%% file: neurips_2022.tex
\documentclass{article}

     \usepackage[final,nonatbib]{neurips_2022}

\usepackage[utf8]{inputenc} 
\usepackage[T1]{fontenc}    
\usepackage{hyperref}       
\usepackage{url}            
\usepackage{booktabs}       
\usepackage{amsfonts}       
\usepackage{nicefrac}       
\usepackage{microtype}      
\usepackage{xcolor}         

\usepackage{amsmath}
\usepackage{amssymb}
\usepackage{multirow}
\usepackage{colortbl}
\usepackage{algorithm}
\usepackage{hyperref}
\usepackage[pdftex]{graphicx}
\usepackage[font=footnotesize]{subcaption}
\usepackage{listings}
\usepackage{wrapfig}
\usepackage{etoolbox}
\usepackage{nicematrix}
\usepackage{tabu}

\definecolor{Primary100}{RGB}{231,234,247}
\definecolor{Primary}{RGB}{37,75,177}

\newcommand{\wtmat}[2]{W_{#1 #2}}

\newcommand{\xfor}{\overrightarrow{x}}
\newcommand{\xback}{\overleftarrow{x}}
\newcommand{\hforfor}{\overrightarrow{h_{\rm for}}}

\newcommand{\hbackfor}{\overrightarrow{h_{\rm back}}}
\newcommand{\hbackback}{\overleftarrow{h_{\rm back}}}
\newcommand{\hver}{\mathbf{H}^{\rm ver}}
\newcommand{\hhor}{\mathbf{H}^{\rm hor}}
\newcommand{\igate}{i}
\newcommand{\fgate}{f}
\newcommand{\ogate}{o}
\newcommand{\state}{c}
\newcommand{\lstmfor}[1]{{\rm LSTM}_{\rm for}(#1)}
\newcommand{\lstmback}[1]{{\rm LSTM}_{\rm back}(#1)}
\newcommand{\bilstm}[1]{{\rm BiLSTM}(#1)}
\newcommand{\fc}[1]{{\rm FC}(#1)}
\newcommand{\concat}[1]{{\tt concatenate}(#1)}
\newcommand{\blocka}[4]{\multirow{3}{*}{$\left[\begin{array}{l}\text{BiLSTM2D: }#1\text{d} \\ D=#2 \\ \text{MLP: }#3 \text{ exp. ratio}\end{array}\right] \times#4$}}
\newcommand{\blockb}[4]{\multirow{3}{*}{$\left[\begin{array}{l}\text{BiLSTM: }#1\text{d} \\ D=#2 \\ \text{MLP: }#3 \text{ exp. ratio}\end{array}\right] \times#4$}}

\makeatletter
\AfterEndEnvironment{algorithm}{\let\@algcomment\relax}
\AtEndEnvironment{algorithm}{\kern2pt\hrule\relax\vskip3pt\@algcomment}
\let\@algcomment\relax
\newcommand\algcomment[1]{\def\@algcomment{\footnotesize#1}}
\renewcommand\fs@ruled{\def\@fs@cfont{\bfseries}\let\@fs@capt\floatc@ruled
  \def\@fs@pre{\hrule height.8pt depth0pt \kern2pt}%
  \def\@fs@post{}%
  \def\@fs@mid{\kern2pt\hrule\kern2pt}%
  \let\@fs@iftopcapt\iftrue}
\makeatother

\title{Sequencer: Deep LSTM for Image Classification}

\author{
Yuki~Tatsunami$^{1,2}$\quad
Masato~Taki$^{1}$\quad \\
  $^{1}$Rikkyo University, Tokyo, Japan \\
  $^{2}$AnyTech Co., Ltd., Tokyo, Japan \\
\texttt{\{y.tatsunami, taki\_m\}@rikkyo.ac.jp}
}

\begin{document}

\maketitle

\begin{abstract}
In recent computer vision research, the advent of the Vision Transformer (ViT) has rapidly revolutionized various architectural design efforts: ViT achieved state-of-the-art image classification performance using self-attention found in natural language processing, and MLP-Mixer achieved competitive performance using simple multi-layer perceptrons. In contrast, several studies have also suggested that carefully redesigned convolutional neural networks (CNNs) can achieve advanced performance comparable to ViT without resorting to these new ideas. Against this background, there is growing interest in what inductive bias is suitable for computer vision. Here we propose \textit{Sequencer}, a novel and competitive architecture alternative to ViT that provides a new perspective on these issues. Unlike ViTs, Sequencer models long-range dependencies using LSTMs rather than self-attention layers. We also propose a two-dimensional version of Sequencer module, where an LSTM is decomposed into vertical and horizontal LSTMs to enhance performance. Despite its simplicity, several experiments demonstrate that Sequencer performs impressively well: Sequencer2D-L, with 54M parameters, realizes 84.6\% top-1 accuracy on only ImageNet-1K. Not only that, we show that it has good transferability and the robust resolution adaptability on double resolution-band. Our source code is available at \url{https://github.com/okojoalg/sequencer}.
\end{abstract}

\section{Introduction}
\begin{wrapfigure}{r}{0.45\textwidth}
  \centering
  \includegraphics[width=0.45\textwidth]{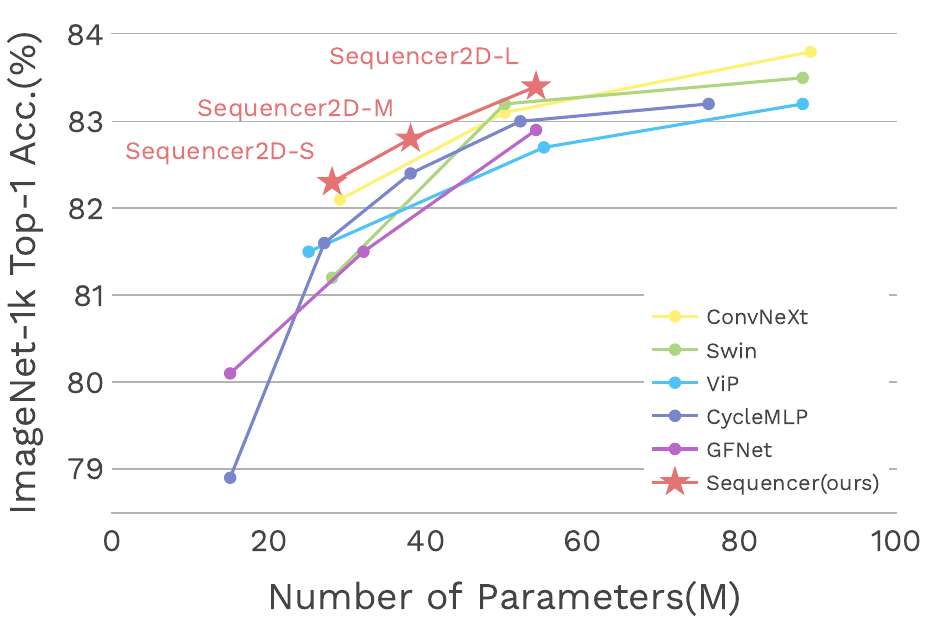}
  \caption{\textbf{IN-1K top-1 accuracy v.s. model parameters.} All models are trained on IN-1K at resolution 224$^2$ from scratch.}
  \label{figure:main}
\end{wrapfigure}
The de-facto standard for computer vision has been convolutional neural networks (CNNs)~\cite{krizhevsky2012imagenet, simonyan2014very, he2016deep, szegedy2015going, szegedy2016rethinking, chollet2017xception, huang2017densely, tan2019efficientnet}. However, inspired by the many breakthroughs in natural language processing (NLP) achieved by Transformers~\cite{vaswani2017attention, devlin2018bert, radford2018improving}, applications of Transformers for computer vision are now being actively studied. In particular, Vision Transformer (ViT)~\cite{dosovitskiy2020image} is a pure Transformer applied to image recognition and achieves performance competitive with CNNs. Various studies triggered by ViT have shown that the state-of-the-art (SOTA) performance can be achieved for a wide range of vision tasks using self-attention alone~\cite{wang2021pyramid, liu2021swin, touvron2021going, liu2022swin, dong2022cswin}, without convolution.

The reason for this success is thought to be due to the ability of self-attention to model long-range dependencies. However, it is still unclear how essential the self-attention is to the effectiveness of Transformers for vision tasks. Indeed, the MLP-Mixer~\cite{tolstikhin2021mlp} based only on multi-layer perceptrons (MLPs) is proposed as an appealing alternative to Vision Transformers (ViTs). In addition, some studies~\cite{liu2022convnet, ding2022scaling} have shown that carefully designed CNNs are still competitive enough with Transformers in computer vision. Therefore, identifying which architectural designs are inherently effective for computer vision tasks is of great interest for current research~\cite{yu2021metaformer}. This paper provides a new perspective on this issue by proposing a novel and competitive alternative to these vision architectures.

We propose the {\it{Sequencer}} architecture, that uses the long short-term memory (LSTM)~\cite{hochreiter1997long} rather than the self-attention for sequence modeling. The macro-architecture design of Sequencer follows ViTs, which iteratively applies token mixing and channel mixing, but the self-attention layer is replaced by one based on LSTMs. In particular, Sequencer uses bidirectional LSTM (BiLSTM)~\cite{schuster1997bidirectional} as a building block. While simple BiLSTM shows a certain level of performance, Sequencer can be further improved by using ideas similar to Vision Permutator (ViP) \cite{hou2022vision}. The key idea in ViP is to process the vertical and horizontal axes in parallel. We also introduce two BiLSTMs for top/bottom and left/right directions in parallel. This modification improves the efficiency and accuracy of Sequencer because this structure reduces the length of the sequence and yields a spatially meaningful receptive field.

When pre-trained on ImageNet-1K (IN-1K) dataset, our new attention-free architecture outperforms advanced architectures such as Swin~\cite{liu2021swin} and ConvNeXt~\cite{liu2022convnet} of comparable size, see Figure \ref{figure:main}. It also outperforms other attention-free and CNN-free architectures such as MLP-Mixer~\cite{tolstikhin2021mlp} and GFNet~\cite{rao2021global}, making Sequencer an attractive new alternative to the self-attention mechanism in vision tasks.

This study also aims to propose novel architecture with practicality by employing LSTM for spatial pattern processing. Notably, Sequencer exhibits robust resolution adaptability, which strongly prevents accuracy degradation even when the input's resolution is increased double during inference. Moreover, fine-tuning Sequencer on high-resolution data can achieve higher accuracy than Swin-B \cite{liu2021swin} and Sequencer is also useful for semantic segmentation. On peak memory, Sequencer tends to be more economical than ViTs and recent CNNs for high-resolution input. Although Sequencer requires more FLOPs than other models due to recursion, the higher resolution improves the relative efficiency of peak memory, enhancing the accuracy/cost trade-off at a high-resolution regime. Therefore, Sequencer also has attractive properties as a practical image recognition model.

\section{Related works}

Inspired by the success of Transformers in NLP~\cite{vaswani2017attention,devlin2018bert,radford2018improving, radford2019language, brown2020language, raffel2020exploring}, various applications of self-attention have been studied in computer vision. For example, in iGPT~\cite{chen2020generative}, an attempt was made to apply autoregressive pre-training with causal self-attention~\cite{radford2018improving} to image classification. However, due to the computational cost of pixel-wise attention, it could only be applied to low-resolution images, and its ImageNet classification performance was significantly inferior to the SOTA. ViT~\cite{dosovitskiy2020image}, on the other hand, quickly brought Transformer's image classification performance closer to SOTA with its idea of applying bidirectional self-attention~\cite{devlin2018bert} to image patches rather than pixels. Various architectural and training improvements~\cite{touvron2020training, yuan2021tokens, wang2021pyramid, zhou2021deepvit, liu2021swin, touvron2021going, chen2021crossvit} have been attempted for ViT~\cite{dosovitskiy2020image}. In this paper, we do not improve self-attention itself but propose a completely new module for image classification to replace it.

The extent to which attention-based cross-token communication inherently contributes to ViT's success is not yet well understood, starting with MLP-Mixer~\cite{tolstikhin2021mlp}, which completely replaced ViT's self-attention with MLP, various MLP-based architectures~\cite{touvron2021resmlp,liu2021pay,hou2022vision, tatsunami2021raftmlp, tang2021sparse, ding2021repmlpnet} have achieved competitive performance on the ImageNet dataset. We refer to these architectures as global MLPs (GMLPs) because they have global receptive fields. This series of studies cast doubt on the need for self-attention. From a practical standpoint, however, these MLP-based models have a drawback: they need to be finetuned to cope with flexible input sizes during inference by modifying the shape of their token-mixing MLP blocks. This resolution adaptability problem has been improved in CycleMLP~\cite{chen2022cyclemlp}, for example, by the idea of realizing a local kernel with a cyclic MLP. There are similar ideas such as \cite{yu2022s2,yu2021s,lian2021mlp,guo2021hire} which are collectively referred to as local MLPs (LMLPs). Besides the MLP-based idea, several other interesting self-attention alternatives have been found. GFNet~\cite{rao2021global} uses Fourier transformation of the tokens and mixes the tokens by global filtering in the frequency domain. PoolFormer~\cite{yu2021metaformer}, on the other hand, achieved competitive performance with only local pooling of tokens, demonstrating that simple local operations are also a suitable alternative. Our proposed Sequencer is a new alternative to self-attention that differs from both of the above, and Sequencer is an attempt to realize token mixing in vision architectures using only LSTM. It achieved competitive performance with SOTA on the IN-1K benchmark, especially with an architecture that can flexibly adapt to higher resolution.

The idea of spatial axis decomposition has been used several times in neural architecture in computer vision. For example, SqueezeNeXt~\cite{gholami2018squeezenext} decomposes a 3x3 convolution layer into 1x3 and 3x1 convolution layers, resulting in a lightweight model. Criss-cross attention~\cite{huang2019ccnet} reduces memory usage and computational complexity by restricting the attention to only vertical and horizontal portions. Current architectures such as CSwin~\cite{dong2022cswin}, Couplformer~\cite{lan2021couplformer}, ViP~\cite{hou2022vision}, RaftMLP~\cite{tatsunami2021raftmlp}, SparseMLP~\cite{tang2021sparse}, and MorphMLP~\cite{zhang2021morphmlp} have included similar ideas to improve efficiency and performance.

In the early days of deep learning, there were attempts to use RNNs for image recognition. The earliest study that applied RNNs to image recognition is \cite{graves2007multi}. The primary difference between our study and \cite{graves2007multi} is that we utilize a usual RNN in place of a 2-multi-dimensional RNN(2MDRNN). The 2MDRNN requires $H+W$ sequential operations; The LSTM requires $H$ sequential operations, where $H$ and $W$ are height and width, respectively. For subsequent work on image recognition using 2MDRNNs, see ~\cite{graves2008offline, kalchbrenner2015grid, byeon2015scene, liang2016semantic}. \cite{byeon2015scene} proposed an architecture in which information is collected from four directions (upper left, lower left, upper right, and lower right) by RNNs for understanding natural scene images. \cite{liang2016semantic} proposed a novel 2MDRNN for semantic object parsing that integrates global and local context information, called LG-LSTM. The overall architecture design is structured to input deep ConvNet features into the LG-LSTM, unlike Sequencer which stacks LSTMs. ReNet~\cite{visin2015renet} is most relevant to our work; ReNet~\cite{visin2015renet} uses a 4-way LSTM and non-overlapping patches as input. In this respect, it is similar to Sequencer. Meanwhile, there are three differences. First, Sequencer is the first MetaFormer~\cite{yu2021metaformer} realized by adopting LSTM as the token mixing block. Sequencer also adopts a larger patch size than ReNet~\cite{visin2015renet}. The benefit of adopting these designs is that we can modernize LSTM-based vision architectures and fairly compare LSTM-based models with ViT. As a result, our results provide further evidence for the extremely interesting hypothesis MetaFormer~\cite{yu2021metaformer}. Second, the way vertical BiLSTMs and horizontal BiLSTMs are connected is different. Our work connects them in parallel, allowing us to gather vertical and horizontal information simultaneously. On the other hand, in ReNet~\cite{visin2015renet}, the output of the horizontal BiLSTM is used as input to the vertical BiLSTM. Finally, we trained Sequencer on large datasets such as ImageNet, whereas ReNet~\cite{visin2015renet} is limited to small datasets as MNIST~\cite{lecun1998gradient}, CIFAR-10~\cite{krizhevsky2009learning}, and SVHN~\cite{netzer2011reading}, and has not shown the effectiveness of LSTM for larger datasets. ReSeg~\cite{visin2016reseg} applied ReNet to semantic segmentation. RNNs have been applied not only to image recognition, but also to generative models: PixcelRNN~\cite{van2016pixel} is a pixel-channel autoregressive generative model of images using Row RNN, which consists of a 1D-convolution and a usual RNN, and Diagonal BiLSTM, which is computationally expensive.

In NLP, attempts have been made to avoid the computational cost of attention by approximating causal self-attention with recurrent neural network (RNN)~\cite{katharopoulos2020transformers} or replacing it with RNN after training~\cite{kasai2021finetuning}. In particular, in~\cite{katharopoulos2020transformers}, an autoregressive pixel-wise image generation task is experimented with an architecture where the attentions in iGPT are approximated by RNNs. These studies are specific to unidirectional Transformers, in contrast to our token-based Sequencer which is the bidirectional analog of them.

\section{Method}
\label{sec:method}
In this section, we briefly recap the preliminary background on LSTM and further describe the details of the proposed architectures.

\begin{figure}[tb]
  \centering
  \begin{minipage}[b]{0.38\hsize}
    \centering
    \includegraphics[height=\linewidth,angle=270]{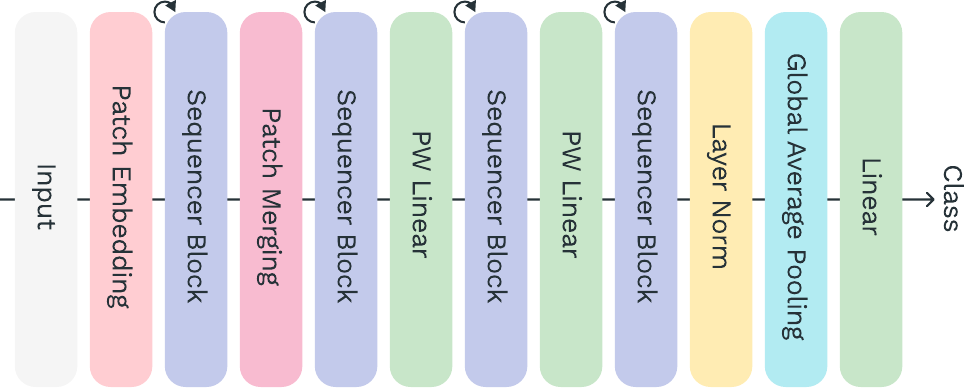}
    \subcaption{Sequencer}
    \label{figure:sequencer}
  \end{minipage}
  \begin{minipage}[b]{0.38\hsize}
    \centering
    \includegraphics[width=\linewidth]{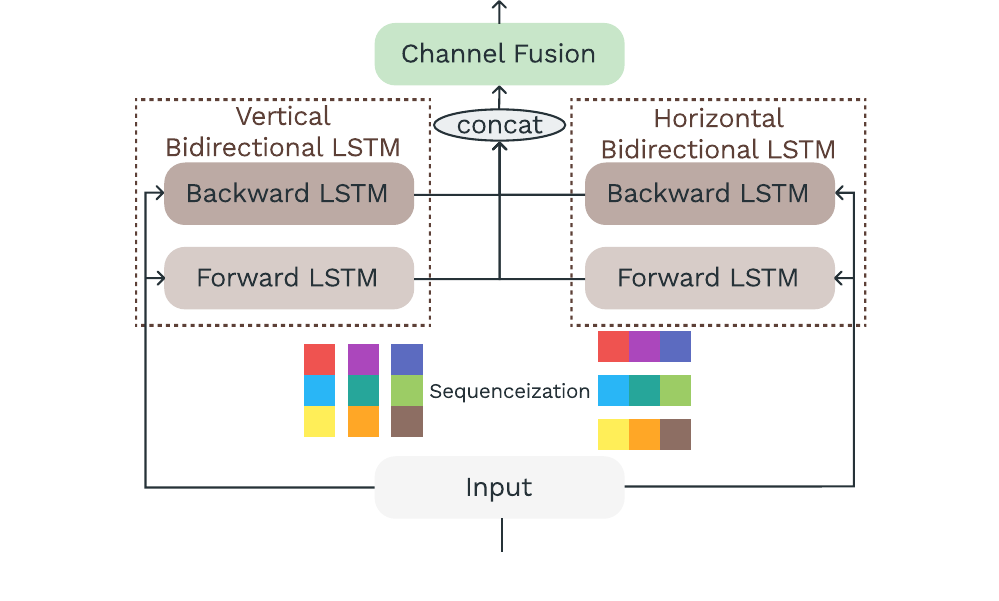}
    \subcaption{BiLSTM2D layer}
    \label{figure:bilstm2d}
  \end{minipage}
  \\
  \begin{minipage}[b]{0.15\hsize}
    \centering
    \includegraphics[width=\linewidth]{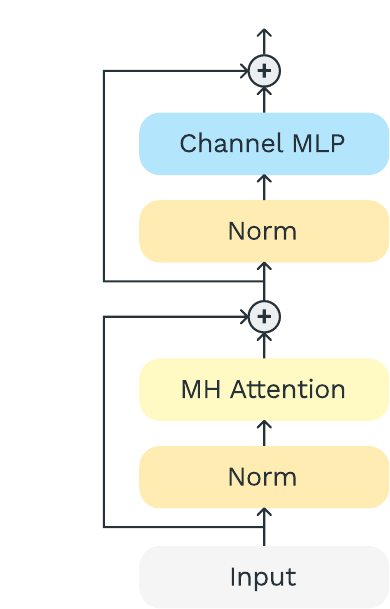}
    \subcaption{Transformer \\block}
    \label{figure:transformer}
  \end{minipage}
  \hspace{0.04\columnwidth}
  \begin{minipage}[b]{0.15\hsize}
    \centering
    \includegraphics[width=\linewidth]{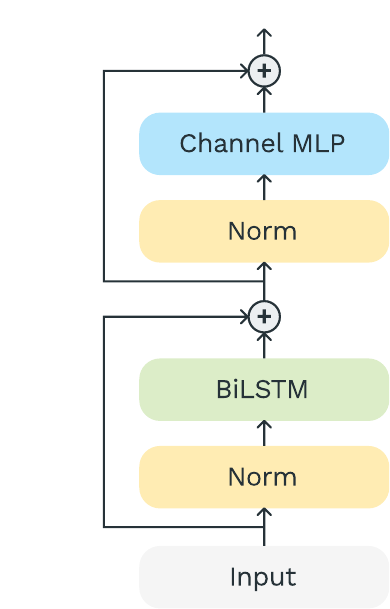}
    \subcaption{Vanilla \\Sequencer block}
    \label{figure:vanila_sequencer}
  \end{minipage}
  \hspace{0.04\columnwidth}
  \begin{minipage}[b]{0.15\hsize}
    \centering
    \includegraphics[width=\linewidth]{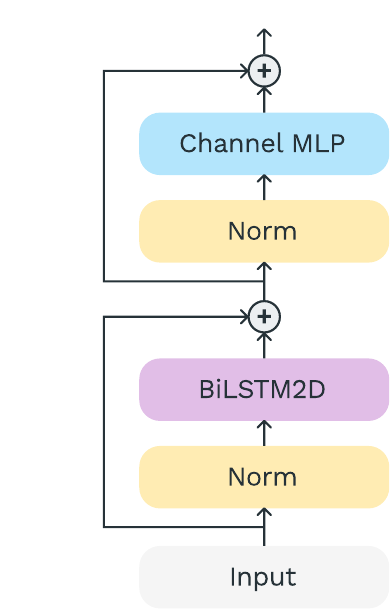}
    \subcaption{Sequencer2D \\block}
    \label{figure:sequencer2d}
  \end{minipage}
  \caption{(\subref{figure:sequencer}) The architecture of Sequencers; (\subref{figure:bilstm2d}) The figure outlines the BiLSTM2D layer, which is the main component of Sequencer2D. (\subref{figure:transformer}) Transformer block consist of multi-head attention. In contrast, (\subref{figure:vanila_sequencer}) Vanilla Sequencer block and (\subref{figure:sequencer2d}) Sequencer2D block, utilized on our archtecture, composed of BiLSTM or BiLSTM2D instead of multi-head attention.}
    \label{figure:overall}
\end{figure}

\subsection{Preliminaries: Long short-term memory}
\label{subsec:lstm}
LSTM~\cite{hochreiter1997long} is a specialized recurrent neural network (RNN) for modeling long-term dependencies of sequences. Plain LSTM has an input gate $\igate_t$ that controls the storage of inputs, a forget gate $\fgate_t$ that controls the forgetting of the former cell state $\state_{t-1}$ and an output gate $\ogate_t$ that controls the cell output $h_t$ from the current cell state $\state_t$. Plain LSTM is formulated as follows:
\begin{align}
\igate_t &= \sigma\left(\wtmat{x}{\igate} x_t + \wtmat{h}{\igate} h_{t-1} + b_\igate\right),\;\;
\fgate_t = \sigma\left(\wtmat{x}{\fgate} x_t + \wtmat{h}{\fgate} h_{t-1} + b_\fgate \right),\\
\state_t &= \fgate_t \odot \state_{t-1} + \igate_t \odot \tanh \left(\wtmat{x}{\state} x_t + \wtmat{h}{\state} h_{t-1} + b_\state \right),\;\;
\ogate_t = \sigma\left(\wtmat{x}{\ogate} x_t + \wtmat{h}{\ogate} h_{t-1} + b_\ogate\right),\\
h_t &= \ogate_t \odot \tanh(\state_t),
\end{align}
where $\sigma$ is the logistic sigmoid function and $\odot$ is Hadamard product.

BiLSTM~\cite{schuster1997bidirectional} is profitable for sequences where mutual dependencies are expected. A BiLSTM consists of two plain LSTMs. Let $\xfor$ be the input series and $\xback$ be the rearrangement of $\xfor$ in reverse order. $\hforfor$ and $\hbackback$ are the outputs obtained by processing $\xfor$ and $\xback$ with the corresponding LSTMs, respectively. Let $\hbackfor$ be the output $\hbackback$ rearranged in the original order, and the output of BiLSTM is obtained as follows:
\begin{align}
\hforfor,\,\hbackback &= \lstmfor{\xfor},\,\lstmback{\xback},\;\; h = \concat{\hforfor, \hbackfor}\,.
\end{align}
Assume that both $\hforfor$ and $\hbackfor$ have the same hidden dimension $D$, which is hyperparameter of BiLSTM. Accordingly, vector $h$ has dimension $2D$.

\subsection{Sequencer architecture}
\label{subsec:sequencer}
\paragraph{Overall architecture}
In the last few years, ViT and its many variants based on self-attention~\cite{dosovitskiy2020image,  touvron2020training, liu2021swin, zhou2021refiner} have attracted much attention in computer vision. Following these, several works~\cite{tolstikhin2021mlp, touvron2021resmlp, liu2021pay, hou2022vision} have been proposed to replace self-attention with MLP. There have also been studies of replacing self-attention with a hard local induced bias module~\cite{chen2022cyclemlp, yu2021metaformer} and with a global filter~\cite{rao2021global} using the fast Fourier transform algorithm (FFT)~\cite{cooley1965algorithm}. This paper continues this trend and attempts to replace the self-attention layer with LSTM~\cite{hochreiter1997long}: we propose a new architecture aiming at memory saving by mixing spatial information with LSTM, which is memory-economical compared to ViT, parameter-saving, and has the ability to learn long-range dependencies.

Figure \ref{figure:sequencer} shows the overall structure of Sequencer architecture. Sequencer architecture takes non-overlapping patches as input and projects them onto the feature map. Sequencer block, which is a core component of Sequencer, consists of the following sub-components:
(1) BiLSTM layer can mix spatial information more memory-economically for high-resolution images than Transformer layer and more globally than CNN.
(2) MLP for channel-mixing as well as~\cite{dosovitskiy2020image, tolstikhin2021mlp}.
Sequencer block is called Vanilla Sequencer block when plain BiLSTM layers are used as BiLSTM layers as Figure \ref{figure:vanila_sequencer} and Sequencer2D block when BiLSTM2D layers are used as Figure \ref{figure:sequencer2d}. We define BiLSTM2D layer later. The output of the last block is sent to the linear classifier via the global average pooling layer, as in most other architectures.

\paragraph{BiLSTM2D layer}
We propose the BiLSTM2D layer as a technique to mix 2D spatial information efficaciously. It has two plain BiLSTMs: a vertical BiLSTM and a horizontal one. For an input $\mathbf{X}\in\mathbb{R}^{H\times W\times C}$, $\{\mathbf{X}_{:, w, :}\in \mathbb{R}^{H\times C}\}_{w=1}^W$ is viewed as a set of sequences, where $H$ is the number of tokens in the vertical direction, $W$ is the number of sequences in the horizontal direction, and $C$ is the channel dimension. All sequences $\mathbf{X}_{:, w, :}$ are input into the vertical BiLSTM with shared weights and hidden dimension $D$:
\begin{align}
\hver_{:, w, :} &= \bilstm{\mathbf{X}_{:, w, :}}.
\end{align}
In a very similar manner, $\{\mathbf{X}_{h, :, :}\in \mathbb{R}^{W\times C}\}_{h=1}^H$ is viewed as a set of sequences, and all sequences $\mathbf{X}_{h, :, :}$ are input into the horizontal BiLSTM with shared weights and hidden dimension $D$ as well:
\begin{align}
\hhor_{h, :, :} &= \bilstm{\mathbf{X}_{h, :, :}}.
\end{align}
We combine $\{\hver_{:, w, :}\in \mathbb{R}^{H\times 2D}\}_{w=1}^W$ into $\hver\in\mathbb{R}^{W\times H\times 2D}$ and $\{\hhor_{h, :, :}\in\mathbb{R}^{W\times 2D}\}_{h=1}^H$ into $\hhor\in\mathbb{R}^{W\times H\times 2D}$.
They are then concatenated and processed point-wisely in a fully-connection layer. These processes are formulated as follows:
\begin{align}
\mathbf{H} &= \concat{\hver, \hhor},\;\; \hat{\mathbf{X}} = \fc{\mathbf{H}},
\end{align}
where $\fc{\cdot}$ denotes the fully-connected layer with weight $W\in \mathbb{R}^{C\times 4D}$. The \texttt{PyTorch}-like pseudocode is shown in Appendix \ref{subsec:pseudocode}.

BiLSTM2D is more memory-economical and throughput-efficiency than multi-head-attention of ViT for high-resolution input. BiLSTM2D involves $(WC+HC)/2$ dimensional cell states, while a multi-head-attention involves $h*(HW)^2$ dimensional attention map where $h$ is a number of heads. Thus, as $H$ and $W$ increase, the memory cost of an attention map increases more rapidly than the cost of a cell state. On throughput, the computational complexity of self-attention is $\mathcal{O}(W^4 C)$, whereas the computational complexity of BiLSTM is $\mathcal{O}(WC^2)$ where we assume $W=H$ for simplicity. There are $\mathcal{O}(W)$ sequential operations for BiLSTM2D. Therefore, assuming we use a sufficiently efficient LSTM cell implementation, such as official PyTorch LSTMs we are using, the increase of the complexity of self-attention is much more rapid than BiLSTM2D. It implies a lower throughput of attention compared to BiLSTM2D. See an experiment in Section \ref{subsec:analysis_visualization}.

\paragraph{Architecture variants}

For comparison between models of different depths consisting of Sequencer2D blocks, we have prepared  three models with different depths: 18, 24, and 36. The names of the models are \textit{Sequencer2D-S}, \textit{Sequencer2D-M}, and \textit{Sequencer2D-L}, respectively. The hidden dimension is set to $D=C/4$. Details of these models are provided in Appendix \ref{subsec:architecture_details}.

As shown in subsection \ref{subsec:in1k}, these architectures outperform typical models. Interestingly, however, subsection \ref{subsec:ablation_studies} shows that replacing Sequencer2D block with the simpler Vanilla Sequencer block maintains moderate accuracy. We denote such a model as Vanilla Sequencer. Note that some of the explicit positional information is lost in the \textit{Vanilla Sequencer} because the model treats patches as a 1D sequence.

\section{Experiments}
\label{sec:expriments}

\begin{table}[tb]
\centering
\caption{The table shows the top-1 accuracy when trained on IN-1K, comparing our model with other similar scale representative models. Training and inference throughput and their peak memory were measured with 16 images per batch on a single V100 GPU. The left sides of the slashes are values during training, and the right sides of the slashes are values during inference.
 Fine-tuned models marked with "$\uparrow$". Note Sequencer2D-L$\uparrow$ are compared to Swin-B$\uparrow$ and ConvNeXt-B$\uparrow$ with more parameters since Swin and ConvNeXt have not fine-tuned models of similar parameters with Sequencer2D-L$\uparrow$ in the original papers.}
\small
\setlength\tabcolsep{2.36pt}
\begin{NiceTabular}{lcccrrrcc}[colortbl-like]
      \multirow{2}{*}{Model} & \multirow{2}{*}{Family} & \multirow{2}{*}{Res.} & \multirow{2}{*}{\#Param.} & \multirow{2}{*}{FLOPs} & Throughput & Peak Mem. & Top-1 & Pre FT Top-1 \\
            &        &         &       &       & (image/s) & (MB) &  Acc.(\%) &  Acc.(\%) \\
\midrule
\multicolumn{9}{c}{Training from scratch} \\
\midrule
RegNetY-4GF~\cite{radosavovic2020designing} &    CNN & 224$^2$ &     21M &  4.0G & 228/823 & 1136/225 &        80.0 & non-fine-tune \\
ConvNeXt-T~\cite{liu2022convnet} &    CNN & 224$^2$ &     29M &  4.5G & 337/1124 & 1418/248 &        82.1 & as above \\
DeiT-S~\cite{touvron2020training} & Trans. & 224$^2$ &     22M & 4.6G & 480/1569 & 1195/180 &        79.9 & as above \\
Swin-T~\cite{liu2021swin} & Trans. & 224$^2$ &     28M &  4.5G & 268/894 & 1613/308 &        81.2 & as above \\
ViP-S/7~\cite{hou2022vision} &  GMLP & 224$^2$ &     25M &  6.9G & 214/702 & 1587/195 &        81.5 & as above \\
CycleMLP-B2~\cite{chen2022cyclemlp} &  LMLP & 224$^2$ &     27M &  3.9G & 158/586 & 1357/234 &        81.6 & as above \\
PoolFormer-S24~\cite{yu2021metaformer} &  LMLP & 224$^2$ &     21M & 3.6G & 313/988 & 1461/183 &        80.3 & as above \\
\rowcolor{Primary100}Sequencer2D-S &   Seq. & 224$^2$ &     28M & 8.4G & 	110/347 & 1799/196 &         \textbf{82.3} & as above \\
\midrule
RegNetY-8GF~\cite{radosavovic2020designing} &    CNN & 224$^2$ &     39M &  8.0G & 	211/751 & 1776/333 &        81.7 & as above \\
T2T-ViT$_t$-19~\cite{yuan2021tokens} & Trans. & 224$^2$ &     39M & 9.8G & 197/654 & 3520/1140 &        82.2 & as above \\
CycleMLP-B3~\cite{chen2022cyclemlp} &  LMLP & 224$^2$ &     38M &  6.9G & 100/367 & 2326/287 &        82.6 & as above \\
PoolFormer-S36~\cite{yu2021metaformer} &  LMLP & 224$^2$ &     31M & 5.2G & 	213/673 & 2187/220 &        81.4 & as above \\
GFNet-H-S~\cite{rao2021global} &  FFT & 224$^2$ &     32M & 4.5G & 227/755 & 1740/282 &        81.5 & as above \\
\rowcolor{Primary100}Sequencer2D-M &   Seq. & 224$^2$ &    38M & 11.1G & 	83/270 & 2311/244 &        \textbf{82.8} & as above\\
\midrule
RegNetY-12GF   ~\cite{radosavovic2020designing} &    CNN & 224$^2$ &     46M &  12.0G & 	199/695 & 2181/440 &        82.4 & as above \\
ConvNeXt-S~\cite{liu2022convnet} &    CNN & 224$^2$ &     50M &  8.7G & 	212/717 & 2265/341 &        83.1 & as above \\
 
Swin-S~\cite{liu2021swin} & Trans. & 224$^2$ &     50M &  8.7G & 	165/566 & 2635/390 &        83.2 & as above \\
Mixer-B/16~\cite{tolstikhin2021mlp} &  GMLP & 224$^2$ &     59M & 12.7G & 	338/1011 & 1864/407 &        76.4 & as above \\
ViP-M/7~\cite{hou2022vision} &  GMLP & 224$^2$ &     55M & 16.3G & 	130/395 & 3095/396 &        82.7 & as above \\
CycleMLP-B4~\cite{chen2022cyclemlp} &  LMLP & 224$^2$ &     52M &  10.1G & 	70/259 & 3272/338 &        83.0 & as above \\
PoolFormer-M36~\cite{yu2021metaformer} &  LMLP & 224$^2$ &     56M & 9.1G & 	171/496 & 3191/368 &        82.1 & as above \\
GFNet-H-B~\cite{rao2021global} &  FFT & 224$^2$ &     54M & 8.4G & 144/482 & 2776/367 &        82.9 & as above \\
\rowcolor{Primary100}Sequencer2D-L &   Seq. & 224$^2$ &    54M & 16.6G & 	54/173 & 3516/322 &    \textbf{83.4} & as above \\
\midrule
\multicolumn{9}{c}{Fine-tuning} \\
\midrule
ConvNeXt-B$\uparrow$~\cite{liu2022convnet} &    CNN & 384$^2$ &     89M &  45.1G & 78/234 & 7329/870 &        85.1(+1.3) & 83.8 \\
Swin-B$\uparrow$~\cite{liu2021swin} & Trans. & 384$^2$ &     88M &  47.1G & 54/156 & 12933/1532 &        84.5(+1.0) & 83.5 \\
GFNet-B$\uparrow$~\cite{rao2021global} &  FFT & 384$^2$ &     47M & 23.2G & 137/390 & 3710/416 &   82.1(+0.8) & 82.9     \\
\rowcolor{Primary100}Sequencer2D-L$\uparrow$ &   Seq. & 392$^2$ &     54M & 50.7G & 26/84 & 9062/481 & 84.6(+1.2) & 83.4    \\
\bottomrule
\end{NiceTabular}
\label{table:main}
\end{table}
In this section, we compare Sequencers with previous studies on the IN-1K benchmark~\cite{krizhevsky2012imagenet}. We also carry out ablation studies, transfer learning studies, and analysis of the results to demonstrate the effectiveness of Sequencers. We adopt \texttt{PyTorch}~\cite{paszke2019pytorch} and \texttt{timm}~\cite{rw2019timm} library to implement models in the conduct of all experiments. See Appendix \ref{sec:impl_details} for more setup details.

\subsection{Scratch training on IN-1K}
\label{subsec:in1k}
We utilize IN-1K~\cite{krizhevsky2012imagenet}, which has 1000 classes and contains 1,281,167 training images and 50,000 validation images. We adopt AdamW optimizer~\cite{loshchilov2017decoupled}. Following the previous study~\cite{touvron2020training}, we adopt the base learning rate $\frac{\rm {batch\, size}}{512} \times 5 \times 10^{-4}$.  The batch sizes for Sequencer2D-S, Sequencer2D-M, and Sequencer2D-L are 2048, 1536, and 1024, respectively. As a regularization method, stochastic depth~\cite{huang2016deep} and label smoothing~\cite{szegedy2016rethinking} are employed. As data augmentation methods, mixup~\cite{zhang2017mixup}, cutout~\cite{devries2017improved}, cutmix~\cite{yun2019cutmix}, random erasing~\cite{zhong2020random}, and randaugment~\cite{cubuk2020randaugment} are applied. 

Table \ref{table:main} shows the results that are comparing the proposed models to others with a comparable number of parameters to our models, including models with local and global receptive fields such as CNNs, ViTs, and MLP-based and FFT-based models. Sequencers has the disadvantage that its throughput is slower than other models because it uses RNNs. In the scratch training on IN-1K, however, they outperform these recent comparative models in accuracy across their parameter bands. In particular, Seqeuncer2D-L is competitive with recently discussed models with comparable parameters such as ConvNeXt-S~\cite{liu2022convnet} and Swin-S~\cite{liu2021swin}, with accuracy outperformance of 0.3\% and 0.2\%, respectively.

Table \ref{table:main} demonstrates that Sequencer’s throughput is not good. The training throughput is about three times the inference throughput for all these models. Compared to other models, both measured inference and training time are not good.

\subsection{Fine-tuning on IN-1K}
\label{subsec:ft_in1k}
In this fine-tuning study, Sequencer2D-L pre-trained on IN-1K at 224$^2$ resolution is fine-tuned on IN-1K at 392$^2$ resolution. We compare it with the other models fine-tuned on IN-1K at 384$^2$ resolution. Since 392 is divisible by 14, the input at this resolution can be split into patches without padding. However, note that this is not the case with a resolution of 384$^2$.

As Table \ref{table:main} indicates, even when higher-resolution Sequencer is fine-tuned, it is competitive with the latest models such as ConvNeXt~\cite{liu2022convnet}, Swin~\cite{liu2021swin}, and GFNet~\cite{rao2021global}.

\subsection{Ablation studies}
\label{subsec:ablation_studies}
\begin{table}[tb]
\small
\centering
\caption{\textbf{Sequencer ablation experiments}. We adopt Sequencer2D-S variant for these ablation studies. \textbf{C1} denotes vertical BiLSTM, \textbf{C2} denotes horizontal BiLSTM, and \textbf{C3} denotes channel fusion component. When vertical BiLSTM only, horizontal BiLSTM only or unidirectional BiLSTM2D, its hidden dimension needs to be doubled from the original setting because it compensates the output dimension for the excluded LSTM and matches the dimensions.}
\begin{minipage}[t]{0.2\linewidth}{
\subcaption{Components\label{table:components}}
\setlength\tabcolsep{2.36pt}
\begin{center}
\begin{NiceTabular}{cccc}[colortbl-like]
C1 & C2 & C3 & Acc. \\
\hline
\checkmark &  & & 75.6 \\
 & \checkmark & & 75.0  \\
\checkmark & \checkmark & & 81.6 \\
\rowcolor{Primary100}\checkmark & \checkmark & \checkmark & \textbf{82.3} \\
\end{NiceTabular}
\end{center}
}\end{minipage}
\begin{minipage}[t]{0.2\linewidth}{
\subcaption{LSTM Direction\label{table:directional}}
\setlength\tabcolsep{2.36pt}
\begin{center}
\begin{NiceTabular}{cc}[colortbl-like]
Bidirectional & Acc. \\
\hline
 & 79.7 \\
\rowcolor{Primary100}\checkmark & \textbf{82.3} \\
\end{NiceTabular}
\end{center}
}\end{minipage}
\begin{minipage}[t]{0.45\linewidth}{
\subcaption{Vanilla Sequencer\label{table:vanilla}}
\setlength\tabcolsep{2.36pt}
\begin{center}
\begin{NiceTabular}{lccc}[colortbl-like]
    Model & \#Params. & FLOPs & Acc. \\
\hline
VSequencer-S & 33M & 8.4G & 78.0 \\
VSequencer(H)-S & 28M & 8.4G & 78.8 \\
VSequencer(PE)-S & 33M & 8.4G & 78.1 \\
\rowcolor{Primary100}Sequencer2D-S & 28M & 8.4G & \textbf{82.3} \\
\end{NiceTabular}
\end{center}
}
\end{minipage}
\begin{minipage}[t]{0.45\linewidth}{
\subcaption{Hidden dimension\label{table:hidden_dimension}}
\setlength\tabcolsep{2.36pt}
\begin{center}
\begin{NiceTabular}{cccc}[colortbl-like]
Hidden dim. ratio & \#Params. & FLOPs & Acc. \\
\hline
\rowcolor{Primary100}1x & \textbf{28M} & \textbf{8.4G} & 82.3 \\
2x & 45M & 13.9G & \textbf{82.6} \\
\end{NiceTabular}
\end{center}
}\end{minipage}
\begin{minipage}[t]{0.43\linewidth}{
\setlength\tabcolsep{2.36pt}
\subcaption{Various RNNs\label{table:various_rnns}}
\begin{center}
\begin{NiceTabular}{lccc}[colortbl-like]
Model & \#Params. & FLOPs & Acc. \\
\hline
RNN-Sequencer2D & 19M & 5.8G & 80.6 \\
GRU-Sequencer2D & 25M & 7.5G & \textbf{82.3} \\
\rowcolor{Primary100}Seqeucer2D-S & 28M & 8.4G & \textbf{82.3} \\
\end{NiceTabular}
\end{center}
}\end{minipage}
\label{table:ablations}
\end{table}

This subsection presents ablation studies based on Sequencer2D-S for further understanding of Sequencer. We seek to clarify the effectiveness and validity of the Sequencers architecture in terms of the importance of each component, bidirectional necessaries, setting of the hidden dimension, and the comparison with simple BiLSTM.

We show where and how relevant the components of BiLSTM2D are: The BiLSTM2D is composed of vertical BiLSTM, horizontal BiLSTM, and channel fusion elements. We want to see the validity of vertical BiLSTM, horizontal BiLSTM, and channel fusion. For this purpose, we examine the removal of channel fusion and vertical or horizontal BiLSTM. Table \ref{table:components} shows the results. Removing channel fusion shows that the performance degrades from 82.3\% to 81.6\%. Furthermore, the additional removal of vertical or horizontal BiLSTM exposes a 6.0\% or 6.6\% performance drop, respectively. Hence, each component discussed here is necessary for Sequencer2D.

We show that the bidirectionality for BiLSTM2D is important for Sequencer. We compare Sequencer2D-S with a version that replaces the vertical and horizontal BiLSTMs with vertical and horizontal unidirectional LSTMs. Table \ref{table:directional} shows that the unidirectional model is 2.6\% less accurate than the bidirectional model. This result attests to the significance of using not unidirectional LSTM but BiLSTM.

It is important to set the hidden dimension of LSTM to a reasonable size. As described in subsection \ref{subsec:sequencer}, Sequencer2D sets the hidden dimension $D$ of BiLSTM to $D=C/4$, but this is not necessary if the model has channel fusions. Table \ref{table:hidden_dimension} compares Sequencer2D-S with the model with increased $D$. Although accuracy is 0.3\% improved, FLOPs increase by 65\%, and the number of parameters increases by 60\%. Namely, the accuracy has not improved for the increase in FLOPs. Moreover, the increase in dimension causes overfitting, which is discussed in Appendix \ref{subsec:overfitting}.

Vanilla Sequencer can also achieve accuracy that outperforms MLP-Mixer \cite{tolstikhin2021mlp}, but is not as accurate as Sequencer2D. Following experimental result supports the claim. We experiment with the Sequencer2D-S variants, where Vanilla Sequencer blocks replace the Sequencer2D blocks, called VSequencer-S(H), with incomplete positional information. In addition, we experiment with a variant of VSequencer-S(H) without the hierarchical structure, which we call VSequencer-S. VSequencer-S(PE) is VSequencer-S using ViTs-style learned positional embedding (PE) ~\cite{dosovitskiy2020image}. Table \ref{table:vanilla} indicates effectiveness for combination of LSTM and ViTs-like architecture. Surprisingly, even with Vanilla Sequencer and Vanilla Sequencer(H) without PE, the performance reduction from Sequencer2D-S is only 4.3\% and 3.5\%, respectively. According to these results, there is no doubt that Vanilla Sequencer using BiLSTMs is significant enough, although not as accurate as Sequencer2D.

All LSTMs in the BiLSTM2D layer can be replaced with other recurrent networks such as gated recurrent units (GRUs)~\cite{cho2014properties} or tanh-RNNs to define BiGRU2D layer or BiRNN2D layer. We also trained these models on IN-1K, so see Table \ref{table:various_rnns} for the results. The table suggests that all of these variants, including RNN-cell, work well. Also, tanh-RNN performs slightly worse than others, probably due to its lower ability to model long-range dependence.

\subsection{Transfer learning and semantic segmentation}
\label{subsec:transfer_learning}
Sequencers perform well on IN-1K, and they have good transferability. In other words, they have satisfactory generalization performance for a new domain, which is shown below. We utilize the commonly used CIFAR-10~\cite{krizhevsky2009learning}, CIFAR-100~\cite{krizhevsky2009learning}, Flowers-102~\cite{nilsback2008automated}, and Stanford Cars~\cite{krause20133d} for this experiment. See the references and Appendix \ref{subsec:tfl_settings} for details on the datasets. The results of the proposed model and the results in previous studies of models with comparable capacity are presented in Table \ref{table:transfer_learning}. In particular, Sequencer2D-L achieves results that are competitive with CaiT-S-36~\cite{touvron2021going} and EfficientNet-B7~\cite{tan2019efficientnet}.

We experiment for semantic segmentation on ADE20K\cite{zhou2017scene} dataset. See Appendix \ref{subsec:segmentation} for details on the setup. Sequencer outperforms PVT~\cite{wang2021pyramid} and PoolFormer~\cite{yu2021metaformer} with similar parameters; compared to PoolFormer, mIoU is about 6 pts higher.

We have investigated a commonly object detection model with Sequencer as the backbone. Its performance is not much different from the case of ResNet~\cite{he2016deep} backbone. Its improvement is the future work. See Appendix \ref{subsec:detection}.

\begin{table}[tb]
\caption{\textbf{Left.} Results on transfer learning. We transfer models trained on IN-1K to datasets from different domains. Sequencers use 224$^2$ resolution images, while ViT-B/16 and EfficientNet-B7 work on higher resolution, see Res. column. \textbf{Right.} Semantic segmentation results on ADE20K~\cite{zhou2017scene}. All models are Semantic FPN~\cite{kirillov2019panoptic} based. We show mIoU for the ADE20k validation set.}
\begin{minipage}[t]{0.64\textwidth}
\vspace{0pt}
\label{table:transfer_learning}
\centering
\small
\setlength\tabcolsep{2.36pt}
\begin{NiceTabular}{lcrrcccc}[colortbl-like]
Model & Res. & \#Pr. & FLOPs & CF$_{10}$ & CF$_{100}$ & Flowers & Cars \\
\hline
ResNet50~\cite{he2016deep} & 224$^2$ & 26M & 4.1G & - & - & 96.2 & 90.0 \\
EN-B7~\cite{tan2019efficientnet} & 600$^2$ & 26M & 37.0G & 98.9 & 91.7 & 98.8 & 94.7 \\
ViT-B/16~\cite{dosovitskiy2020image} & 384$^2$ & 86M & 55.4G & 98.1 & 87.1 & 89.5 & - \\
DeiT-B~\cite{touvron2020training} & 224$^2$ & 86M & 17.5G & 99.1 & 90.8 & 98.4 & 92.1 \\
CaiT-S-36~\cite{touvron2021going} & 224$^2$ & 68M & 13.9G & 99.2 & 92.2 & 98.8 & 93.5 \\
ResMLP-24~\cite{touvron2021resmlp} & 224$^2$ & 30M & 6.0G & 98.7 & 89.5 & 97.9 & 89.5 \\
GFNet-H-B~\cite{rao2021global} & 224$^2$ & 54M & 8.6G & 99.0 & 90.3 & 98.8 & 93.2 \\
\midrule
Sequencer2D-S & 224$^2$ & 28M & 8.4G & 99.0 & 90.6 & 98.2 & 93.1 \\
Sequencer2D-M & 224$^2$ & 38M & 11.1G & 99.1 & 90.8 & 98.2 & 93.3 \\
Sequencer2D-L & 224$^2$ & 54M & 16.6G & 99.1 & 91.2 & 98.6 & 93.1 \\
\bottomrule
\end{NiceTabular}
\end{minipage}
\begin{minipage}[t]{0.35\textwidth}
\vspace{0pt}
\centering
\small
\setlength\tabcolsep{2.36pt}
\begin{NiceTabular}{lcc}[colortbl-like]
Model         & \#Pr. & mIoU \\
\hline
PVT-Small~\cite{wang2021pyramid}& 28M& 39.8 \\
PoolFormer-S24~\cite{yu2021metaformer}& 23M& 40.3 \\
\rowcolor{Primary100}Sequencer2D-S & 32M & 46.1 \\ \hline
PVT-Medium~\cite{wang2021pyramid}& 48M& 41.6 \\
PoolFormer-S36\cite{yu2021metaformer}& 35M& 42.0 \\
\rowcolor{Primary100}Sequencer2D-M & 42M & 47.3 \\ \hline
PVT-Large~\cite{wang2021pyramid}& 65M& 42.1 \\
PoolFormer-M36~\cite{yu2021metaformer}& 60M& 42.4 \\
\rowcolor{Primary100}Sequencer2D-L & 58M & 48.6 \\ \hline
\end{NiceTabular}
\label{table:segmentation}
\end{minipage}
\end{table}

\subsection{Analysis and visualization}
\label{subsec:analysis_visualization}
In this subsection, we investigate the properties of Sequencer in terms of resolution adaptability and efficiency. Furthermore, effective receptive field (ERF)~\cite{luo2016understanding} and visualization of the hidden states provides insight into the question of how Sequencer recognizes images.

\begin{figure}[ht]
\centering
\begin{minipage}[t]{0.75\textwidth}
\vspace{0pt}
\addtolength{\tabcolsep}{-1pt}
\captionsetup{width=0.92\textwidth}
\small
  \centering
  \begin{minipage}[b]{0.49\hsize}
    \centering
    \includegraphics[width=\linewidth]{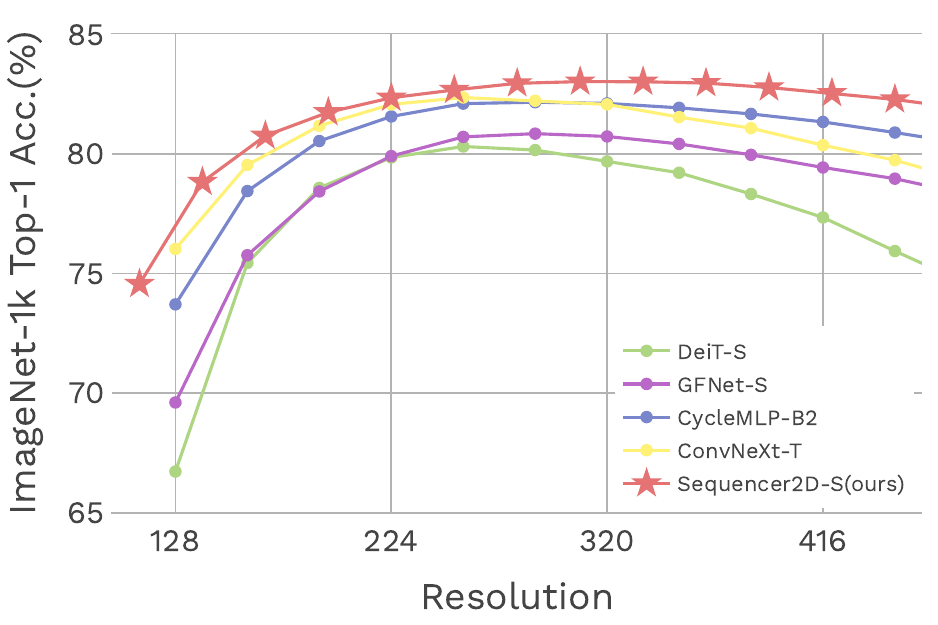}
    \subcaption{Absolute top-1
acc.}
    \label{figure:resolution_adaptability_org}
  \end{minipage}
  \begin{minipage}[b]{0.49\hsize}
    \centering
    \includegraphics[width=\linewidth]{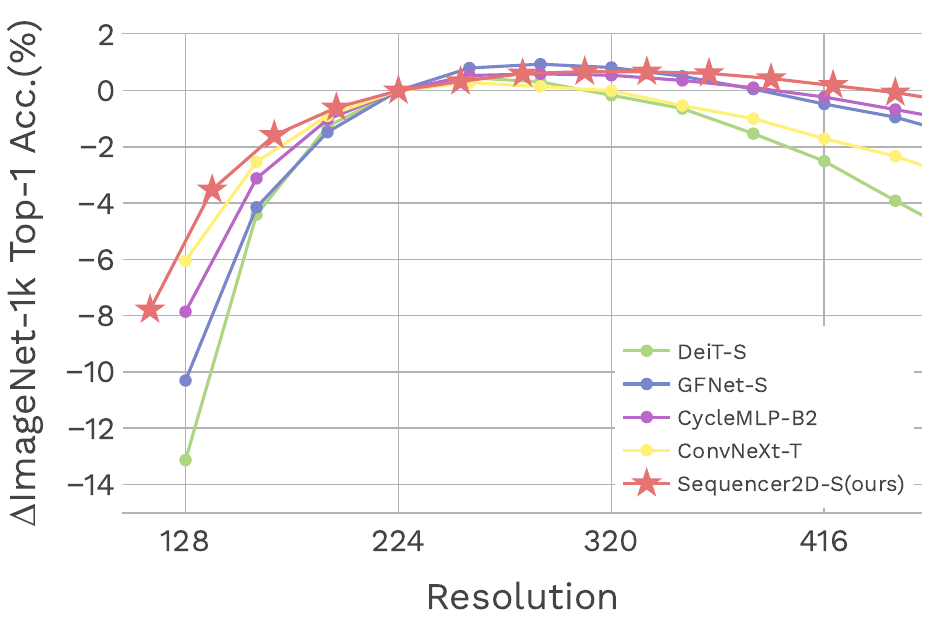}
    \subcaption{Relative top-1 acc. to 224$^2$ res.}
    \label{figure:resolution_adaptability_delta}
  \end{minipage}
  \centering
    \begin{minipage}[b]{0.49\hsize}
    \centering
    \includegraphics[width=\linewidth]{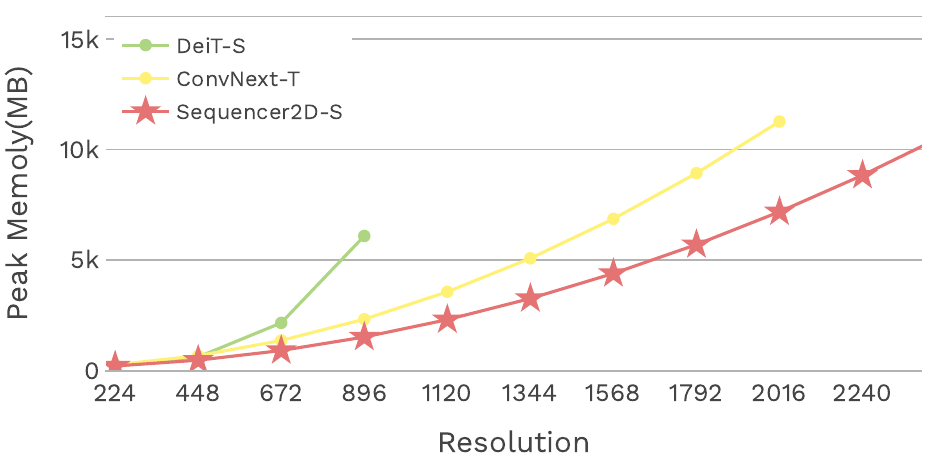}
    \subcaption{GPU peak memory}
    \label{figure:efficiency_peak_memory}
  \end{minipage}
  \begin{minipage}[b]{0.49\hsize}
    \centering
    \includegraphics[width=\linewidth]{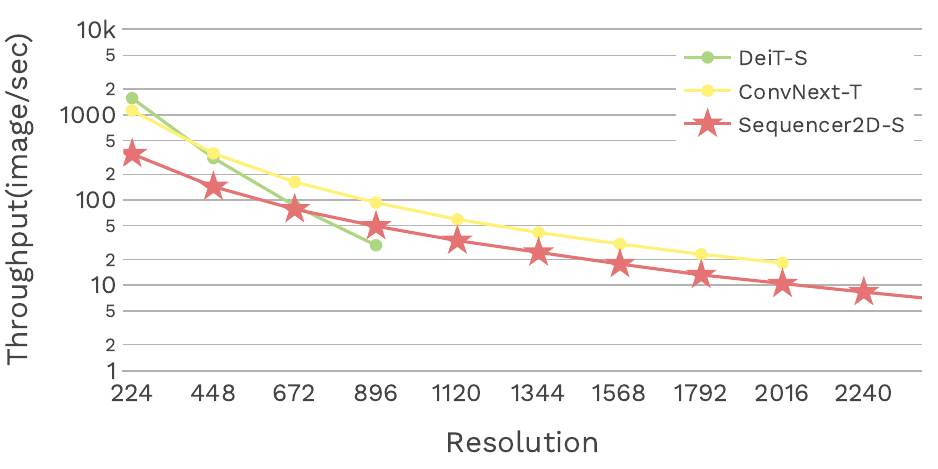}
    \subcaption{Throughput}
    \label{figure:efficiency_throughput}
  \end{minipage}
    \caption{\textbf{Top.} Resolution adaptability. Every model is trained at 224$^2$ resolution and evaluated at various resolutions with no fine-tuning. \textbf{Bottom.} Comparisons among Sequencer2D-S, DeiT-S~\cite{touvron2020training}, and ConvNeXt-T~\cite{liu2022convnet} in (\subref{figure:efficiency_peak_memory}) GPU peak memory and (\subref{figure:efficiency_throughput}) throughput for different input image resolutions. Measured for each increment of 224$^2$ resolution, points not plotted are when GPU memory is exhausted. The measurements are founded on a batch size of 16 and a single V100.}
    \label{figure:resolution_adaptability_efficiency}
\end{minipage}
\begin{minipage}[t]{0.23\textwidth}
\vspace{0pt}
\centering
  \centering
    \begin{minipage}[h]{0.48\hsize}
    \centering
    \includegraphics[width=\linewidth]{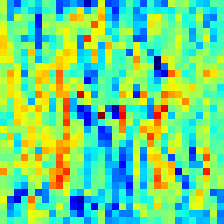}
  \end{minipage}
    \begin{minipage}[h]{0.48\hsize}
    \centering
    \includegraphics[width=\linewidth]{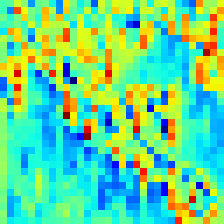}
  \end{minipage}
  \\ \vspace{1.5ex}
    \begin{minipage}[h]{0.48\hsize}
    \centering
    \includegraphics[width=\linewidth]{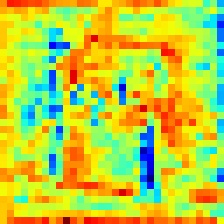}
  \end{minipage}
    \begin{minipage}[h]{0.48\hsize}
    \centering
    \includegraphics[width=\linewidth]{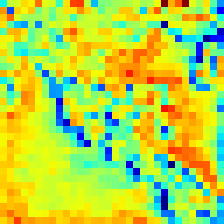}
  \end{minipage}
  \\ \vspace{1.5ex}
      \begin{minipage}[h]{0.48\hsize}
    \centering
    \includegraphics[width=\linewidth]{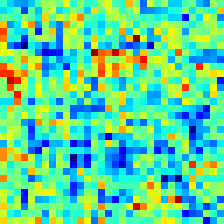}
  \end{minipage}
      \begin{minipage}[h]{0.48\hsize}
    \centering
    \includegraphics[width=\linewidth]{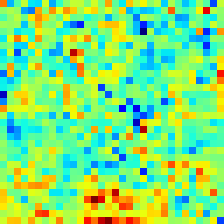}
  \end{minipage}
  \\ \vspace{1.5ex}
  \begin{minipage}[h]{0.48\hsize}
    \centering
    \includegraphics[width=\linewidth]{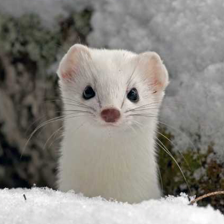}
  \end{minipage}
  \begin{minipage}[h]{0.48\hsize}
    \centering
    \includegraphics[width=\linewidth]{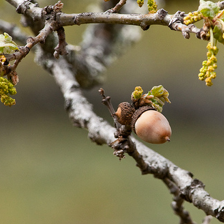}
  \end{minipage} \vspace{0.2ex}
  \caption{Part of states of the last BiLSTM2D layer in the Sequencer block of stage 1. From top to bottom: outputs of ver-LSTM, hor-LSTM, and ch-fusions and original images.}
  \label{figure:viz_lstm}
\end{minipage}
\end{figure}

One of the attractive properties of Sequencer is its flexible adaptability to the resolution, with minimal impact on accuracy even when the resolution of the input image is varied from one-half to twice. In comparison, architectures like MLP-Mixer~\cite{tolstikhin2021mlp} have a fixed input resolution, and GFNet~\cite{rao2021global} requires interpolation of weights in the Fourier domain when inputting images with a resolution different from training images. We evaluate the resolution adaptability of models comparatively by inputting different resolution images to each model, without fine-tuning, with pre-trained weights on IN-1K at the resolution of 224$^2$. Figure \ref{figure:resolution_adaptability_org} compares absolute top-1 accuracy on IN-1K, and Figure \ref{figure:resolution_adaptability_delta} compares relative one to the input image with the resolution of 224$^2$. By increasing the resolution by 28 for Sequencer2D-S and by 32 for other models, we avoid padding and prevent the effect of padding on accuracy. Compared to DeiT-S~\cite{touvron2020training}, GFNet-S~\cite{rao2021global}, CycleMLP-B2~\cite{chen2022cyclemlp}, and ConvNeXt-T~\cite{liu2022convnet}, Sequencer-S's performance is more sustainable. The relative accuracy is consistently better than ConvNeXt~\cite{liu2022convnet}, which is influential in the lower-resolution band, and, at 448$^2$ resolution, 0.6\% higher than CycleMLP~\cite{chen2022cyclemlp}, which is influential in the double-resolution band. It is noteworthy that Sequencer continues to maintain high accuracy on double resolution.

The higher the input resolution, the higher memory-efficiency and throughput of Sequencers when compared to DeiT~\cite{touvron2020training}. Figure \ref{figure:resolution_adaptability_efficiency} shows the efficiency of Sequencer2D-S when compared to DeiT-S and ConvNeXt-T~\cite{liu2022convnet}. Memory consumption increases rapidly in DeiT-S and ConvNeXt-T with increasing input resolution, but more gradual increase in Sequencer2D-S. The result strongly implies that it has more practical potential as the resolution increases than the ViTs. At a resolution of 224$^2$, it is behind DeiT in throughput, but it stands ahead of DeiT when images with a resolution of $896^2$ are input.

\begin{figure}[htb]
  \centering
  \begin{minipage}[b]{0.14\hsize}
    \centering
    \includegraphics[width=\linewidth]{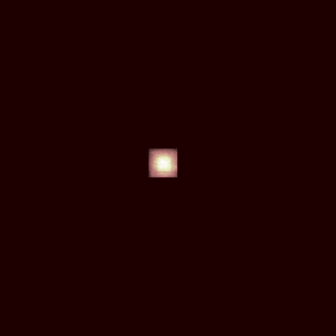}
    \subcaption{RN50/first}
    \label{figure:erf_resnets}
  \end{minipage}
  \begin{minipage}[b]{0.14\hsize}
    \centering
    \includegraphics[width=\linewidth]{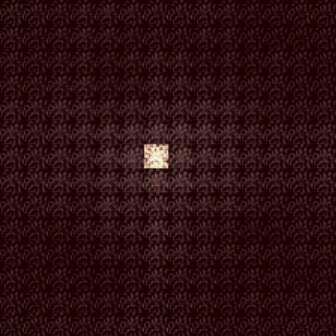}
    \subcaption{DeiTS/first}
    \label{figure:erf_deits}
  \end{minipage}
    \begin{minipage}[b]{0.14\hsize}
    \centering
    \includegraphics[width=\linewidth]{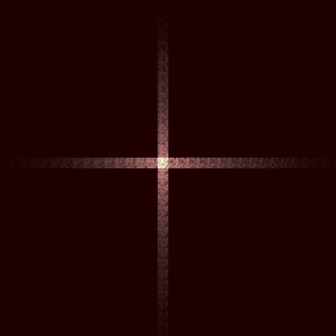}
    \subcaption{SeqS/first}
    \label{figure:erf_seqs}
  \end{minipage}
    \begin{minipage}[b]{0.14\hsize}
    \centering
    \includegraphics[width=\linewidth]{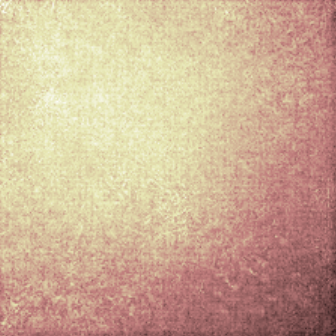}
    \subcaption{RN50/last}
    \label{figure:erf_resnetd}
  \end{minipage}
    \begin{minipage}[b]{0.14\hsize}
    \centering
    \includegraphics[width=\linewidth]{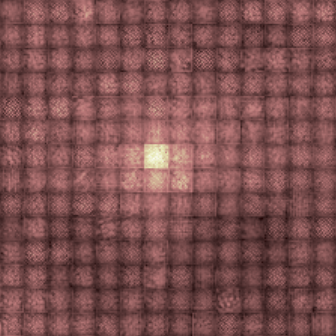}
    \subcaption{DeiTS/last}
    \label{figure:erf_deitd}
  \end{minipage}
    \begin{minipage}[b]{0.14\hsize}
    \centering
    \includegraphics[width=\linewidth]{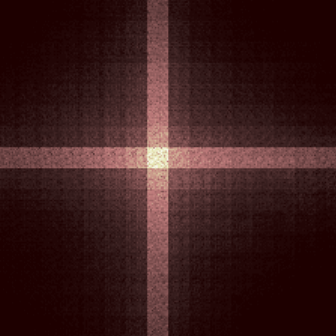}
    \subcaption{SeqS/last}
    \label{figure:erf_seqd}
  \end{minipage}
  \caption{The visualizations are the ERFs of Sequencer2D-S and comparative models such as ResNet-50 and DeiT-S. The left of the slash denotes the model name, and the right of the slash denotes the location of the block of output used to generate the ERFs. The ERFs are rescaled from 0 to 1. The brighter and more influential the region is, the closer to 1, and the darker, the closer to 0.}
    \label{figure:erf_main}
\end{figure}

In general, CNNs have localized, layer-by-layer expanding receptive fields, and ViTs without shifted windows capture global dependencies, working the self-attention mechanism. In contrast, in the case of Sequencer, it is not clear how information is processed in Sequencer block. We calculated ERF~\cite{luo2016understanding} for ResNet-50~\cite{he2016deep}, DeiT-S~\cite{touvron2020training}, and Sequencer2D-S as shown in Figure \ref{figure:erf_main}. ERFs of Sequencer2D-S form a cruciform shape in all layers. The trend distinguishes it from well-known models such as DeiT-S and ResNet-50. More remarkably, in shallow layers, Sequencer2D-S has a wider ERF than ResNet-50, although not as wide as DeiT. This observation confirms that LSTMs in Sequencer can model long-term dependencies as expected and that Sequencer recognizes sufficiently long vertical or horizontal regions. Thus, it can be argued that Sequencer recognizes an image in a very different way than CNNs or ViTs. For more details on ERF and additional visualization, see Appendix \ref{sec:erf_more}.

Moreover we also visualized a hidden state of vertical and horizontal BiLSTM, and a feature map after channel fusion, and the results are visualized in Figure \ref{figure:viz_lstm}. It demonstrates that our Sequencer has the hidden states interact with each other over the vertical and horizontal directions. The closer tokens are in position, the stronger their interaction tends to be; the farther tokens are in position, the less their interaction tends to be.

\section{Conclusions}
\label{sec:conclusions}
We propose a novel and simple architecture that leverages LSTM for computer vision. It is demonstrated that new modeling with LSTM instead of the self-attention layer can achieve competitive performance with current state-of-the-art models. Our experiments show that Sequencer has a good memory-resource/accuracy and parameter/accuracy tradeoffs, comparable to the main existing methods. Despite the impact of recursion on throughput, we have demonstrated benefits over it. We believe that these results raise a number of interesting issues. Improving Sequencer's poor throughput is one example. Moreover, we expect that investigating the internal mechanisms of our model using methods other than ERF will further our understanding of how this architecture works. In addition, it would be important to analyze in more detail the features learned by Sequencer in comparison to other architectures. We hope this will lead to a better understanding of the role of various inductive biases in computer vision. Furthermore, we expect that our results trigger further study beyond the domain or research area. Especially, it would be a very interesting open question to see if such a design works with time-series data in vision such as video or in a multi-modal problem setting combined with another modality such as video with audio.

\begin{ack}
Our colleagues at AnyTech Co., Ltd. provided valuable comments on the early versions and encouragement. We thank them for their cooperation. In particular, We thank Atsushi Fukuda for organizing discussion opportunities. We also thank people who support us, belonging to Graduate School of Artificial Intelligence and Science, Rikkyo University.
\end{ack}

{
\small
\bibliographystyle{plain}
\bibliography{neurips_2022}
}


\clearpage
\appendix

\section{Societal impact}
\label{sec:societal_impact}
The impact of this study on society has both positive and negative aspects. Here we discuss each.

On the positive side, our proposal would promote modeling methods using LSTMs in computer vision. This study takes image patches as tokens and models their relationships with LSTMs. Although LSTMs have been used in computer vision, designing image recognition with a module that includes LSTMs in the spatial direction as the main elements, as our study does, is new. It is exciting to see if this design benefits computer vision tasks other than image classification. Thus, our study would be an impetus for further research on its application to various computer vision tasks.

On the other side, our architecture may increase the carbon dioxide footprint: the study of new architectures for vision, such as Sequencer, requires iterative training of models for long periods to optimize the model's design. In particular, Sequencer is not a FLOPs-friendly design, and the amount of carbon dioxide emitted during training is likely to be high. Therefore, considering the environmental burden caused by the training of Sequencers, research to reduce the computational cost of Sequencers is also desired by society.

\section{Implementation details}
\label{sec:impl_details}
In this section, implementation details are supplemented. We describe the pseudocode of the BiLSTM2D layer, the architecture details, settings for training on IN-1K, and introduce settings for transfer learning.

\subsection{Pseudocode}
\label{subsec:pseudocode}
\begin{algorithm}[h]
\caption{Pseudocode of BiLSTM2D layer.}
\label{alg:lstm2d}
\lstset{
  backgroundcolor=\color{white},
  basicstyle=\fontsize{7.2pt}{7.2pt}\ttfamily\selectfont,
  columns=fullflexible,
  breaklines=true,
  captionpos=b,
  commentstyle=\fontsize{7.2pt}{7.2pt}\color{Primary},
  keywordstyle=\fontsize{7.2pt}{7.2pt},
}
\begin{lstlisting}[language=python]
# B: batch size H: height, W: width, C: channel, D: hidden dimension
# x: input tensor of shape (B, H, W, C)
### initialization ###
self.rnn_v = nn.LSTM(C, D, num_layers=1, batch_first=True, bias=True, bidirectional=True)
self.rnn_h = nn.LSTM(C, D, num_layers=1, batch_first=True, bias=True, bidirectional=True)
self.fc = nn.Linear(4 * D, C)
### forward ###
def forward(self, x):
    v, _ = self.rnn_v(x.permute(0, 2, 1, 3).reshape(-1, H, C))
    v = v.reshape(B, W, H, -1).permute(0, 2, 1, 3)
    h, _ = self.rnn_h(x.reshape(-1, W, C))
    h = h.reshape(B, H, W, -1)
    x = torch.cat([v, h], dim=-1)
    x = self.fc(x)
    return x
\end{lstlisting}
\end{algorithm}

\subsection{Architecture details}
\label{subsec:architecture_details}

This subsection describes Sequencer's architecture. The architectural details are shown in Table \ref{table:model_family} and \ref{table:vanila_family}.

Sequencer2D-S is based on a ViP-S/7-like architecture. We intend to directly compare the BiLSTM2D layer in Sequencer2D, which has a similar structure, with the Permute-MLP layer in ViP-S/7. Table \ref{table:model_family} is a summary of the architecture. In keeping with ViP, the first stage of Sequencers involves patch embedding with a 7x7 kernel. The second stage of Sequencers performs patch embedding with a 2x2 kernel, but the following two stages have no downsampling. The classifier of Sequencers then continues with layer normalization (LN)~\cite{ba2016layer}, followed by global average pooling and a linear layer. The number of blocks of Sequencer2D-S, Sequencer2D-M, and Sequencer2D-L correspond to ViP-S/7, ViP-M/7, and ViP-L/7, respectively. However, as described in Appendix \ref{sec:more_results}, we configure the dimension of the block to be different from ViP-M/7 and ViP-L/7 for Sequencer2D-M and Sequencer2D-L, respectively, because high dimension causes over-fitting.

VSequencer is a bit different from Sequencer2D in that it is non-hierarchical architecture. Table \ref{table:vanila_family} define that no downsampling is performed in the second stage, instead of downsampling with 14x14 kernel for patch embedding in the first stage. In addition, we match the dimension of the blocks in the first stage to the dimension of the subsequent blocks.

Following the overall architecture, we describe the details of the modules not mentioned in the main text. Sequencer2D block and the Vanilla Sequencer block use LNs~\cite{ba2016layer} for the normalization layers. We follow previous studies for the channel MLPs of these blocks and employ MLPs with Gaussian Error Linear Units (GELUs)~\cite{hendrycks2016gaussian} as the activation function; the ratio of increasing dimension in MLPs is uniformly 3x, as shown in Table \ref{table:model_family} and \ref{table:vanila_family}.

\subsection{IN-1K settings}
\label{subsec:in1k_settings}
On IN-1K dataset \cite{krizhevsky2012imagenet}, we utilize the hyper-parameters displayed in Table \ref{table:in_setting} to scratch train models in subsections \ref{subsec:in1k} and \ref{subsec:ablation_studies}. All Sequencer variants, including the models in the ablation study, follow almost the same settings for pre-training. However, the stochastic depth rate and batch size are adjusted depending on the model variant. The models in the ablation study are Sequencer2D-S based because of the following Sequencer2D-S settings.

The fine-tuning Sequencer2D-L$\uparrow$ 392$^2$ in subsection \ref{subsec:ft_in1k} has slightly different hyper-parameters than the pre-training models. There are changes in the settings for the number of epochs and learning rate because it uses trained weights, so there is no need to increase these hyper-parameters. In addition, we used crop ratio 0.875 during testing in the pre-training models instead of crop ratio 1.0 in the fine-tuning model.

\subsection{Transfer learning settings}
\label{subsec:tfl_settings}
Details of the datasets used for transfer learning in subsection \ref{subsec:transfer_learning} are shown in Table \ref{table:transfer_dataset}. This summary includes for each dataset CIFAR-10~\cite{krizhevsky2009learning}, CIFAR-100~\cite{krizhevsky2009learning}, Flowers-102~\cite{nilsback2008automated}, and Stanford Cars~\cite{krause20133d}, the number of training images, test images, and number of categories are listed.

Table \ref{table:in_setting} demonstrates the hyperparameters used in transfer learning with these datasets. The training epochs are especially adjusting to the datasets and changing them. The reason for this is attributable to the different sizes of the datasets.

\begin{table}[tb]
\centering
\caption{\textbf{Variants of Sequencer2D and details}. "d" denotes the input/output dimension, and $D$ denotes the hidden dimension as above. "$\downarrow n$" (e.g., $\downarrow$2) shows the stride of the downsampling is $n$}
\small
\setlength\tabcolsep{3.54pt}
\begingroup
\renewcommand{\arraystretch}{1.5}
\begin{NiceTabular}{clll}[colortbl-like]
    & \multicolumn{1}{c}{Sequencer2D-S} & \multicolumn{1}{c}{Sequencer2D-M} & \multicolumn{1}{c}{Sequencer2D-L} \\
\midrule
    \multirow{4}{*}{stage 1} & \multicolumn{1}{c}{Patch Embedding$\downarrow$7} & \multicolumn{1}{c}{Patch Embedding$\downarrow$7} & \multicolumn{1}{c}{Patch Embedding$\downarrow$7} \\
    & \blocka{192}{48}{3}{4} & \blocka{192}{48}{3}{4} & \blocka{192}{48}{3}{8} \\
    &&& \\
    &&& \\
\midrule
    \multirow{4}{*}{stage 2} & \multicolumn{1}{c}{Patch Embedding$\downarrow$2} & \multicolumn{1}{c}{Patch Embedding$\downarrow$2} & \multicolumn{1}{c}{Patch Embedding$\downarrow$2} \\
    & \blocka{384}{96}{3}{3} & \blocka{384}{96}{3}{3} & \blocka{384}{96}{3}{8} \\
    &&& \\
    &&& \\
\midrule
    \multirow{4}{*}{stage 3} & \multicolumn{1}{c}{Point-wise Linear} & \multicolumn{1}{c}{Point-wise Linear} & \multicolumn{1}{c}{Point-wise Linear} \\ 
    & \blocka{384}{96}{3}{8} & \blocka{384}{96}{3}{14} & \blocka{384}{96}{3}{16} \\
    &&& \\
    &&& \\
\midrule
    \multirow{4}{*}{stage 4} & \multicolumn{1}{c}{Point-wise Linear} & \multicolumn{1}{c}{Point-wise Linear} & \multicolumn{1}{c}{Point-wise Linear} \\ 
    & \blocka{384}{96}{3}{3} & \blocka{384}{96}{3}{3} & \blocka{384}{96}{3}{4} \\
    &&& \\
    &&& \\
\midrule
classifier & \multicolumn{3}{c}{Layer Norm., Global Average Pooling, Linear}\\
\bottomrule
\end{NiceTabular}
\endgroup
\label{table:model_family}
\end{table}

\begin{table}[tb]
\centering
\caption{\textbf{Variants of VSequencer and details}. "d" denotes the input/output dimension, and $D$ denotes the hidden dimension as above. "$\downarrow n$" (e.g., $\downarrow$2) shows the stride of the downsampling is $n$}
\small
\setlength\tabcolsep{3.54pt}
\begingroup
\renewcommand{\arraystretch}{1.5}
\begin{NiceTabular}{clll}[colortbl-like]
    & \multicolumn{1}{c}{VSequencer-S} & \multicolumn{1}{c}{VSequencer-S(H)} & \multicolumn{1}{c}{VSequencer-S(PE)} \\
\midrule
    \multirow{4}{*}{stage 1} & \multicolumn{1}{c}{Patch Embedding$\downarrow$14} & \multicolumn{1}{c}{Patch Embedding$\downarrow$7} & \multicolumn{1}{c}{Patch Embedding$\downarrow$14} \\
    & \blockb{384}{192}{3}{4} & \blockb{192}{96}{3}{4} & \blockb{384}{192}{3}{4} \\
    &&& \\
    &&& \\
\midrule
    \multirow{4}{*}{stage 2} & \multicolumn{1}{c}{Point-wise Linear} & \multicolumn{1}{c}{Patch Embedding$\downarrow$2} & \multicolumn{1}{c}{Point-wise Linear} \\
    & \blockb{384}{192}{3}{3} & \blockb{384}{192}{3}{3} & \blockb{384}{192}{3}{3} \\
    &&& \\
    &&& \\
\midrule
    \multirow{4}{*}{stage 3} & \multicolumn{1}{c}{Point-wise Linear} & \multicolumn{1}{c}{Point-wise Linear} & \multicolumn{1}{c}{Point-wise Linear} \\ 
    & \blockb{384}{192}{3}{8} & \blockb{384}{192}{3}{8} & \blockb{384}{192}{3}{8} \\
    &&& \\
    &&& \\
\midrule
    \multirow{4}{*}{stage 4} & \multicolumn{1}{c}{Point-wise Linear} & \multicolumn{1}{c}{Point-wise Linear} & \multicolumn{1}{c}{Point-wise Linear} \\ 
    & \blockb{384}{192}{3}{3} & \blockb{384}{192}{3}{3} & \blockb{384}{192}{3}{3} \\
    &&& \\
    &&& \\
\midrule
classifier & \multicolumn{3}{c}{Layer Norm., Global Average Pooling, Linear}\\
\bottomrule
\end{NiceTabular}
\endgroup
\label{table:vanila_family}
\end{table}

\begin{table}[tb]
\centering
\caption{\textbf{Hyper-parameters}. $\uparrow$ denotes fine-tuning pre-trained model on IN-1K.  Multiple values are for each model, respectively.}
\small
\begin{NiceTabular}{@{\hskip -0.05ex}l|c@{\hskip 1ex}c@{\hskip 1ex}c}[colortbl-like]
\multirow{2}{*}{Training config.} & Sequencer2D-S/M/L & Sequencer2D-L$\uparrow$ & Sequencer2D-S$\uparrow$/M$\uparrow$/L$\uparrow$ \\
& 224$^2$ & 392$^2$ & 224$^2$ \\
\hline
dataset & IN-1K~\cite{krizhevsky2012imagenet} & IN-1K~\cite{krizhevsky2012imagenet} & CIFAR$^{10,\, 100}$, Flowers, Cars \\
optimizer & AdamW~\cite{loshchilov2017decoupled} & AdamW~\cite{loshchilov2017decoupled} & AdamW~\cite{loshchilov2017decoupled}\\
base learning rate & 2e-3/1.5e-3/1e-3 & 5e-5 & 1e-4 \\
weight decay & 0.05 & 1e-8 & 1e-4 \\
optimizer $\epsilon$ & 1e-8 & 1e-8 & 1e-8 \\
optimizer momentum & $\beta_1,=0.9, \beta_2{=}0.999$ & $\beta_1,=0.9, \beta_2{=}0.999$ & $\beta_1,=0.9, \beta_2{=}0.999$ \\
batch size & 2048/1536/1024 & 512 & 512 \\
training epochs & 300 & 30 & CIFAR: 200, Others: 1000 \\
learning rate schedule & cosine decay & cosine decay & cosine decay \\
lower learning rate bound & 1e-6 & 1e-6 & 1e-6 \\
warmup epochs & 20 & None & 5 \\
warmup schedule & linear & None & linear \\
warmup learning rate & 1e-6 & None & 1e-6 \\
cooldown epochs & 10 & None & 10 \\
crop ratio & 0.875 & 1.0 & 0.875 \\
randaugment~\cite{cubuk2020randaugment} & (9, 0.5) & (9, 0.5) & (9, 0.5) \\
mixup $\alpha$~\cite{zhang2017mixup} & 0.8 & 0.8 & 0.8 \\
cutmix $\alpha$~\cite{yun2019cutmix} & 1.0 & 1.0 & 1.0 \\
random erasing~\cite{zhong2020random} & 0.25 & 0.25 & None \\
label smoothing~\cite{szegedy2016rethinking} & 0.1 & 0.1 & 0.1 \\
stochastic depth~\cite{huang2016deep} & 0.1/0.2/0.4 & 0.4 & 0.1/0.2/0.4 \\
gradient clip & None & None & 1 \\
\bottomrule
\end{NiceTabular}
\label{table:in_setting}
\end{table}

\begin{table}[t]
  \centering
  \caption{\textbf{Transfer learning datasets}.} 
  \small
\begin{NiceTabular}{l|rrr}[colortbl-like]
Dataset & Train Size & Test size & \#Classes \\\hline
   CIFAR-10~\cite{krizhevsky2009learning} & 50,000 & 10,000 & 10 \\
   CIFAR-100~\cite{krizhevsky2009learning} & 50,000 & 10,000 & 100 \\
   Flowers-102~\cite{nilsback2008automated} & 2,040 & 6,149 & 102 \\ 
   Stanford Cars~\cite{krause20133d} & 8,144 & 8,041 & 196 \\ \bottomrule
\end{NiceTabular}
\label{table:transfer_dataset}
\end{table}

\section{More results}
\label{sec:more_results}
This section discusses additional results that could not be addressed in the main text. The contents of the experiment consist of three parts: an evaluation of robustness in subsection \ref{subsec:rob}, an evaluation of generalization performance in subsection \ref{subsec:gen}, and a discussion of over-fitting in subsection \ref{subsec:overfitting}.

\subsection{Robustness}
\label{subsec:rob}
In this subsection, we evaluate the robustness of Sequencer. There are two main evaluation methods, benchmark datasets and adversarial attacks.

 Evaluation with benchmark datasets reveals nice robustness of Sequencer. The evaluation results are summarized in Table \ref{table:robustress_generalization}. We test our models, trained on only IN-1K, on several datasets such as ImageNet-A/R/Sketch/C (IN-A/R/Sketch/C)~\cite{hendrycks2021natural, hendrycks2021many, wang2019learning, hendrycks2018benchmarking} to evaluate robustness. We evaluate our models on IN-C with mean corruption error (mCE), and on other datasets with top-1 accuracy. This result leads us to suggest that for models with a similar number of parameters, Sequencer is conquered by Swin and is robust enough to be competitive with ConvNeXt. Table \ref{table:robustress_details} shows detail evaluation on IN-C. According to the results, it is understood that Sequencer is more immune to corruptions other than Noise than Swin and ConvNeXt, and, in particular, the model is less sensitive to weather conditions.
 
Sequencers are tolerant of principal adversarial attacks. We evaluate robustness using the single-step attack algorithm FGSM~\cite{goodfellow2014explaining} and multi-step attack algorithm PGD~\cite{madry2017towards}. Both algorithms give a perturbation of max magnitude $1$. For PGD, we choose steps $5$ and step size $0.5$. This setup is based on RVT~\cite{mao2021towards}. Table \ref{table:robustress_details} indicates that Sequencer2D-L defeats in both FGSM and PGD compared to other models. Thus, Sequencer has an advantage over conventional models, such as RVT, which tout robustness on these adversarial attacks.

\subsection{Generalization ability}
\label{subsec:gen}
The generalization ability of Sequencers is also impressive. We evaluate our models on ImageNet-Real/V2 (IN-Real/V2)~\cite{beyer2020we, recht2019imagenet} to test their generalization performance: IN-Real is a re-labeled dataset of the IN-1K validation set, and IN-V2 is the dataset that re-collects the IN-1K validation set. Table \ref{table:robustress_generalization} shows the results of evaluating the top-1 accuracy on both datasets. We reveal an understanding of the Sequencer's excellent generalization ability.

\begin{table}[tb]
\centering
\caption{\textbf{The robustness} is evaluated on IN-A~\cite{hendrycks2021natural} (top-1 accuracy), IN-R~\cite{hendrycks2021many} (top-1 accuracy), IN-Sketch~\cite{wang2019learning} (top-1 accuracy), IN-C~\cite{hendrycks2018benchmarking} (mCE), FGSM~\cite{goodfellow2014explaining}  (top-1 accuracy), and PGD~\cite{madry2017towards}  (top-1 accuracy). \textbf{The generalization ability} is evaluated on IN-Real~\cite{beyer2020we} and IN-V2~\cite{recht2019imagenet}. We denote the higher as better value as $\uparrow$ and the lower as better value as $\downarrow$. Rather than those reported in the original paper, the values we observed are marked with $^\dagger$. If the model name has $^\dagger$, it means that we observed all the metrics of the model.}
\small
\setlength\tabcolsep{2.36pt}%
\begin{NiceTabular}{lrrc|cccc|cc|cc}[colortbl-like]
Model         & \multicolumn{1}{c}{\#Param.} & \multicolumn{1}{c}{FLOPs} & Clean($\uparrow$) & A($\uparrow$)   & R($\uparrow$)    & Sk.($\uparrow$)  & C($\downarrow$) & FGSM($\uparrow$) & PGD($\uparrow$) & Real($\uparrow$) & V2($\uparrow$)    \\
\hline
Swin-T~\cite{liu2021swin} & 28M                          & 4.5G                      & 81.2  & 21.6 & 41.3 & 29.1 & 62.0 & 33.7 & 7.3 & 86.7$^\dagger$ & 69.6$^\dagger$ \\
ConvNeXt-T~\cite{liu2022convnet} & 29M                          & 4.5G                      & 82.1  & 24.2 & 47.2 & 33.8 & 53.2 & 37.8$^\dagger$ & 10.5$^\dagger$ & 87.3$^\dagger$ & 71.0$^\dagger$ \\
RVT-S*~\cite{mao2021towards} & 23M & 4.7G & 81.9 & 25.7 & 47.7 & 34.7 & 49.4 & 51.8 & 28.2 & - & - \\
\rowcolor{Primary100}Sequencer2D-S & 28M                          & 8.4G                      & 82.3 & 26.7 & 45.1 & 33.4 & 53.0 & 49.2 & 25.0 & 87.4 & 71.8 \\
\rowcolor{Primary100}Sequencer2D-M & 38M                          & 11.1G                     & 82.8 & 30.5 & 46.3 & 34.7 & 51.8 & 50.8 & 26.3 & 87.6 & 72.5 \\
Swin-S~\cite{liu2021swin}$^\dagger$ & 50M                          & 8.7G                     & 83.2  & 32.5 & 45.2 & 32.3 & 54.9 & 45.9 & 18.1 & 87.7 & 72.1 \\
ConvNeXt-S$^\dagger$~\cite{liu2022convnet} & 50M                          & 8.7G                     & 83.1  & 31.3 & \textbf{49.6} & \textbf{37.1} & 49.5 & 46.1 & 17.7 & \textbf{88.1}  & 72.5 \\
\rowcolor{Primary100}Sequencer2D-L & 54M                          & 16.6G                     & \textbf{83.4}  & \textbf{35.5} & 48.1 & 35.8 & \textbf{48.9} & \textbf{53.1} & \textbf{30.9} & 87.9 & \textbf{73.4} \\
\hline
Swin-B~\cite{liu2021swin} & 88M                          & 15.4G                     & 83.4  & 35.8 & 46.6 & 32.4 & 54.4 & 49.2 & 21.3 & 89.2$^\dagger$ & 75.6$^\dagger$ \\
ConvNeXt-B~\cite{liu2022convnet} & 89M                          & 15.4G                     & 83.8  & 36.7 & 51.3 & 38.2 & 46.8 & 47.5$^\dagger$ & 18.3$^\dagger$ & 88.4$^\dagger$ & 73.7$^\dagger$ \\
RVT-B*~\cite{mao2021towards} & 92M & 17.7G & 82.6 & 28.5 & 48.7 & 36.0 & 46.8 & 53.0 & 29.9 & - & - \\
\bottomrule
\end{NiceTabular}
\label{table:robustress_generalization}
\end{table}

\begin{table}[tb]
\caption{\text{Details of robustness evaluation with IN-C}.}

\small
\setlength\tabcolsep{1.18pt}%
\begin{NiceTabular}{l|c|ccc|cccc|cccc|cccc}[colortbl-like]
\multicolumn{2}{c}{} & \multicolumn{3}{c}{Noise} & \multicolumn{4}{c}{Blur} & \multicolumn{4}{c}{Weather} & \multicolumn{4}{c}{Digital} \\
\toprule
\multicolumn{1}{c|}{Model}         & mCE  & \scriptsize{Gauss.} & \scriptsize{Shot} & \scriptsize{Impulse} & \scriptsize{Defocus} &\scriptsize
{Glass} & \scriptsize{Motion} & \scriptsize{Zoom} & \scriptsize{Snow} & \scriptsize{Frost} & \scriptsize{Fog}  & \scriptsize{Bright} & \scriptsize{Contrast} & \scriptsize{Elastic} & \scriptsize{Pixel} & \scriptsize{JPEG} \\ \hline
Swin-S~\cite{liu2021swin}       & 54.9 & 42.9   & 44.9 & 43.2    & 61.3    & 74.1  & 56.6   & 67.5 & 50.8 & 48.5  & 46.0 & 44.1   & 42.1     & 68.9    & 62.1  & 70.7 \\
ConvNeXt-S~\cite{liu2022convnet}   & 49.5 & \textbf{38.1}   & \textbf{39.1} & \textbf{37.9}    & 57.8    & 72.5  & \textbf{51.8}   & \textbf{61.9} & 46.1 & 43.8  & 44.6 & 39.6   & 37.6     & 66.7    & 55.1  & \textbf{50.1} \\
\rowcolor{Primary100}Sequencer2D-L & \textbf{48.9} & 43.3   & 42.0 & 41.4    & \textbf{55.2}    & \textbf{71.0}  & \textbf{51.8}   & 63.3 & \textbf{44.2} & \textbf{41.0}  & \textbf{41.9} & \textbf{37.1}   & \textbf{33.8}     & \textbf{66.6}    & \textbf{50.4}  & 51.1 \\ \bottomrule
\end{NiceTabular}
\label{table:robustress_details}
\end{table}

\subsection{Over-fitting}
\label{subsec:overfitting}
\begin{figure}[tb]
  \centering
    \begin{minipage}[b]{0.48\hsize}
    \centering
    \includegraphics[width=\linewidth]{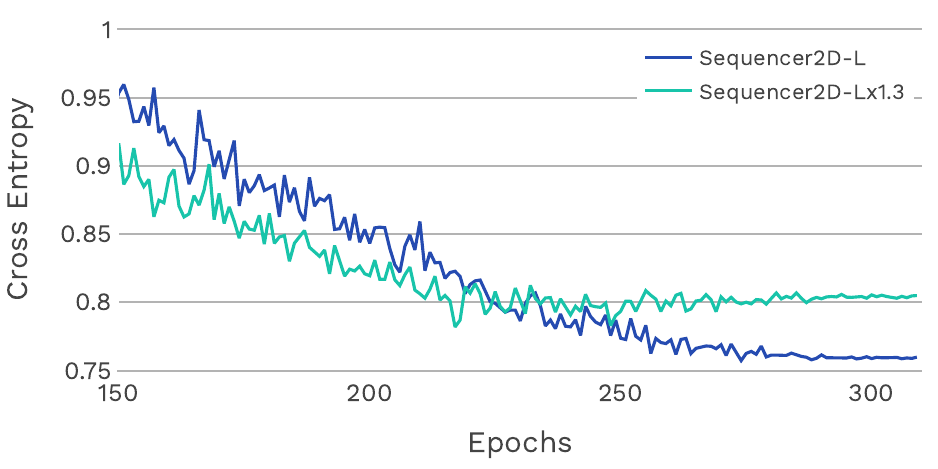}
    \subcaption{Cross entropy}
    \label{figure:overfitting_loss}
  \end{minipage}
  \begin{minipage}[b]{0.48\hsize}
    \centering
    \includegraphics[width=\linewidth]{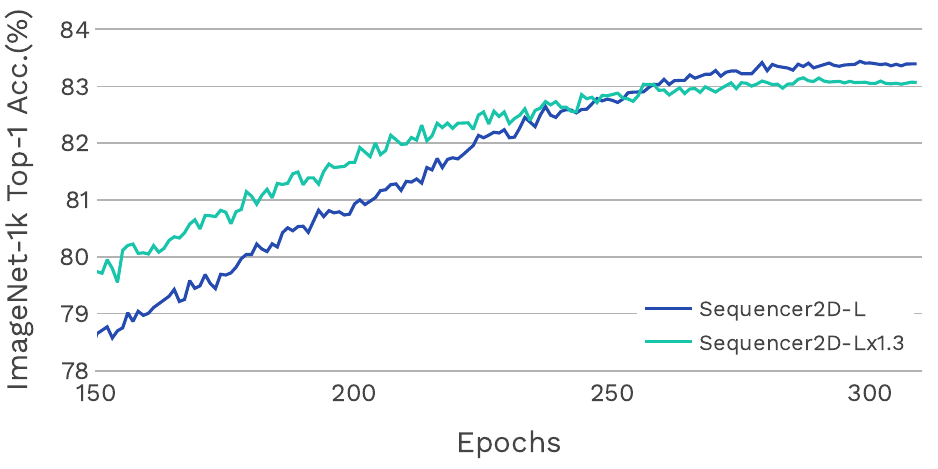}
    \subcaption{Top-1 accuracy}
    \label{figure:overfitting_top1}
  \end{minipage}
  \caption{\textbf{Comparison of different model widths}. (\subref{figure:overfitting_loss}) is cross entropy, (\subref{figure:overfitting_top1}) is top-1 accuracy comparison, on IN-1K validation set. \textbf{The blue curve} represents the original Sequencer2D-L, which did not produce any problems and is learning all the way through. In contrast, \textbf{the green curve} represents the wider Sequencer2D-Lx1.3. This model stalls in the second half and is somewhat degenerate.}
    \label{figure:overfitting}
\end{figure}

Wide Sequencers tend to be over-trained. We scratch-train Sequencer2D-Lx1.3, which has 4/3 times the dimension of each layer of Sequencer2D-L, on IN-1K. The training utilizes the same conditions as Sequencer2D-L. Consequently, as Table \ref{table:overfitting} shows, Sequencer2D-Lx1.3 has 0.8\% less accuracy than Sequencer2D-L. Figure \ref{figure:overfitting} illustrates the cross-entropy evolution and top-1 accuracy on IN-1K validation set for the two models. On the one hand, cross-entropy decreased on Sequencer2D-L in the last 100 epochs. On the other hand, Sequencer2D-Lx1.3 is increasing. Thus, widening Sequencer is counterproductive for training.

\begin{table}[tb]
\small
\centering
\caption{\textbf{Comparison of accuracy for different model widths}.}
\begin{NiceTabular}{cccc}[colortbl-like]
Model & \#Params. & FLOPs & Acc. \\
\hline
\rowcolor{Primary100}Sequencer2D-L & 54M & 16.6G & 83.4 \\
Sequencer2D-Lx1.3 & 96M & 29.4G & 83.0 \\
\bottomrule
\end{NiceTabular}
\label{table:overfitting}
\end{table}

\subsection{Semantic segmentation}
\label{subsec:segmentation}
We evaluate models with Sequencer as the backbone for a semantic segmentation task. We trained and evaluated on ADE20K dataset~\cite{zhou2017scene}, a well-known scene parsing benchmark. The dataset consists of the training set with about 20k images and the validation set with about 2k, covering 150 fine-grained semantic classes. We employed Sequencer as the backbone of SemanticFPN~\cite{kirillov2019panoptic} to train and evaluate semantic segmentation. The training adopts a batch size of 32 and AdamW~\cite{loshchilov2017decoupled} with the initial learning rate of 2e-4, decay in the polynomial decay schedule with a power of 0.9, and 40k iterations of training. These settings follow Metaformer~\cite{yu2021metaformer}. Table \ref{table:segmentation}  of the result indicates that Sequencer has the generalization for segmentation is comparable to other leading models.

\subsection{Object Detection}
\label{subsec:detection}
We evaluate Sequencer on COCO benchmark~\cite{lin2014microsoft}. The dataset consists of 118k training images and 5k validation images. Sequencer with ImageNet pre-trained weights is employed as the backbone of RetinaNet~\cite{lin2017focal}. Following ~\cite{lin2017focal}, we employ AdamW, batch size of 16, and 1$\times$ training schedule. Table \ref{table:detection} shows that Sequencer is not suited for existing standard object detection models such as RetinaNet. It shows no improvement trend for model scaling. It also struggles to detect small objects, making RNN-based object detection models an issue to consider in the future.

\begin{table}[!htp]\centering
\caption{Object detection results on COCO dataset~\cite{lin2014microsoft}}\label{table:detection}
\small
\setlength\tabcolsep{2.36pt}
\begin{NiceTabular}{lccccccc}[colortbl-like]
Backbone & Params (M) &AP &AP$_{50}$ &AP$_{75}$ &AP$_{S}$ &AP$_{M}$ &AP$_{L}$ \\ \hline
ResNet-18~\cite{he2016deep} &21.3 &31.8 &49.6 &33.6 &16.3 &34.3 &43.2 \\
PoolFormer-S12~\cite{yu2021metaformer} &21.7 &36.2 &56.2 &38.2 &20.8 &39.1 &48.0 \\
Sequencer2D-S &37.3 &33.6 &54.8 &34.8 &15.3 &37.5 &50.2 \\
ResNet-50~\cite{he2016deep} &37.7 &36.3 &55.3 &38.6 &19.3 &40.0 &48.8 \\
PoolFormer-S24~\cite{yu2021metaformer} &31.1 &38.9 &59.7 &41.3 &23.3 &42.1 &51.8 \\
Sequencer2D-M &47.9 &34.5 &55.5 &35.9 &15.0 &39.0 &51.6 \\
ResNet-101~\cite{he2016deep} &56.7 &38.5 &57.8 &41.2 &21.4 &42.6 &51.1 \\
PoolFormer-S36~\cite{yu2021metaformer} &40.6 &39.5 &60.5 &41.8 &22.5 &42.9 &52.4 \\
Sequencer2D-L &63.9 &35.0 &56.4 &36.5 &16.5 &39.6 &51.6 \\
\end{NiceTabular}
\end{table}

\subsection{More studies}
\label{subsec:more_studies}

\paragraph{Method of merge}
As shown in Figure \ref{figure:overall}, "\texttt{concatenate}" is used to merge the vertical BiLSTM and horizontal BiLSTM outputs but "\texttt{add}" can also be used. See Table \ref{table:method_of_merge} for the result of the experiment.

\begin{table}[tb]
\small
\centering
\caption{\textbf{More Sequencer ablation experiments}.}
\begin{minipage}[t]{0.45\linewidth}{
\subcaption{Method of merge\label{table:method_of_merge}}
\begin{center}
\begin{NiceTabular}{cccc}[colortbl-like]
Union & \#Params. & FLOPs & Acc. \\
\hline
\texttt{add} & 27M & 8.0G & 82.2 \\
\rowcolor{Primary100}\texttt{concatnate} & 28M & 8.4G & \textbf{82.3} \\
\end{NiceTabular}
\end{center}
}\end{minipage}
\label{table:more_studies}
\end{table}

\section{Effective receptive field}
\label{sec:erf_more}

This section covers in detail the effective receptive fields (ERFs)~\cite{luo2016understanding} used in the visualization in subsection \ref{subsec:analysis_visualization}. First, we explain how the visualized effective receptive fields are obtained. Second, we present other visualization results not addressed in the main text. The ERF's calculations in this paper are based on ~\cite{ding2022scaling}.

\subsection{Calculation of visualized ERFs}
\label{sub:calc_erf}

The ERF~\cite{luo2016understanding} is a technique for calculating the pixels that contribute to the center of a output feature maps of a neural network. Let $\mathbf{I} \in \mathbb{R}^{n\times h \times w \times c}$ be a input image collection and $\mathbf{O} \in \mathbb{R}^{n \times h^\prime \times w^\prime \times c^\prime}$ be the output feature map collection. The center of the output feature map can be expressed as $\mathbf{O}_{:, \lfloor h^\prime/2 \rfloor, \lfloor w^\prime/2 \rfloor,:}$, where $\lfloor \cdot \rfloor$ is the floor function. Each element of the derivative of $\mathbf{O}_{i, \lfloor h^\prime/2\rfloor ,\lfloor w^\prime/2 \rfloor,j}$ to $\mathbf{I}$, i.e., $\frac{\partial \left(\sum_{i,j}\mathbf{O}_{i,\lfloor h^\prime/2, \rfloor \lfloor w^\prime/2 \rfloor,j}\right)}{\partial \mathbf{I}}$, represents to what extent the center of the output feature map changes for each perturbation of each pixel in each input image. Adding these together for all images and channels, we can calculate the average pixel contribution for all input images, which can be activated with a Rectified Linear Unit (ReLU) to get the positively contributing pixel values $\mathbf{P} \in \mathbb{R}^{n\times h\times w\times c}$, defined by
\begin{equation}
    \mathbf{P} = \text{ReLU}\left(\frac{\partial \left(\sum_{i,j}\mathbf{O}_{i,\lfloor h^\prime/2, \rfloor \lfloor w^\prime/2 \rfloor,j}\right)}{\partial \mathbf{I}}\right) \,.
\end{equation}
Furthermore, the score $\mathrm{S}\in\mathbb{R}^{h\times w}$ is calculated by
\begin{equation}
\mathrm{S} = \log_{10} \left(\sum_{i,j}\mathbf{P}_{i,:,:,j} + 1\right) \,,
\end{equation}
and $\mathrm{S}$ is called the effective receptive field.

Next, define a visualized effective receptive field based on the effective receptive field. We want to compare the effective receptive fields across models. We, therefore, calculate the score $\mathrm{S}_{\rm model}$ for each model and rescale $\mathrm{S}_{\rm model}$ from $0$ to $1$ across the models. The tensor calculated in this way is called the visualized effective receptive field.

The derivatives used in these definitions are efficient if they take advantage of the auto-grad mechanism. Indeed, we also relied on the automatic auto-grad function on \texttt{PyTorch}~\cite{paszke2019pytorch} to calculate the effective receptive fields.

\subsection{More visualization of ERFs}
\label{sub:erf_viz}

We introduce additional visualization and concrete visualization method. We experiment with visualization using input images of two different resolutions.

We visualize the effective receptive fields of Sequencer2D-S and comparative models by using 224$^2$ resolution images. The method is applied to the following models for comparing: ResNet-50~\cite{he2016deep}, ConvNeXt-T~\cite{liu2022convnet}, CycleMLP-B2~\cite{chen2022cyclemlp}, DeiT-S~\cite{touvron2020training}, Swin-T~\cite{liu2021swin}, GFNet-S~\cite{rao2021global}, and ViP-S/7~\cite{hou2022vision}. The object to be visualized is the output for each block, and the effective receptive fields are calculated. For example, in the case of Sequencer2D-S, the effective receptive fields are calculated for the output of each Sequencer block. We are rescaling within a value between 0 and 1 for the whole to effective receptive fields for each model block.

The effective receptive fields of Sequencer2D-S and comparative models are then visualized using input images with a resolution of 448$^2$. The reason for running experiments is to verify how the receptive field is affected when the input resolution is increased compared to the 224$^2$ resolution input image. Sequencer2D-S compare with ResNet-50~\cite{he2016deep}, ConvNeXt-T~\cite{liu2022convnet}, CycleMLP-B2~\cite{chen2022cyclemlp}, DeiT-S~\cite{touvron2020training}, and GFNet-S~\cite{hou2022vision}. The method of visualization of the effective receptive field follows the case of input images with a resolution of 224$^2$.

Sequencer has very distinctive cruciform ERFs in all layers. Table \ref{figure:erf_sequencer}, \ref{figure:erf_resnet}, \ref{figure:erf_convnext}, \ref{figure:erf_cyclemlp}, \ref{figure:erf_deit}, \ref{figure:erf_swin}, \ref{figure:erf_gfnet}, and \ref{figure:erf_vip} illustrates this fact for 224$^2$ resolution input images. Furthermore, as shown in Table \ref{figure:erf_sequencer_448}, \ref{figure:erf_resnet_448}, \ref{figure:erf_convnext_448}, \ref{figure:erf_cyclemlp_448} and \ref{figure:erf_deit_448}, we observe the same trend when the double resolution. The ERFs are structurally quite different from the ERFs other than ViP, which have a similar structure. ViP's ERFs have, on average, some also coverage except for the cruciforms. In contrast, Sequencer's ERFs are limited to the cruciform and its neighborhood.

It is interesting to note that Sequencer, with its characteristic ERFs, achieves high accuracy. It will be helpful for future architecture development because of the possibility of creating Sequencer-like ERFs outside of LSTM.

\input{erf_sequencer_224}
\input{erf_resnet_224}
\input{erf_convnext_224}
\input{erf_cyclemlp_224}
\input{erf_deit_224}
\input{erf_swin_224}
\input{erf_gfnet_224}
\input{erf_vip_224}

\input{erf_sequencer_448}
\input{erf_resnet_448}
\input{erf_convnext_448}
\input{erf_cyclemlp_448}
\input{erf_deit_448}

\end{document}

%% file: erf_sequencer_224.tex
\begin{figure}[tb]
  \raggedright
  \begin{minipage}[t]{0.13\hsize}
    \centering
    \includegraphics[width=\linewidth]{figures/erf_results/lstm_sequencer2d_s_cat_224_blocks.0.0.pdf}
    \subcaption{Block 1}
    \label{figure:erf_sequencer:0}
  \end{minipage}
  \begin{minipage}[t]{0.13\hsize}
    \centering
    \includegraphics[width=\linewidth]{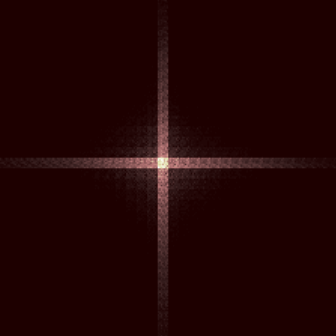}
    \subcaption{Block 2}
    \label{figure:erf_sequencer:1}
  \end{minipage}
  \begin{minipage}[t]{0.13\hsize}
    \centering
    \includegraphics[width=\linewidth]{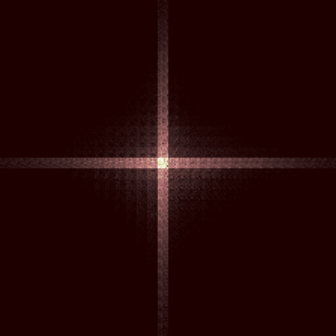}
    \subcaption{Block 3}
    \label{figure:erf_sequencer:2}
  \end{minipage}
  \begin{minipage}[t]{0.13\hsize}
    \centering
    \includegraphics[width=\linewidth]{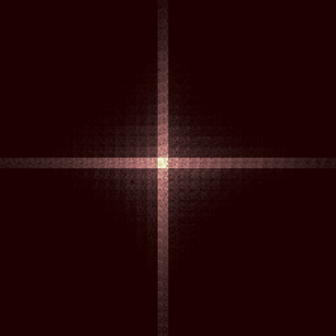}
    \subcaption{Block 4}
    \label{figure:erf_sequencer:3}
  \end{minipage}
  \begin{minipage}[t]{0.13\hsize}
    \centering
    \includegraphics[width=\linewidth]{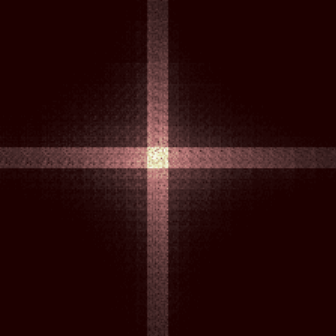}
    \subcaption{Block 5}
    \label{figure:erf_sequencer:4}
  \end{minipage}
  \begin{minipage}[t]{0.13\hsize}
    \centering
    \includegraphics[width=\linewidth]{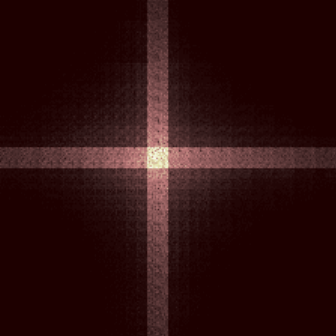}
    \subcaption{Block 6}
    \label{figure:erf_sequencer:5}
  \end{minipage}
  \begin{minipage}[t]{0.13\hsize}
    \centering
    \includegraphics[width=\linewidth]{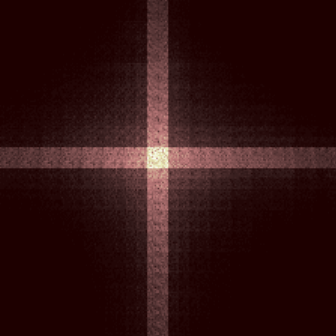}
    \subcaption{Block 7}
    \label{figure:erf_sequencer:6}
  \end{minipage}
  \begin{minipage}[t]{0.13\hsize}
    \centering
    \includegraphics[width=\linewidth]{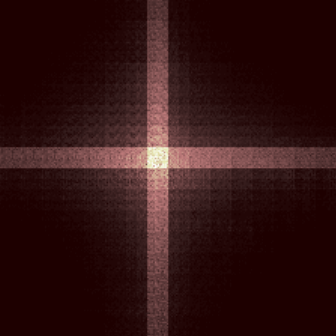}
    \subcaption{Block 8}
    \label{figure:erf_sequencer:7}
  \end{minipage}
  \begin{minipage}[t]{0.13\hsize}
    \centering
    \includegraphics[width=\linewidth]{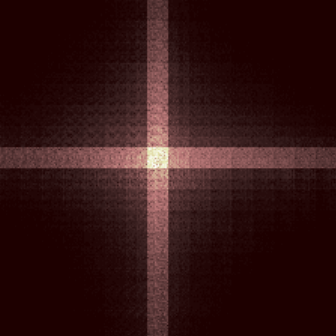}
    \subcaption{Block 9}
    \label{figure:erf_sequencer:8}
  \end{minipage}
  \begin{minipage}[t]{0.13\hsize}
    \centering
    \includegraphics[width=\linewidth]{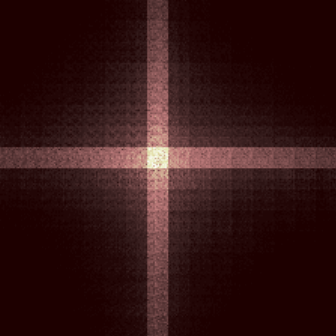}
    \subcaption{Block 10}
    \label{figure:erf_sequencer:9}
  \end{minipage}
  \begin{minipage}[t]{0.13\hsize}
    \centering
    \includegraphics[width=\linewidth]{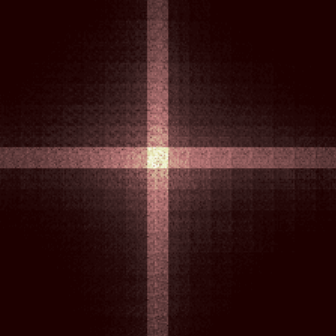}
    \subcaption{Block 11}
    \label{figure:erf_sequencer:10}
  \end{minipage}
  \begin{minipage}[t]{0.13\hsize}
    \centering
    \includegraphics[width=\linewidth]{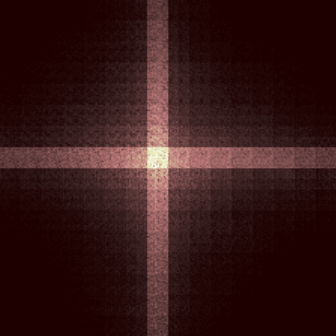}
    \subcaption{Block 12}
    \label{figure:erf_sequencer:11}
  \end{minipage}
  \begin{minipage}[t]{0.13\hsize}
    \centering
    \includegraphics[width=\linewidth]{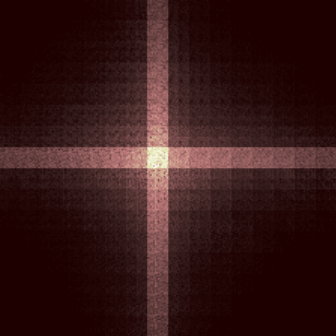}
    \subcaption{Block 13}
    \label{figure:erf_sequencer:12}
  \end{minipage}
  \begin{minipage}[t]{0.13\hsize}
    \centering
    \includegraphics[width=\linewidth]{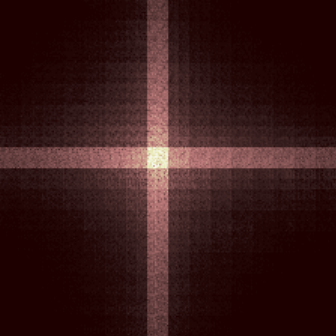}
    \subcaption{Block 14}
    \label{figure:erf_sequencer:13}
  \end{minipage}
  \begin{minipage}[t]{0.13\hsize}
    \centering
    \includegraphics[width=\linewidth]{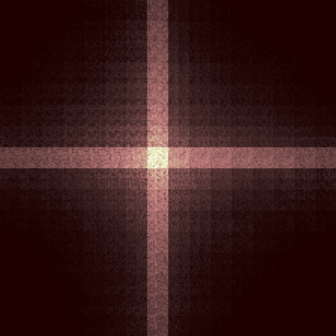}
    \subcaption{Block 15}
    \label{figure:erf_sequencer:14}
  \end{minipage}
  \begin{minipage}[t]{0.13\hsize}
    \centering
    \includegraphics[width=\linewidth]{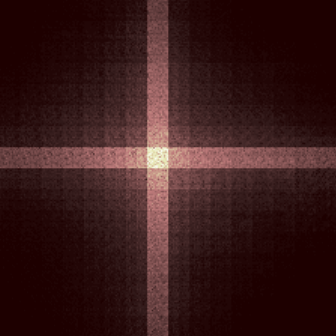}
    \subcaption{Block 16}
    \label{figure:erf_sequencer:15}
  \end{minipage}
  \begin{minipage}[t]{0.13\hsize}
    \centering
    \includegraphics[width=\linewidth]{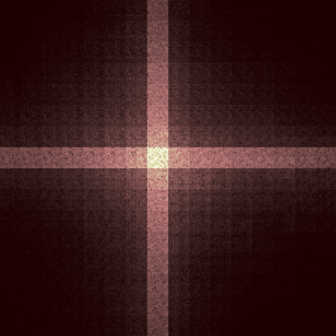}
    \subcaption{Block 17}
    \label{figure:erf_sequencer:16}
  \end{minipage}
  \begin{minipage}[t]{0.13\hsize}
    \centering
    \includegraphics[width=\linewidth]{figures/erf_results/lstm_sequencer2d_s_cat_224_blocks.3.2.pdf}
    \subcaption{Block 18}
    \label{figure:erf_sequencer:17}
  \end{minipage}
  \caption{ERFs in Sequencer2D-S on images with resolution $224^2$.}
    \label{figure:erf_sequencer}
\end{figure}

%% file: erf_resnet_224.tex
\begin{figure}[tb]
  \raggedright
  \begin{minipage}[t]{0.13\hsize}
    \centering
    \includegraphics[width=\linewidth]{figures/erf_results/resnet50_224_layer1.0.pdf}
    \subcaption{Block 1}
    \label{figure:erf_resnet:0}
  \end{minipage}
  \begin{minipage}[t]{0.13\hsize}
    \centering
    \includegraphics[width=\linewidth]{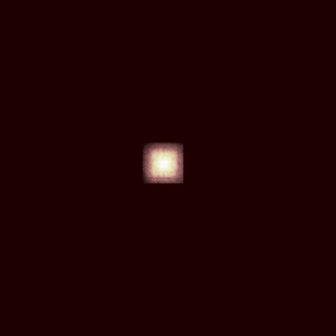}
    \subcaption{Block 2}
    \label{figure:erf_resnet:1}
  \end{minipage}
  \begin{minipage}[t]{0.13\hsize}
    \centering
    \includegraphics[width=\linewidth]{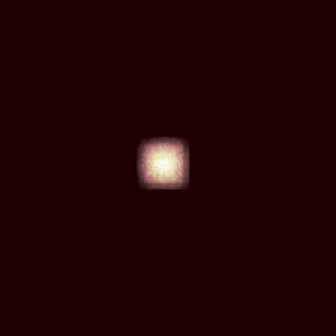}
    \subcaption{Block 3}
    \label{figure:erf_resnet:2}
  \end{minipage}
  \begin{minipage}[t]{0.13\hsize}
    \centering
    \includegraphics[width=\linewidth]{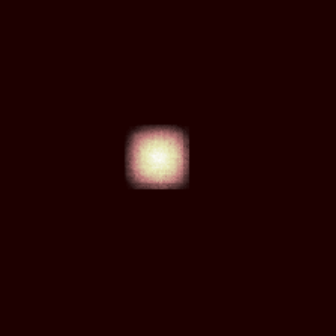}
    \subcaption{Block 4}
    \label{figure:erf_resnet:3}
  \end{minipage}
  \begin{minipage}[t]{0.13\hsize}
    \centering
    \includegraphics[width=\linewidth]{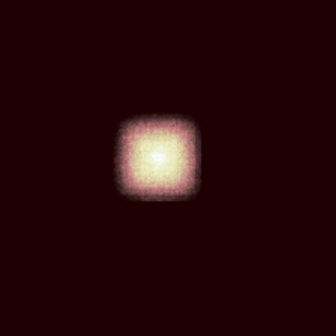}
    \subcaption{Block 5}
    \label{figure:erf_resnet:4}
  \end{minipage}
  \begin{minipage}[t]{0.13\hsize}
    \centering
    \includegraphics[width=\linewidth]{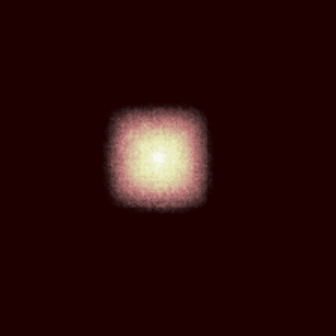}
    \subcaption{Block 6}
    \label{figure:erf_resnet:5}
  \end{minipage}
  \begin{minipage}[t]{0.13\hsize}
    \centering
    \includegraphics[width=\linewidth]{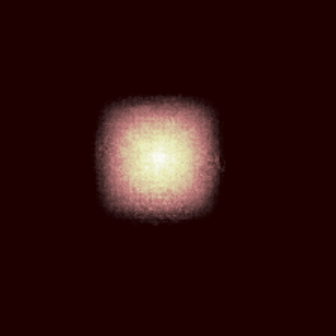}
    \subcaption{Block 7}
    \label{figure:erf_resnet:6}
  \end{minipage}
  \begin{minipage}[t]{0.13\hsize}
    \centering
    \includegraphics[width=\linewidth]{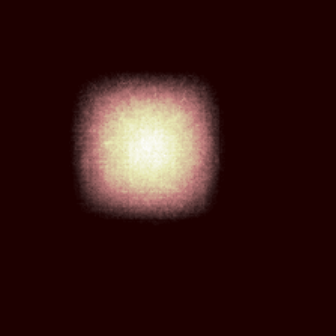}
    \subcaption{Block 8}
    \label{figure:erf_resnet:7}
  \end{minipage}
  \begin{minipage}[t]{0.13\hsize}
    \centering
    \includegraphics[width=\linewidth]{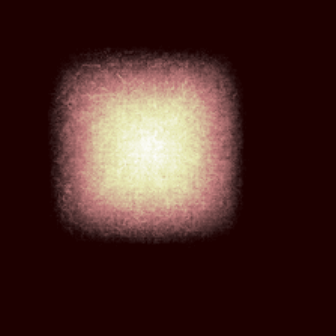}
    \subcaption{Block 9}
    \label{figure:erf_resnet:8}
  \end{minipage}
  \begin{minipage}[t]{0.13\hsize}
    \centering
    \includegraphics[width=\linewidth]{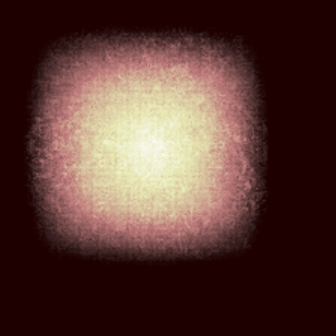}
    \subcaption{Block 10}
    \label{figure:erf_resnet:9}
  \end{minipage}
  \begin{minipage}[t]{0.13\hsize}
    \centering
    \includegraphics[width=\linewidth]{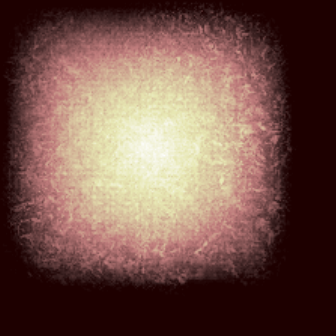}
    \subcaption{Block 11}
    \label{figure:erf_resnet:10}
  \end{minipage}
  \begin{minipage}[t]{0.13\hsize}
    \centering
    \includegraphics[width=\linewidth]{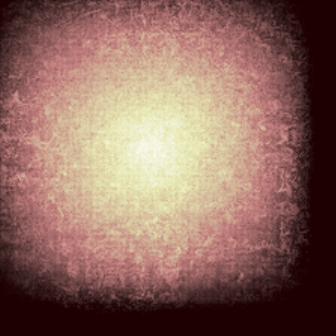}
    \subcaption{Block 12}
    \label{figure:erf_resnet:11}
  \end{minipage}
  \begin{minipage}[t]{0.13\hsize}
    \centering
    \includegraphics[width=\linewidth]{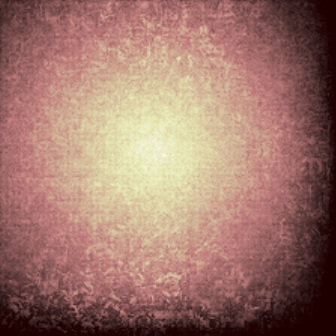}
    \subcaption{Block 13}
    \label{figure:erf_resnet:12}
  \end{minipage}
  \begin{minipage}[t]{0.13\hsize}
    \centering
    \includegraphics[width=\linewidth]{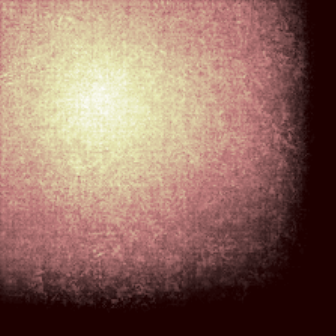}
    \subcaption{Block 14}
    \label{figure:erf_resnet:13}
  \end{minipage}
  \begin{minipage}[t]{0.13\hsize}
    \centering
    \includegraphics[width=\linewidth]{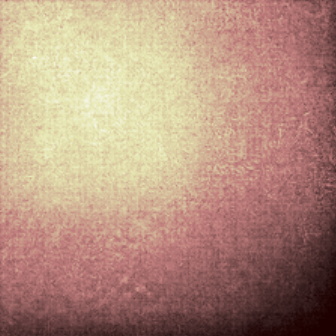}
    \subcaption{Block 15}
    \label{figure:erf_resnet:14}
  \end{minipage}
  \begin{minipage}[t]{0.13\hsize}
    \centering
    \includegraphics[width=\linewidth]{figures/erf_results/resnet50_224_layer4.2.pdf}
    \subcaption{Block 16}
    \label{figure:erf_resnet:15}
  \end{minipage}
  \caption{ERFs in ResNet-50 \cite{he2016deep} on images with resolution $224^2$.}
  \label{figure:erf_resnet}
\end{figure}

%% file: erf_convnext_224.tex
\begin{figure}[tb]
  \raggedright
  \begin{minipage}[t]{0.13\hsize}
    \centering
    \includegraphics[width=\linewidth]{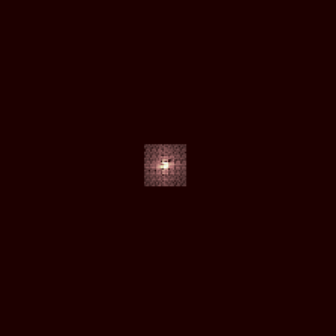}
    \subcaption{Block 1}
    \label{figure:erf_convnext:0}
  \end{minipage}
  \begin{minipage}[t]{0.13\hsize}
    \centering
    \includegraphics[width=\linewidth]{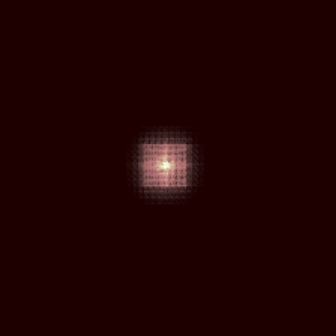}
    \subcaption{Block 2}
    \label{figure:erf_convnext:1}
  \end{minipage}
  \begin{minipage}[t]{0.13\hsize}
    \centering
    \includegraphics[width=\linewidth]{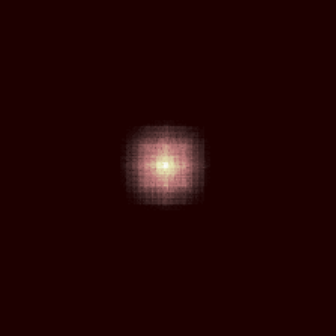}
    \subcaption{Block 3}
    \label{figure:erf_convnext:2}
  \end{minipage}
  \begin{minipage}[t]{0.13\hsize}
    \centering
    \includegraphics[width=\linewidth]{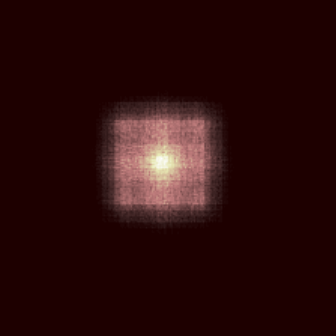}
    \subcaption{Block 4}
    \label{figure:erf_convnext:3}
  \end{minipage}
  \begin{minipage}[t]{0.13\hsize}
    \centering
    \includegraphics[width=\linewidth]{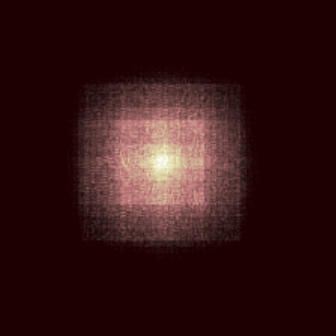}
    \subcaption{Block 5}
    \label{figure:erf_convnext:4}
  \end{minipage}
  \begin{minipage}[t]{0.13\hsize}
    \centering
    \includegraphics[width=\linewidth]{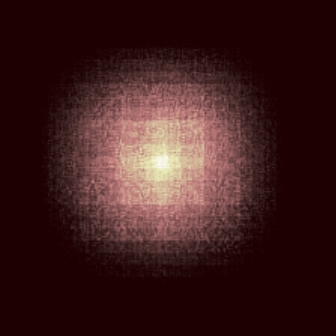}
    \subcaption{Block 6}
    \label{figure:erf_convnext:5}
  \end{minipage}
  \begin{minipage}[t]{0.13\hsize}
    \centering
    \includegraphics[width=\linewidth]{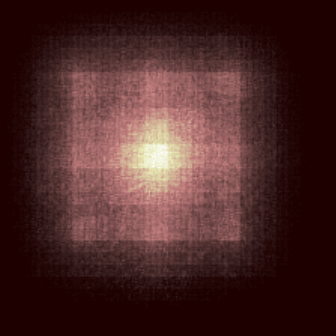}
    \subcaption{Block 7}
    \label{figure:erf_convnext:6}
  \end{minipage}
  \begin{minipage}[t]{0.13\hsize}
    \centering
    \includegraphics[width=\linewidth]{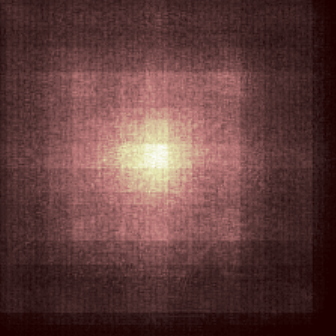}
    \subcaption{Block 8}
    \label{figure:erf_convnext:7}
  \end{minipage}
  \begin{minipage}[t]{0.13\hsize}
    \centering
    \includegraphics[width=\linewidth]{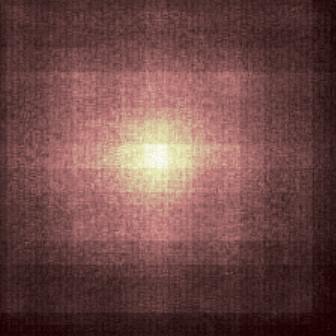}
    \subcaption{Block 9}
    \label{figure:erf_convnext:8}
  \end{minipage}
  \begin{minipage}[t]{0.13\hsize}
    \centering
    \includegraphics[width=\linewidth]{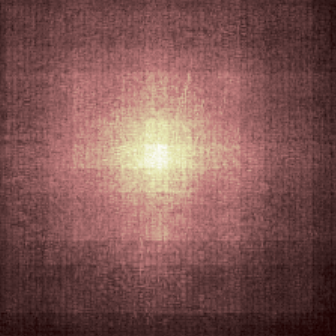}
    \subcaption{Block 10}
    \label{figure:erf_convnext:9}
  \end{minipage}
  \begin{minipage}[t]{0.13\hsize}
    \centering
    \includegraphics[width=\linewidth]{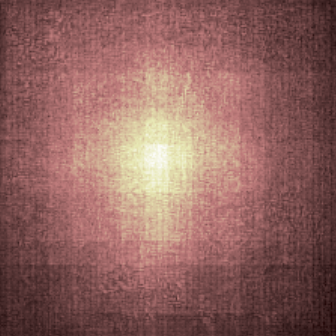}
    \subcaption{Block 11}
    \label{figure:erf_convnext:10}
  \end{minipage}
  \begin{minipage}[t]{0.13\hsize}
    \centering
    \includegraphics[width=\linewidth]{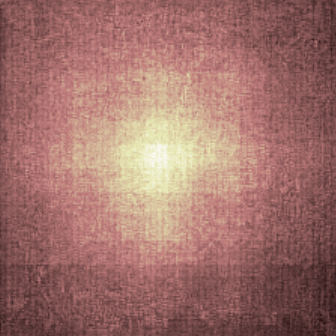}
    \subcaption{Block 12}
    \label{figure:erf_convnext:11}
  \end{minipage}
  \begin{minipage}[t]{0.13\hsize}
    \centering
    \includegraphics[width=\linewidth]{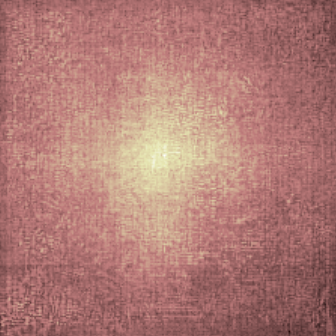}
    \subcaption{Block 13}
    \label{figure:erf_convnext:12}
  \end{minipage}
  \begin{minipage}[t]{0.13\hsize}
    \centering
    \includegraphics[width=\linewidth]{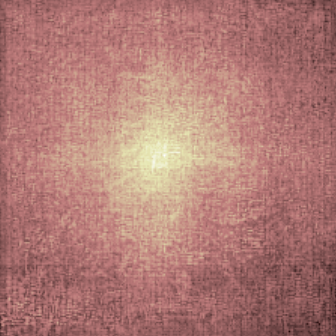}
    \subcaption{Block 14}
    \label{figure:erf_convnext:13}
  \end{minipage}
  \begin{minipage}[t]{0.13\hsize}
    \centering
    \includegraphics[width=\linewidth]{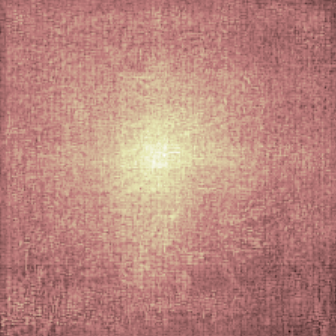}
    \subcaption{Block 15}
    \label{figure:erf_convnext:14}
  \end{minipage}
  \begin{minipage}[t]{0.13\hsize}
    \centering
    \includegraphics[width=\linewidth]{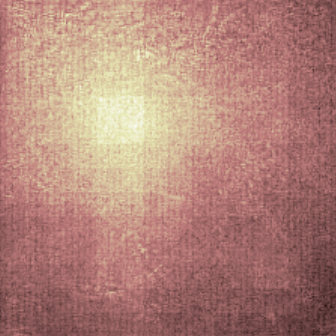}
    \subcaption{Block 16}
    \label{figure:erf_convnext:15}
  \end{minipage}
  \begin{minipage}[t]{0.13\hsize}
    \centering
    \includegraphics[width=\linewidth]{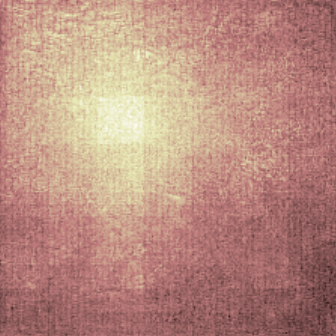}
    \subcaption{Block 17}
    \label{figure:erf_convnext:16}
  \end{minipage}
  \begin{minipage}[t]{0.13\hsize}
    \centering
    \includegraphics[width=\linewidth]{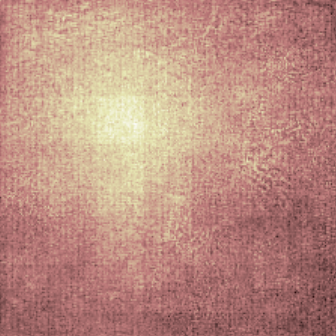}
    \subcaption{Block 18}
    \label{figure:erf_convnext:17}
  \end{minipage}
  \caption{ERFs in ConvNeXt-T \cite{liu2022convnet} on images with resolution $224^2$.}
    \label{figure:erf_convnext}
\end{figure}

%% file: erf_cyclemlp_224.tex
\begin{figure}[tb]
  \raggedright
  \begin{minipage}[t]{0.13\hsize}
    \centering
    \includegraphics[width=\linewidth]{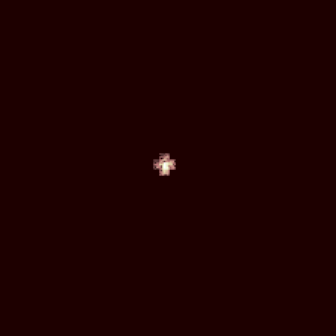}
    \subcaption{Block 1}
    \label{figure:erf_cyclemlp:0}
  \end{minipage}
  \begin{minipage}[t]{0.13\hsize}
    \centering
    \includegraphics[width=\linewidth]{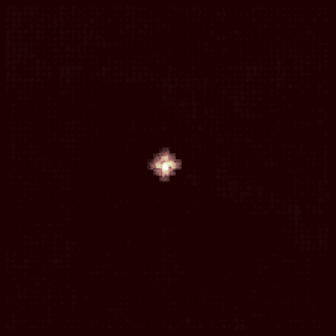}
    \subcaption{Block 2}
    \label{figure:erf_cyclemlp:1}
  \end{minipage}
  \begin{minipage}[t]{0.13\hsize}
    \centering
    \includegraphics[width=\linewidth]{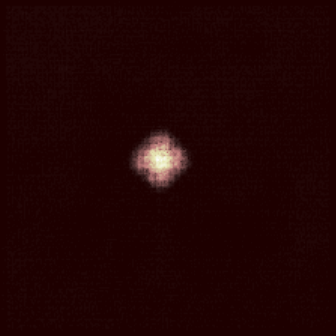}
    \subcaption{Block 3}
    \label{figure:erf_cyclemlp:2}
  \end{minipage}
  \begin{minipage}[t]{0.13\hsize}
    \centering
    \includegraphics[width=\linewidth]{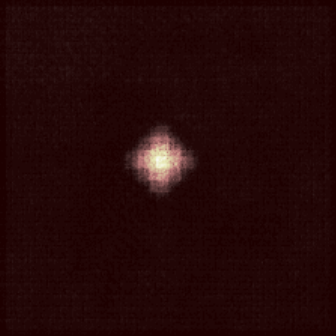}
    \subcaption{Block 4}
    \label{figure:erf_cyclemlp:3}
  \end{minipage}
  \begin{minipage}[t]{0.13\hsize}
    \centering
    \includegraphics[width=\linewidth]{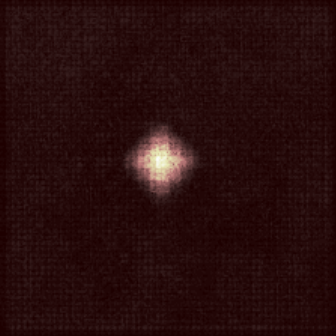}
    \subcaption{Block 5}
    \label{figure:erf_cyclemlp:4}
  \end{minipage}
  \begin{minipage}[t]{0.13\hsize}
    \centering
    \includegraphics[width=\linewidth]{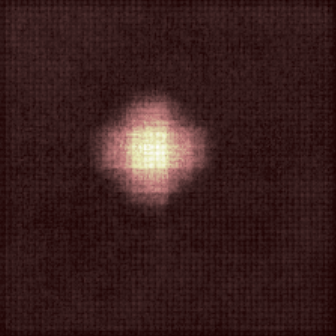}
    \subcaption{Block 6}
    \label{figure:erf_cyclemlp:5}
  \end{minipage}
  \begin{minipage}[t]{0.13\hsize}
    \centering
    \includegraphics[width=\linewidth]{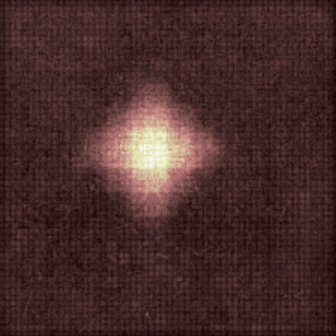}
    \subcaption{Block 7}
    \label{figure:erf_cyclemlp:6}
  \end{minipage}
  \begin{minipage}[t]{0.13\hsize}
    \centering
    \includegraphics[width=\linewidth]{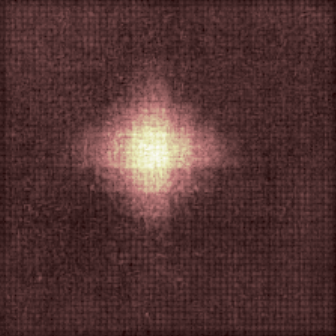}
    \subcaption{Block 8}
    \label{figure:erf_cyclemlp:7}
  \end{minipage}
  \begin{minipage}[t]{0.13\hsize}
    \centering
    \includegraphics[width=\linewidth]{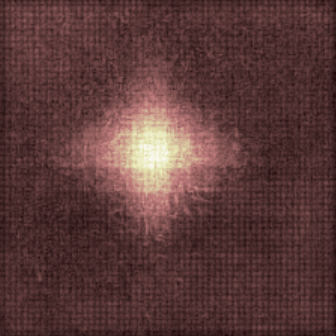}
    \subcaption{Block 9}
    \label{figure:erf_cyclemlp:8}
  \end{minipage}
  \begin{minipage}[t]{0.13\hsize}
    \centering
    \includegraphics[width=\linewidth]{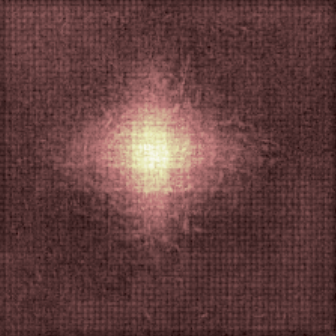}
    \subcaption{Block 10}
    \label{figure:erf_cyclemlp:9}
  \end{minipage}
  \begin{minipage}[t]{0.13\hsize}
    \centering
    \includegraphics[width=\linewidth]{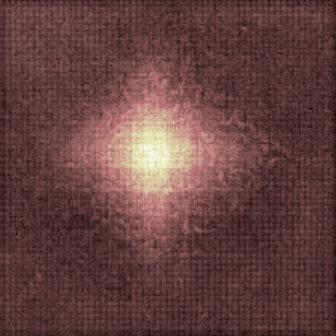}
    \subcaption{Block 11}
    \label{figure:erf_cyclemlp:10}
  \end{minipage}
  \begin{minipage}[t]{0.13\hsize}
    \centering
    \includegraphics[width=\linewidth]{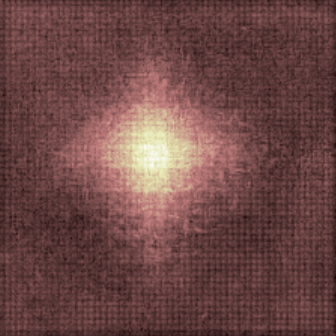}
    \subcaption{Block 12}
    \label{figure:erf_cyclemlp:11}
  \end{minipage}
  \begin{minipage}[t]{0.13\hsize}
    \centering
    \includegraphics[width=\linewidth]{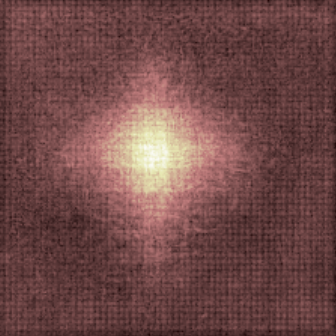}
    \subcaption{Block 13}
    \label{figure:erf_cyclemlp:12}
  \end{minipage}
  \begin{minipage}[t]{0.13\hsize}
    \centering
    \includegraphics[width=\linewidth]{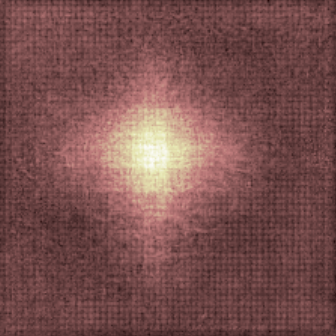}
    \subcaption{Block 14}
    \label{figure:erf_cyclemlp:13}
  \end{minipage}
  \begin{minipage}[t]{0.13\hsize}
    \centering
    \includegraphics[width=\linewidth]{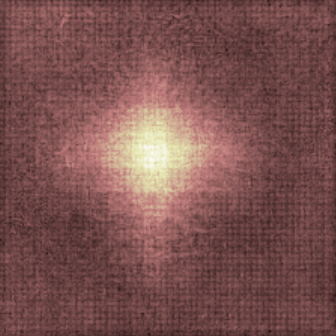}
    \subcaption{Block 15}
    \label{figure:erf_cyclemlp:14}
  \end{minipage}
  \begin{minipage}[t]{0.13\hsize}
    \centering
    \includegraphics[width=\linewidth]{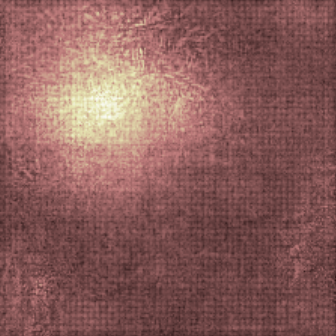}
    \subcaption{Block 16}
    \label{figure:erf_cyclemlp:15}
  \end{minipage}
  \begin{minipage}[t]{0.13\hsize}
    \centering
    \includegraphics[width=\linewidth]{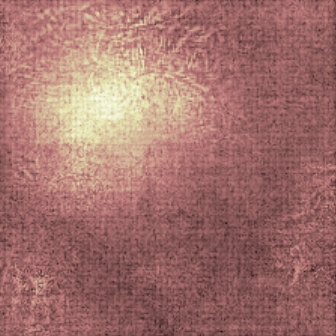}
    \subcaption{Block 17}
    \label{figure:erf_cyclemlp:16}
  \end{minipage}
  \begin{minipage}[t]{0.13\hsize}
    \centering
    \includegraphics[width=\linewidth]{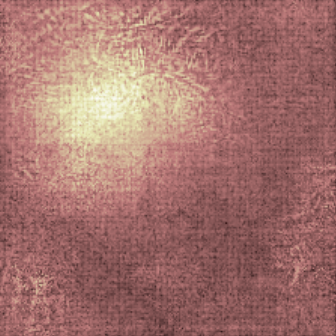}
    \subcaption{Block 18}
    \label{figure:erf_cyclemlp:17}
  \end{minipage}
  \caption{ERFs in CycleMLP-B2 \cite{chen2022cyclemlp} on images with resolution $224^2$.}
    \label{figure:erf_cyclemlp}
\end{figure}

%% file: erf_deit_224.tex
\begin{figure}[tb]
  \raggedright
  \begin{minipage}[t]{0.13\hsize}
    \centering
    \includegraphics[width=\linewidth]{figures/erf_results/deit_small_patch16_224_224_blocks.0.pdf}
    \subcaption{Block 1}
    \label{figure:erf_deit:0}
  \end{minipage}
  \begin{minipage}[t]{0.13\hsize}
    \centering
    \includegraphics[width=\linewidth]{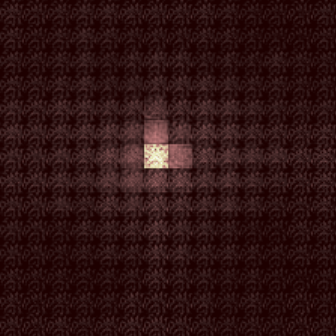}
    \subcaption{Block 2}
    \label{figure:erf_deit:1}
  \end{minipage}
  \begin{minipage}[t]{0.13\hsize}
    \centering
    \includegraphics[width=\linewidth]{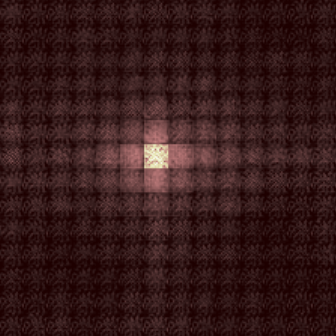}
    \subcaption{Block 3}
    \label{figure:erf_deit:2}
  \end{minipage}
  \begin{minipage}[t]{0.13\hsize}
    \centering
    \includegraphics[width=\linewidth]{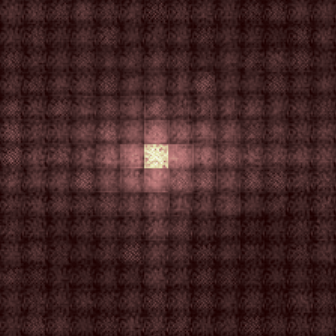}
    \subcaption{Block 4}
    \label{figure:erf_deit:3}
  \end{minipage}
  \begin{minipage}[t]{0.13\hsize}
    \centering
    \includegraphics[width=\linewidth]{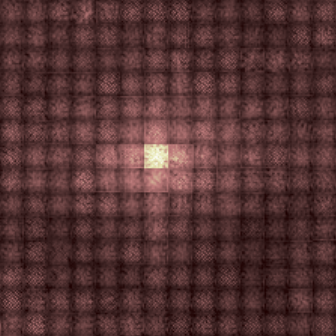}
    \subcaption{Block 5}
    \label{figure:erf_deit:4}
  \end{minipage}
  \begin{minipage}[t]{0.13\hsize}
    \centering
    \includegraphics[width=\linewidth]{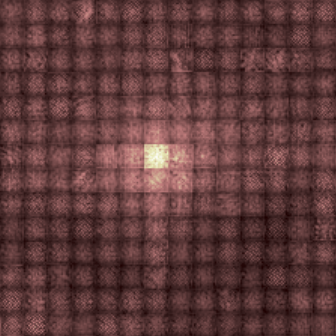}
    \subcaption{Block 6}
    \label{figure:erf_deit:5}
  \end{minipage}
  \begin{minipage}[t]{0.13\hsize}
    \centering
    \includegraphics[width=\linewidth]{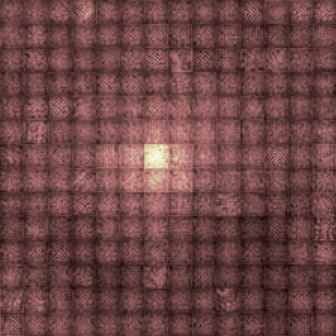}
    \subcaption{Block 7}
    \label{figure:erf_deit:6}
  \end{minipage}
  \begin{minipage}[t]{0.13\hsize}
    \centering
    \includegraphics[width=\linewidth]{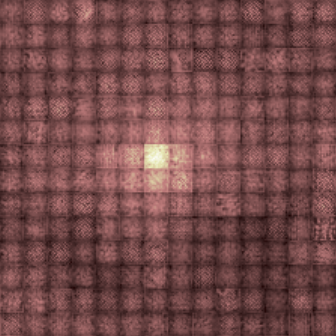}
    \subcaption{Block 8}
    \label{figure:erf_deit:7}
  \end{minipage}
  \begin{minipage}[t]{0.13\hsize}
    \centering
    \includegraphics[width=\linewidth]{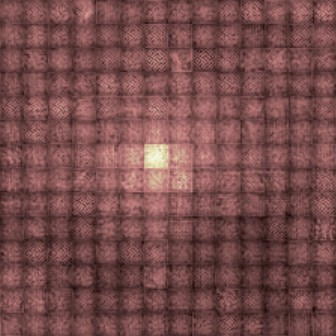}
    \subcaption{Block 9}
    \label{figure:erf_deit:8}
  \end{minipage}
  \begin{minipage}[t]{0.13\hsize}
    \centering
    \includegraphics[width=\linewidth]{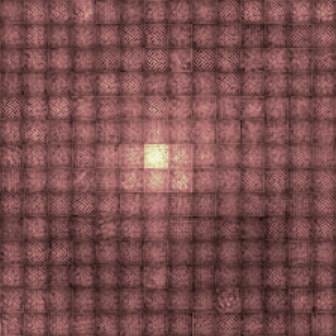}
    \subcaption{Block 10}
    \label{figure:erf_deit:9}
  \end{minipage}
  \begin{minipage}[t]{0.13\hsize}
    \centering
    \includegraphics[width=\linewidth]{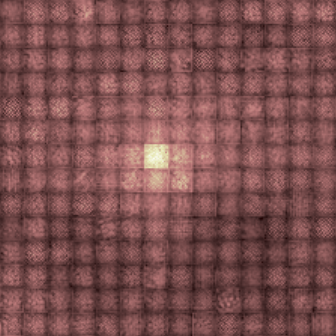}
    \subcaption{Block 11}
    \label{figure:erf_deit:10}
  \end{minipage}
  \begin{minipage}[t]{0.13\hsize}
    \centering
    \includegraphics[width=\linewidth]{figures/erf_results/deit_small_patch16_224_224_blocks.11.pdf}
    \subcaption{Block 12}
    \label{figure:erf_deit:11}
  \end{minipage}
  \caption{ERFs in DeiT-S \cite{touvron2020training} on images with resolution $224^2$.}
    \label{figure:erf_deit}
\end{figure}

%% file: erf_swin_224.tex
\begin{figure}[tb]
  \raggedright
  \begin{minipage}[t]{0.13\hsize}
    \centering
    \includegraphics[width=\linewidth]{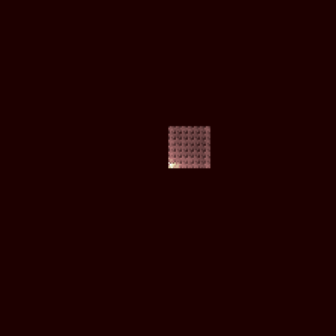}
    \subcaption{Block 1}
    \label{figure:erf_swin:0}
  \end{minipage}
  \begin{minipage}[t]{0.13\hsize}
    \centering
    \includegraphics[width=\linewidth]{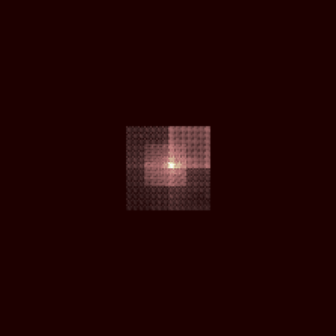}
    \subcaption{Block 2}
    \label{figure:erf_swin:1}
  \end{minipage}
  \begin{minipage}[t]{0.13\hsize}
    \centering
    \includegraphics[width=\linewidth]{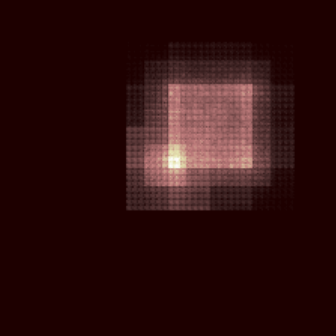}
    \subcaption{Block 3}
    \label{figure:erf_swin:2}
  \end{minipage}
  \begin{minipage}[t]{0.13\hsize}
    \centering
    \includegraphics[width=\linewidth]{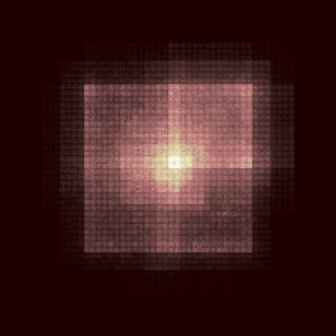}
    \subcaption{Block 4}
    \label{figure:erf_swin:3}
  \end{minipage}
  \begin{minipage}[t]{0.13\hsize}
    \centering
    \includegraphics[width=\linewidth]{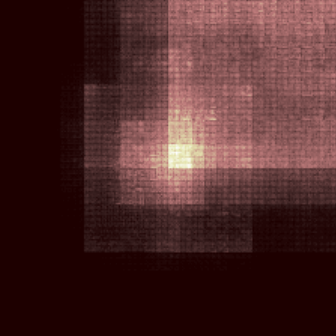}
    \subcaption{Block 5}
    \label{figure:erf_swin:4}
  \end{minipage}
  \begin{minipage}[t]{0.13\hsize}
    \centering
    \includegraphics[width=\linewidth]{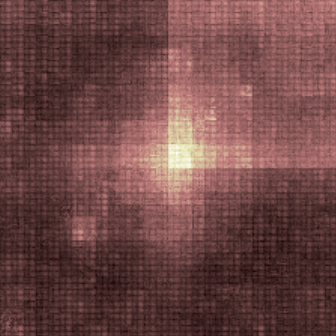}
    \subcaption{Block 6}
    \label{figure:erf_swin:5}
  \end{minipage}
  \begin{minipage}[t]{0.13\hsize}
    \centering
    \includegraphics[width=\linewidth]{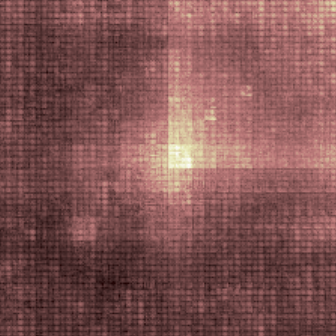}
    \subcaption{Block 7}
    \label{figure:erf_swin:6}
  \end{minipage}
  \begin{minipage}[t]{0.13\hsize}
    \centering
    \includegraphics[width=\linewidth]{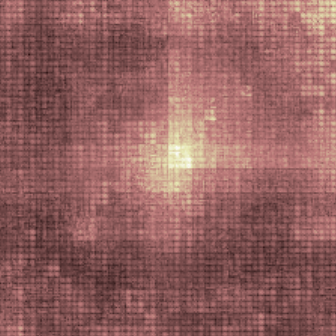}
    \subcaption{Block 8}
    \label{figure:erf_swin:7}
  \end{minipage}
  \begin{minipage}[t]{0.13\hsize}
    \centering
    \includegraphics[width=\linewidth]{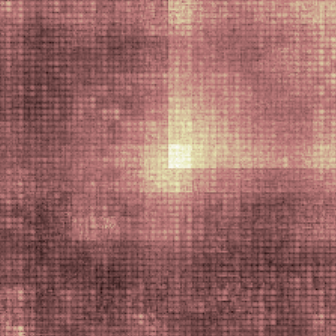}
    \subcaption{Block 9}
    \label{figure:erf_swin:8}
  \end{minipage}
  \begin{minipage}[t]{0.13\hsize}
    \centering
    \includegraphics[width=\linewidth]{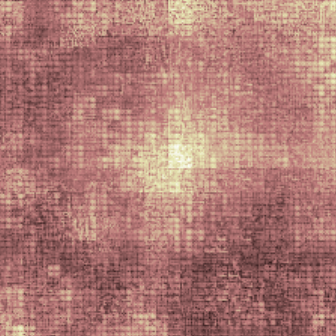}
    \subcaption{Block 10}
    \label{figure:erf_swin:9}
  \end{minipage}
  \begin{minipage}[t]{0.13\hsize}
    \centering
    \includegraphics[width=\linewidth]{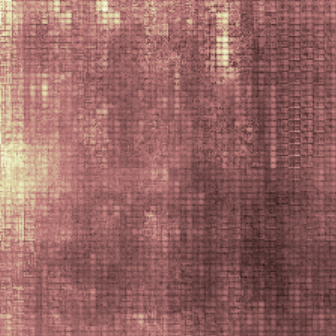}
    \subcaption{Block 11}
    \label{figure:erf_swin:10}
  \end{minipage}
  \begin{minipage}[t]{0.13\hsize}
    \centering
    \includegraphics[width=\linewidth]{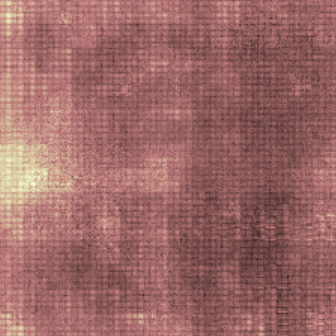}
    \subcaption{Block 12}
    \label{figure:erf_swin:11}
  \end{minipage}
  \caption{ERFs in Swin-T \cite{liu2021swin} on images with resolution $224^2$.}
    \label{figure:erf_swin}
\end{figure}

%% file: erf_gfnet_224.tex
\begin{figure}[tb]
  \raggedright
  \begin{minipage}[t]{0.13\hsize}
    \centering
    \includegraphics[width=\linewidth]{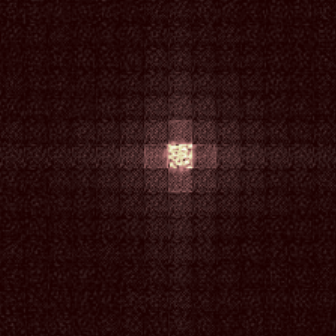}
    \subcaption{Block 1}
    \label{figure:erf_gfnet:0}
  \end{minipage}
  \begin{minipage}[t]{0.13\hsize}
    \centering
    \includegraphics[width=\linewidth]{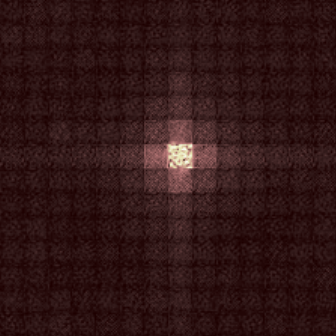}
    \subcaption{Block 2}
    \label{figure:erf_gfnet:1}
  \end{minipage}
  \begin{minipage}[t]{0.13\hsize}
    \centering
    \includegraphics[width=\linewidth]{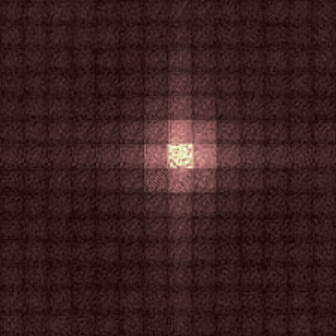}
    \subcaption{Block 3}
    \label{figure:erf_gfnet:2}
  \end{minipage}
  \begin{minipage}[t]{0.13\hsize}
    \centering
    \includegraphics[width=\linewidth]{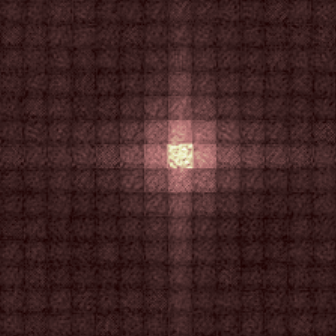}
    \subcaption{Block 4}
    \label{figure:erf_gfnet:3}
  \end{minipage}
  \begin{minipage}[t]{0.13\hsize}
    \centering
    \includegraphics[width=\linewidth]{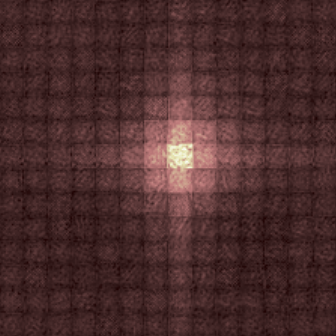}
    \subcaption{Block 5}
    \label{figure:erf_gfnet:4}
  \end{minipage}
  \begin{minipage}[t]{0.13\hsize}
    \centering
    \includegraphics[width=\linewidth]{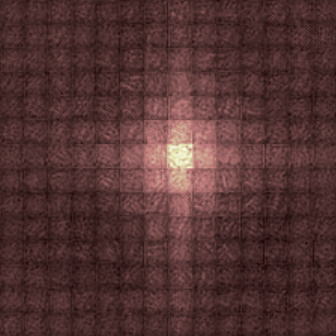}
    \subcaption{Block 6}
    \label{figure:erf_gfnet:5}
  \end{minipage}
  \begin{minipage}[t]{0.13\hsize}
    \centering
    \includegraphics[width=\linewidth]{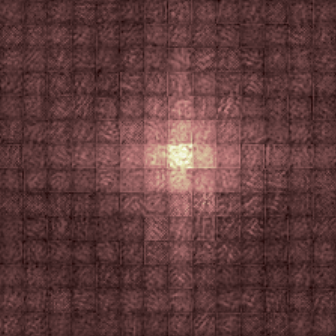}
    \subcaption{Block 7}
    \label{figure:erf_gfnet:6}
  \end{minipage}
  \begin{minipage}[t]{0.13\hsize}
    \centering
    \includegraphics[width=\linewidth]{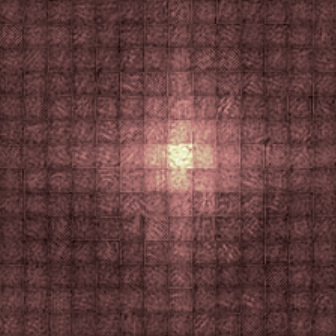}
    \subcaption{Block 8}
    \label{figure:erf_gfnet:7}
  \end{minipage}
  \begin{minipage}[t]{0.13\hsize}
    \centering
    \includegraphics[width=\linewidth]{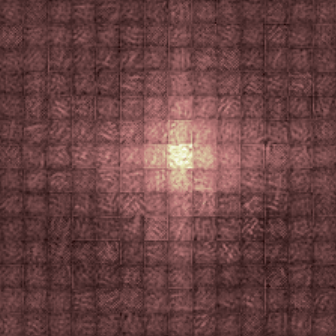}
    \subcaption{Block 9}
    \label{figure:erf_gfnet:8}
  \end{minipage}
  \begin{minipage}[t]{0.13\hsize}
    \centering
    \includegraphics[width=\linewidth]{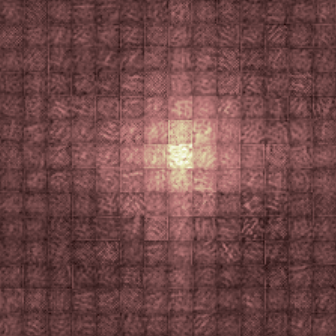}
    \subcaption{Block 10}
    \label{figure:erf_gfnet:9}
  \end{minipage}
  \begin{minipage}[t]{0.13\hsize}
    \centering
    \includegraphics[width=\linewidth]{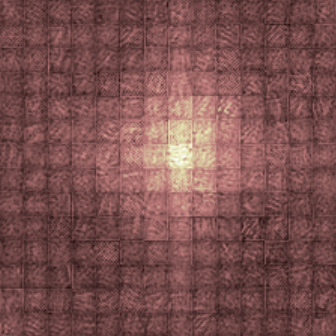}
    \subcaption{Block 11}
    \label{figure:erf_gfnet:10}
  \end{minipage}
  \begin{minipage}[t]{0.13\hsize}
    \centering
    \includegraphics[width=\linewidth]{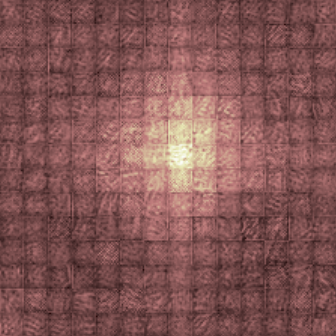}
    \subcaption{Block 12}
    \label{figure:erf_gfnet:11}
  \end{minipage}
  \begin{minipage}[t]{0.13\hsize}
    \centering
    \includegraphics[width=\linewidth]{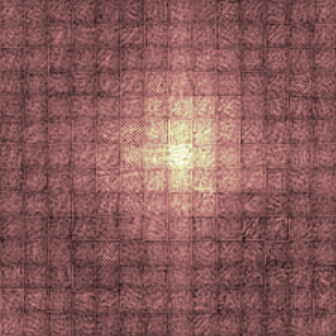}
    \subcaption{Block 13}
    \label{figure:erf_gfnet:12}
  \end{minipage}
  \begin{minipage}[t]{0.13\hsize}
    \centering
    \includegraphics[width=\linewidth]{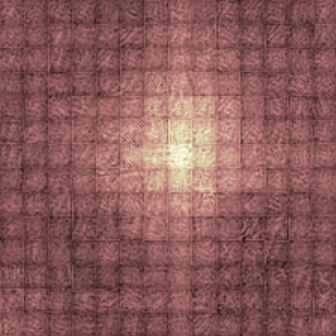}
    \subcaption{Block 14}
    \label{figure:erf_gfnet:13}
  \end{minipage}
  \begin{minipage}[t]{0.13\hsize}
    \centering
    \includegraphics[width=\linewidth]{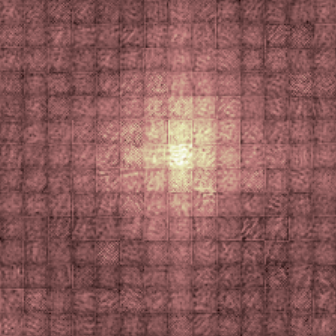}
    \subcaption{Block 15}
    \label{figure:erf_gfnet:14}
  \end{minipage}
  \begin{minipage}[t]{0.13\hsize}
    \centering
    \includegraphics[width=\linewidth]{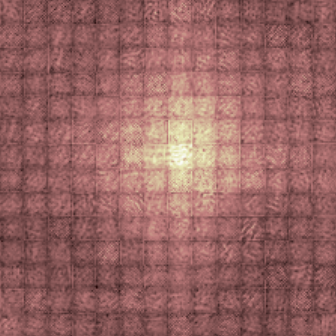}
    \subcaption{Block 16}
    \label{figure:erf_gfnet:15}
  \end{minipage}
  \begin{minipage}[t]{0.13\hsize}
    \centering
    \includegraphics[width=\linewidth]{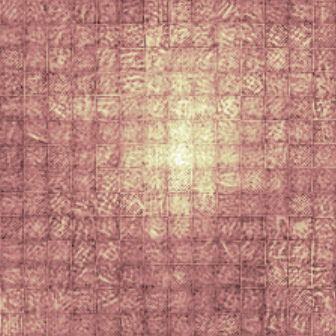}
    \subcaption{Block 17}
    \label{figure:erf_gfnet:16}
  \end{minipage}
  \begin{minipage}[t]{0.13\hsize}
    \centering
    \includegraphics[width=\linewidth]{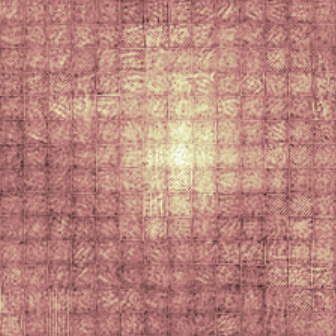}
    \subcaption{Block 18}
    \label{figure:erf_gfnet:17}
  \end{minipage}
  \begin{minipage}[t]{0.13\hsize}
    \centering
    \includegraphics[width=\linewidth]{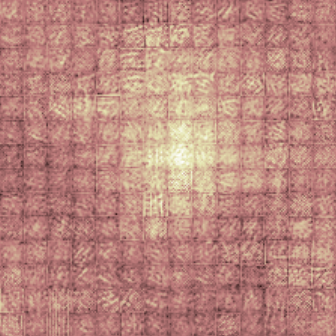}
    \subcaption{Block 19}
    \label{figure:erf_gfnet:18}
  \end{minipage}
  \caption{ERFs in GFNet-S \cite{rao2021global} on images with resolution $224^2$.}
    \label{figure:erf_gfnet}
\end{figure}

%% file: erf_vip_224.tex
\begin{figure}[tb]
  \raggedright
  \begin{minipage}[t]{0.13\hsize}
    \centering
    \includegraphics[width=\linewidth]{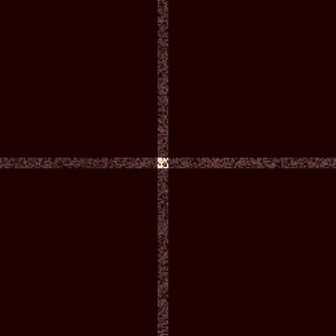}
    \subcaption{Block 1}
    \label{figure:erf_vip:0}
  \end{minipage}
  \begin{minipage}[t]{0.13\hsize}
    \centering
    \includegraphics[width=\linewidth]{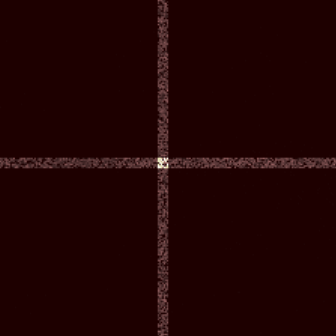}
    \subcaption{Block 2}
    \label{figure:erf_vip:1}
  \end{minipage}
  \begin{minipage}[t]{0.13\hsize}
    \centering
    \includegraphics[width=\linewidth]{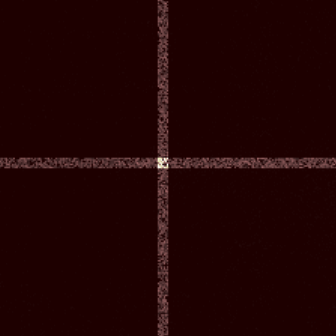}
    \subcaption{Block 3}
    \label{figure:erf_vip:2}
  \end{minipage}
  \begin{minipage}[t]{0.13\hsize}
    \centering
    \includegraphics[width=\linewidth]{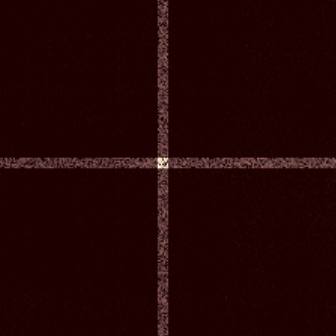}
    \subcaption{Block 4}
    \label{figure:erf_vip:3}
  \end{minipage}
  \begin{minipage}[t]{0.13\hsize}
    \centering
    \includegraphics[width=\linewidth]{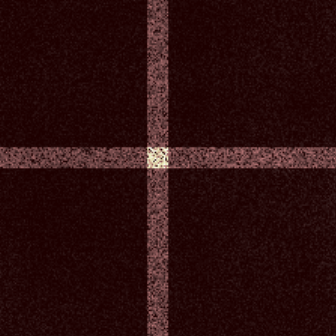}
    \subcaption{Block 5}
    \label{figure:erf_vip:4}
  \end{minipage}
  \begin{minipage}[t]{0.13\hsize}
    \centering
    \includegraphics[width=\linewidth]{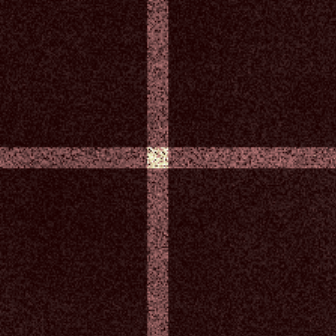}
    \subcaption{Block 6}
    \label{figure:erf_vip:5}
  \end{minipage}
  \begin{minipage}[t]{0.13\hsize}
    \centering
    \includegraphics[width=\linewidth]{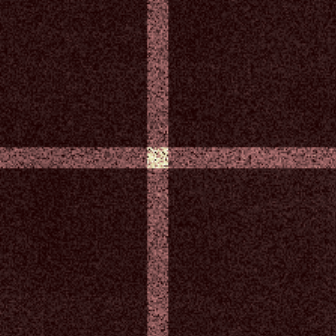}
    \subcaption{Block 7}
    \label{figure:erf_vip:6}
  \end{minipage}
  \begin{minipage}[t]{0.13\hsize}
    \centering
    \includegraphics[width=\linewidth]{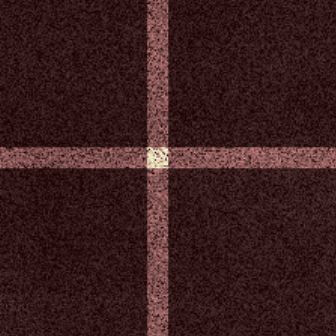}
    \subcaption{Block 8}
    \label{figure:erf_vip:7}
  \end{minipage}
  \begin{minipage}[t]{0.13\hsize}
    \centering
    \includegraphics[width=\linewidth]{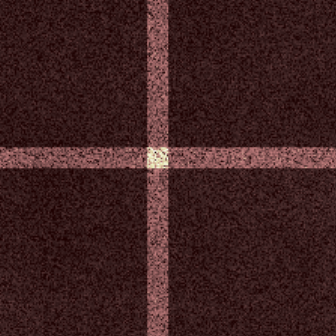}
    \subcaption{Block 9}
    \label{figure:erf_vip:8}
  \end{minipage}
  \begin{minipage}[t]{0.13\hsize}
    \centering
    \includegraphics[width=\linewidth]{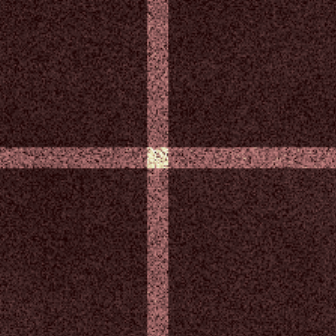}
    \subcaption{Block 10}
    \label{figure:erf_vip:9}
  \end{minipage}
  \begin{minipage}[t]{0.13\hsize}
    \centering
    \includegraphics[width=\linewidth]{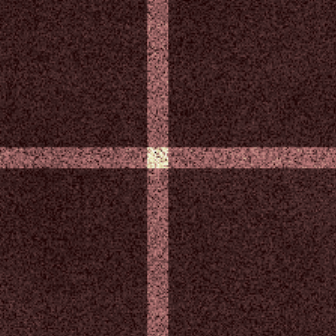}
    \subcaption{Block 11}
    \label{figure:erf_vip:10}
  \end{minipage}
  \begin{minipage}[t]{0.13\hsize}
    \centering
    \includegraphics[width=\linewidth]{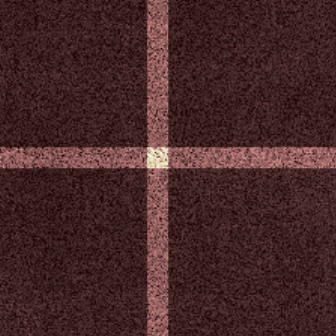}
    \subcaption{Block 12}
    \label{figure:erf_vip:11}
  \end{minipage}
  \begin{minipage}[t]{0.13\hsize}
    \centering
    \includegraphics[width=\linewidth]{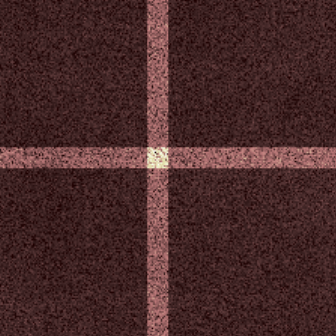}
    \subcaption{Block 13}
    \label{figure:erf_vip:12}
  \end{minipage}
  \begin{minipage}[t]{0.13\hsize}
    \centering
    \includegraphics[width=\linewidth]{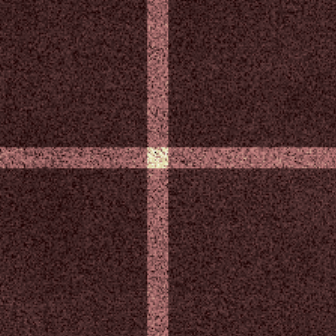}
    \subcaption{Block 14}
    \label{figure:erf_vip:13}
  \end{minipage}
  \begin{minipage}[t]{0.13\hsize}
    \centering
    \includegraphics[width=\linewidth]{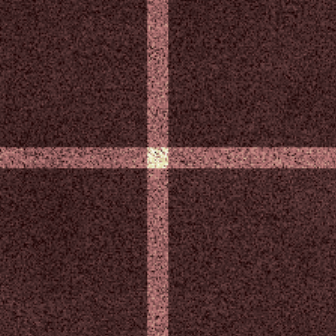}
    \subcaption{Block 15}
    \label{figure:erf_vip:14}
  \end{minipage}
  \begin{minipage}[t]{0.13\hsize}
    \centering
    \includegraphics[width=\linewidth]{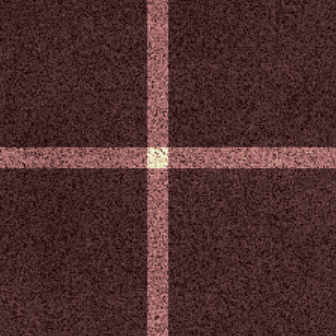}
    \subcaption{Block 16}
    \label{figure:erf_vip:15}
  \end{minipage}
  \begin{minipage}[t]{0.13\hsize}
    \centering
    \includegraphics[width=\linewidth]{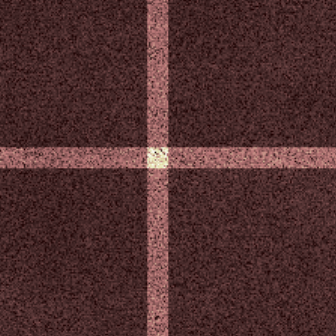}
    \subcaption{Block 17}
    \label{figure:erf_vip:16}
  \end{minipage}
  \begin{minipage}[t]{0.13\hsize}
    \centering
    \includegraphics[width=\linewidth]{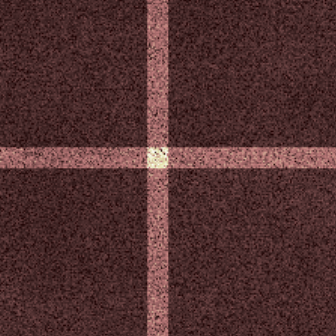}
    \subcaption{Block 18}
    \label{figure:erf_vip:17}
  \end{minipage}
  \caption{ERFs in ViP-S/7 \cite{hou2022vision} on images with resolution $224^2$.}
    \label{figure:erf_vip}
\end{figure}

%% file: erf_sequencer_448.tex
\begin{figure}[tb]
  \raggedright
  \begin{minipage}[t]{0.13\hsize}
    \centering
    \includegraphics[width=\linewidth]{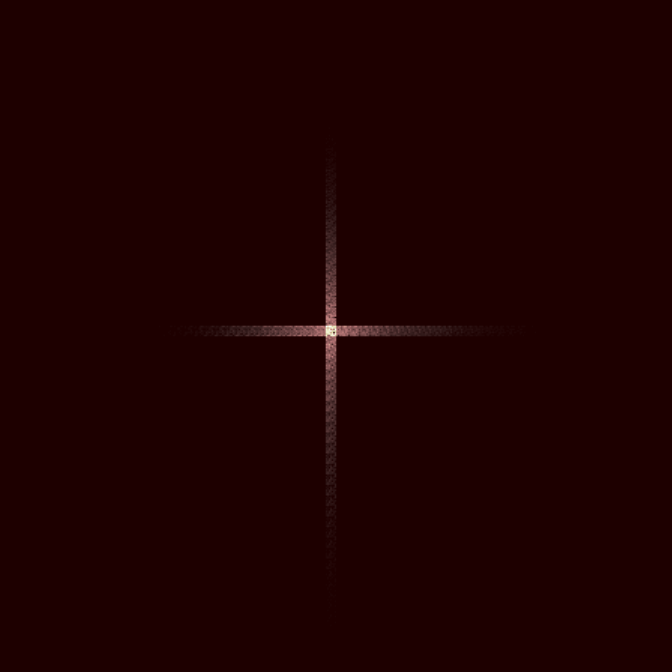}
    \subcaption{Block 1}
    \label{figure:erf_sequencer_448:0}
  \end{minipage}
  \begin{minipage}[t]{0.13\hsize}
    \centering
    \includegraphics[width=\linewidth]{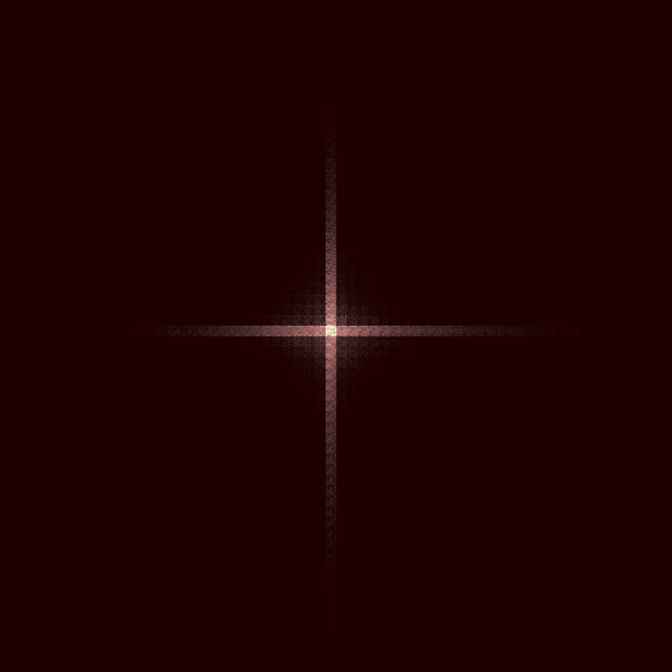}
    \subcaption{Block 2}
    \label{figure:erf_sequencer_448:1}
  \end{minipage}
  \begin{minipage}[t]{0.13\hsize}
    \centering
    \includegraphics[width=\linewidth]{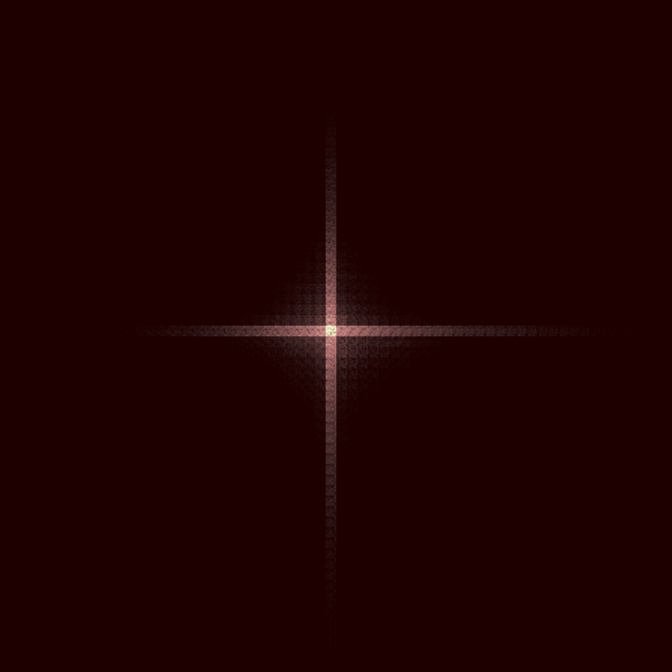}
    \subcaption{Block 3}
    \label{figure:erf_sequencer_448:2}
  \end{minipage}
  \begin{minipage}[t]{0.13\hsize}
    \centering
    \includegraphics[width=\linewidth]{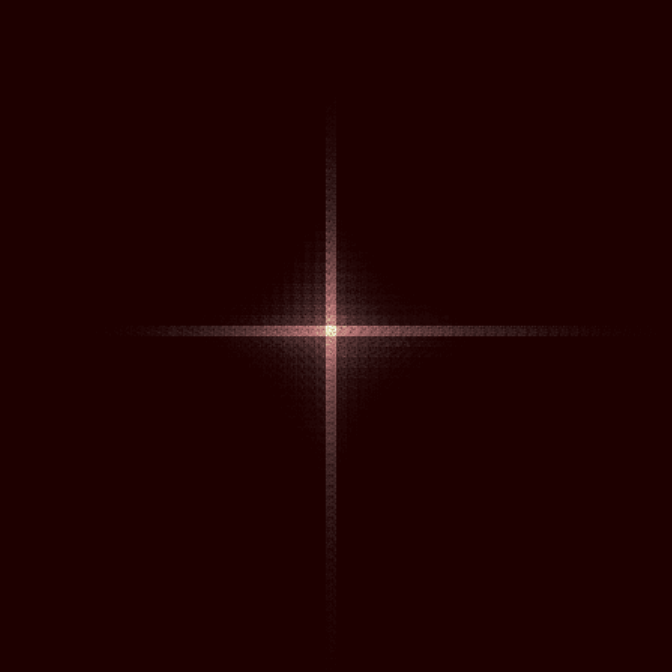}
    \subcaption{Block 4}
    \label{figure:erf_sequencer_448:3}
  \end{minipage}
  \begin{minipage}[t]{0.13\hsize}
    \centering
    \includegraphics[width=\linewidth]{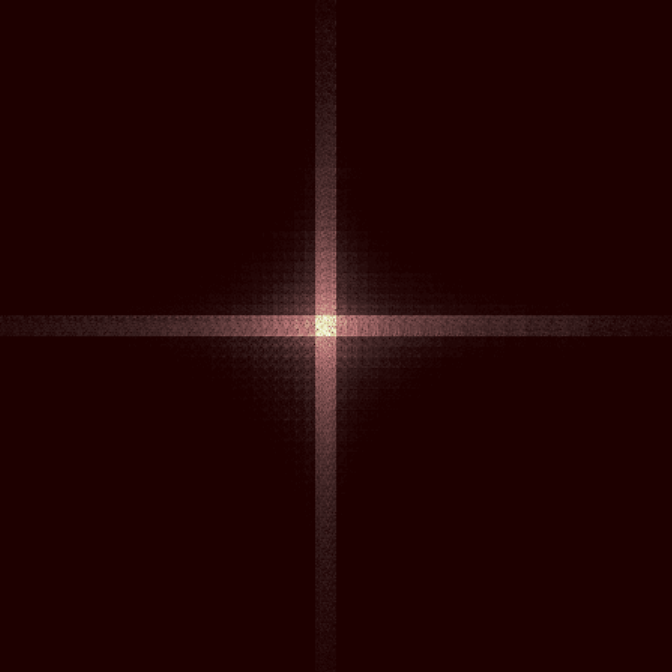}
    \subcaption{Block 5}
    \label{figure:erf_sequencer_448:4}
  \end{minipage}
  \begin{minipage}[t]{0.13\hsize}
    \centering
    \includegraphics[width=\linewidth]{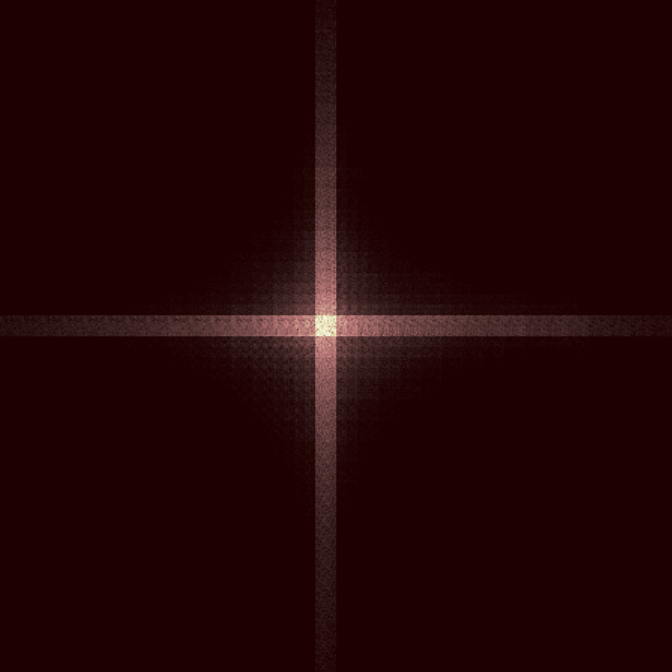}
    \subcaption{Block 6}
    \label{figure:erf_sequencer_448:5}
  \end{minipage}
  \begin{minipage}[t]{0.13\hsize}
    \centering
    \includegraphics[width=\linewidth]{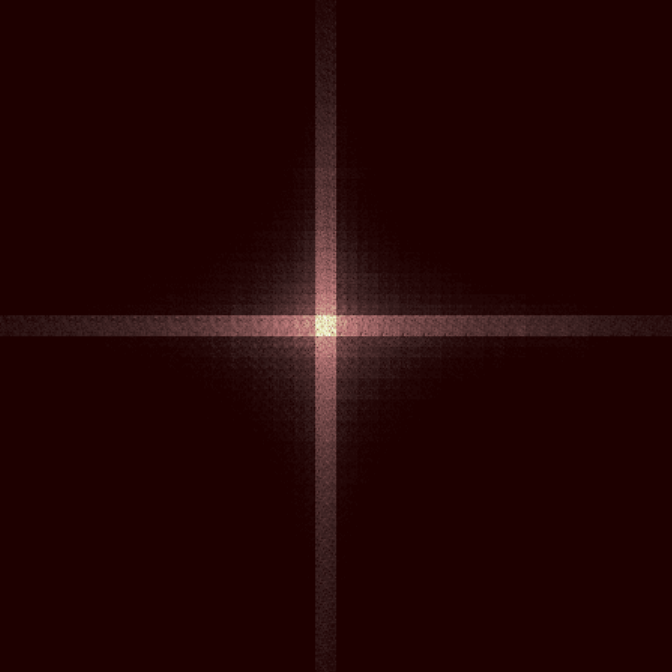}
    \subcaption{Block 7}
    \label{figure:erf_sequencer_448:6}
  \end{minipage}
  \begin{minipage}[t]{0.13\hsize}
    \centering
    \includegraphics[width=\linewidth]{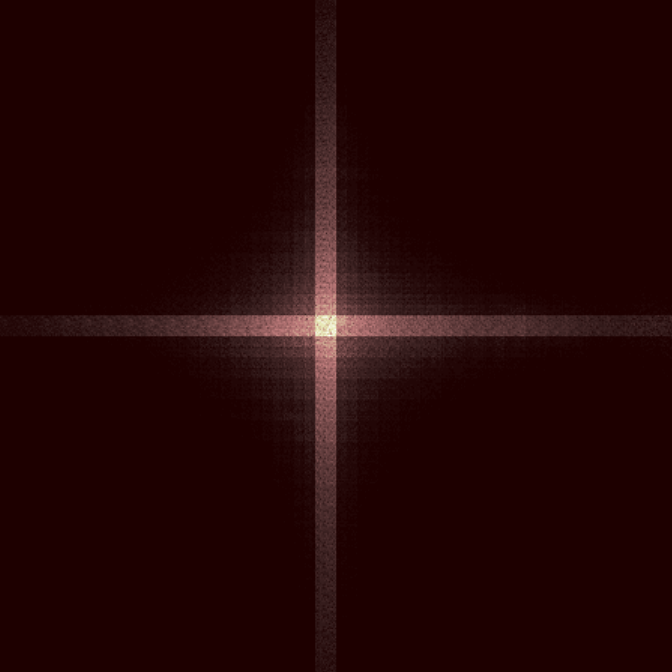}
    \subcaption{Block 8}
    \label{figure:erf_sequencer_448:7}
  \end{minipage}
  \begin{minipage}[t]{0.13\hsize}
    \centering
    \includegraphics[width=\linewidth]{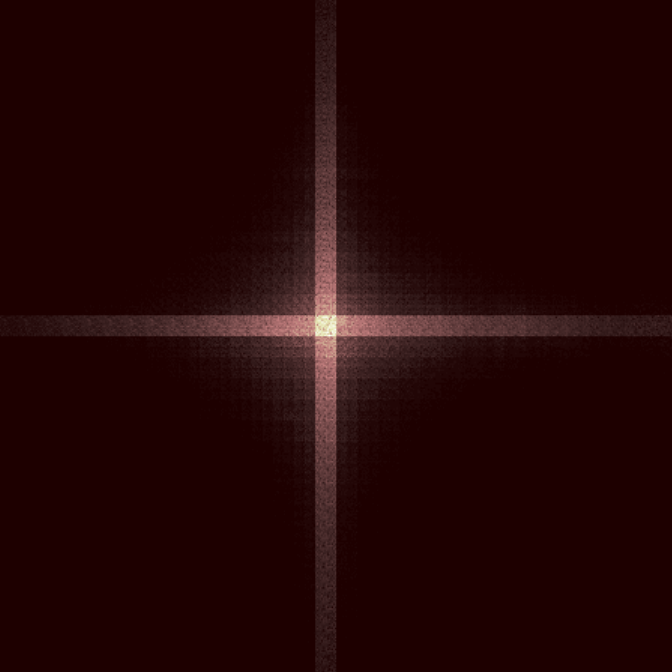}
    \subcaption{Block 9}
    \label{figure:erf_sequencer_448:8}
  \end{minipage}
  \begin{minipage}[t]{0.13\hsize}
    \centering
    \includegraphics[width=\linewidth]{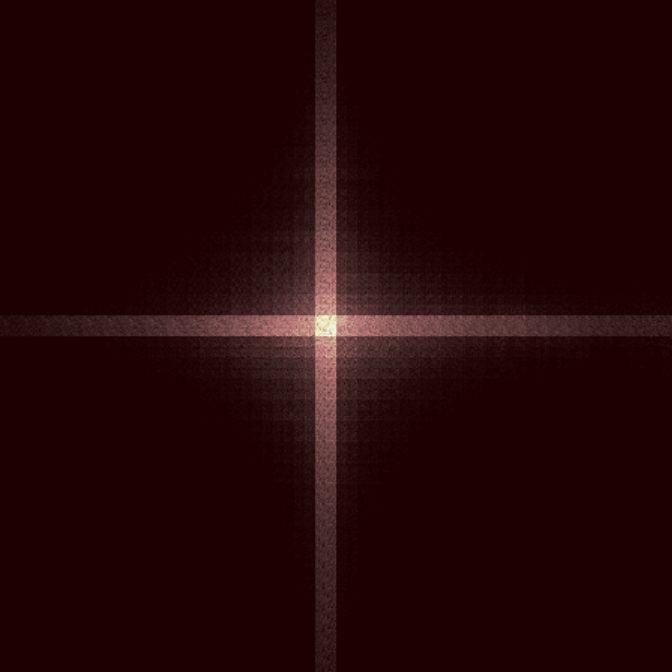}
    \subcaption{Block 10}
    \label{figure:erf_sequencer_448:9}
  \end{minipage}
  \begin{minipage}[t]{0.13\hsize}
    \centering
    \includegraphics[width=\linewidth]{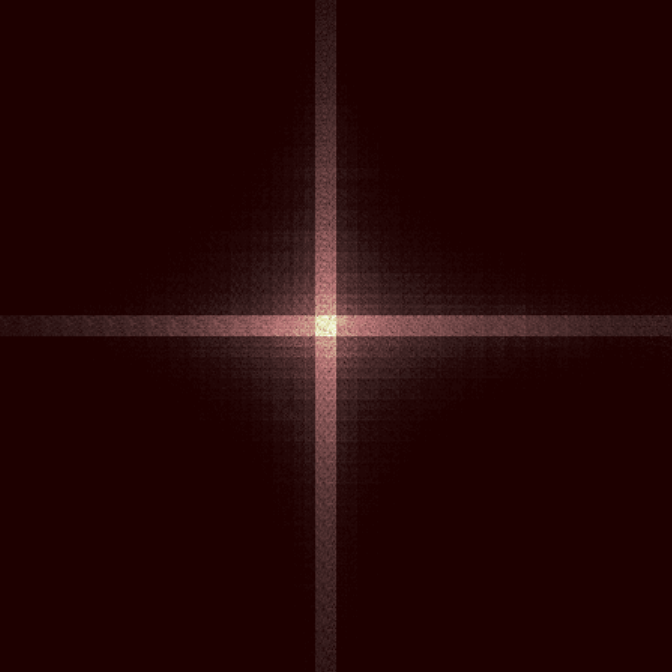}
    \subcaption{Block 11}
    \label{figure:erf_sequencer_448:10}
  \end{minipage}
  \begin{minipage}[t]{0.13\hsize}
    \centering
    \includegraphics[width=\linewidth]{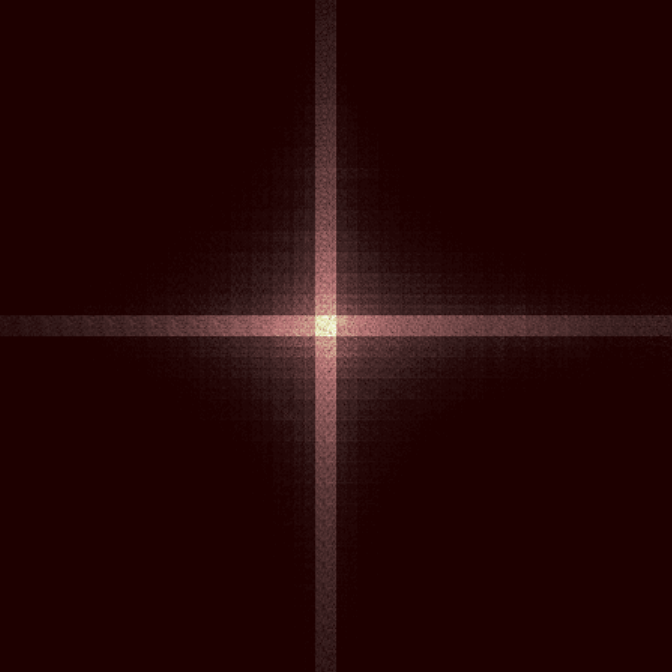}
    \subcaption{Block 12}
    \label{figure:erf_sequencer_448:11}
  \end{minipage}
  \begin{minipage}[t]{0.13\hsize}
    \centering
    \includegraphics[width=\linewidth]{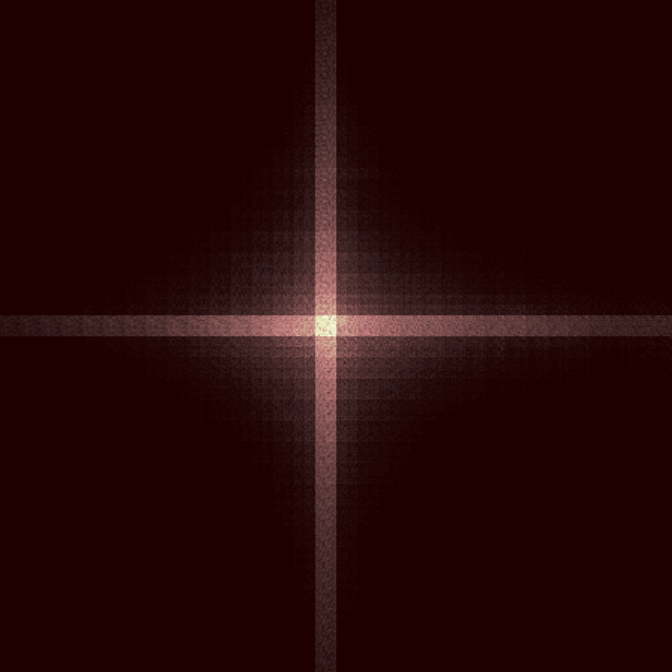}
    \subcaption{Block 13}
    \label{figure:erf_sequencer_448:12}
  \end{minipage}
  \begin{minipage}[t]{0.13\hsize}
    \centering
    \includegraphics[width=\linewidth]{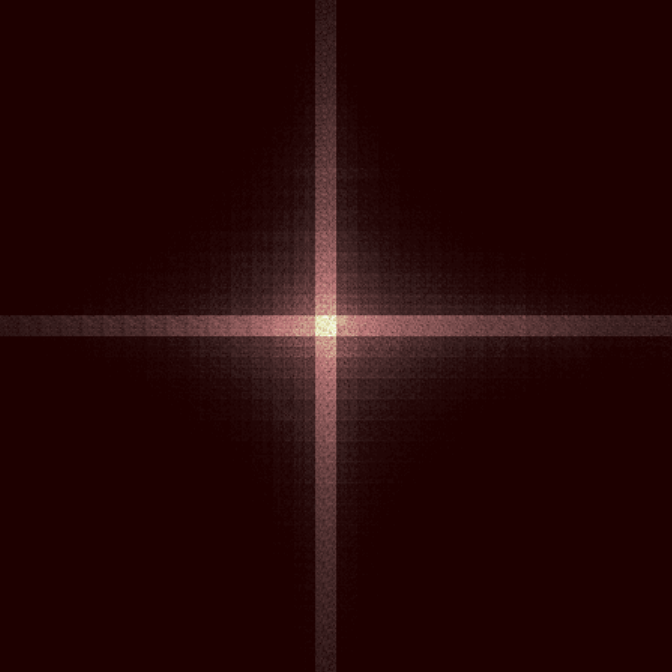}
    \subcaption{Block 14}
    \label{figure:erf_sequencer_448:13}
  \end{minipage}
  \begin{minipage}[t]{0.13\hsize}
    \centering
    \includegraphics[width=\linewidth]{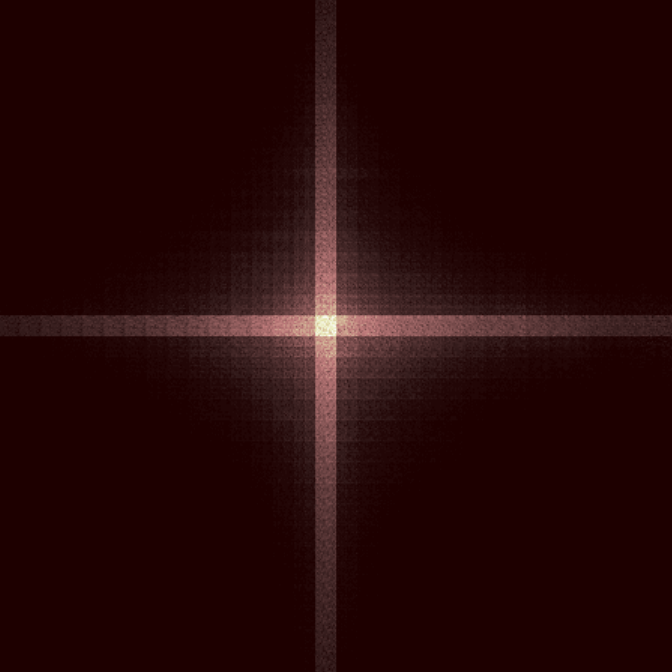}
    \subcaption{Block 15}
    \label{figure:erf_sequencer_448:14}
  \end{minipage}
  \begin{minipage}[t]{0.13\hsize}
    \centering
    \includegraphics[width=\linewidth]{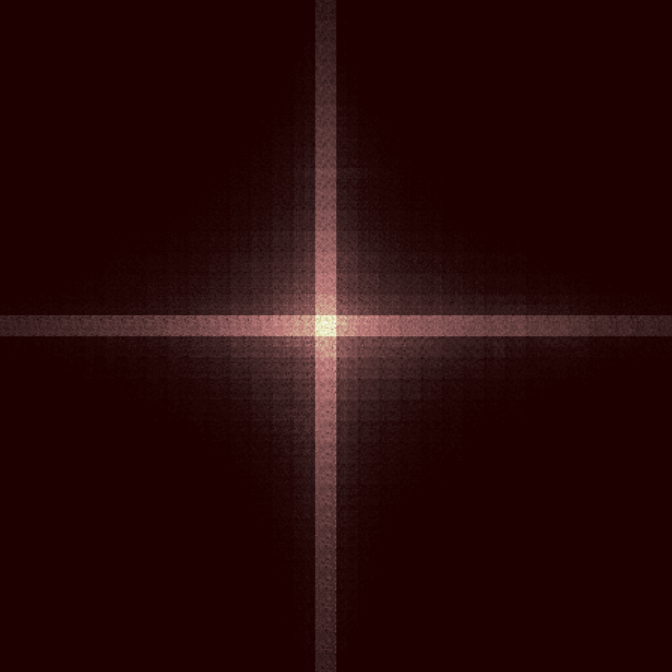}
    \subcaption{Block 16}
    \label{figure:erf_sequencer_448:15}
  \end{minipage}
  \begin{minipage}[t]{0.13\hsize}
    \centering
    \includegraphics[width=\linewidth]{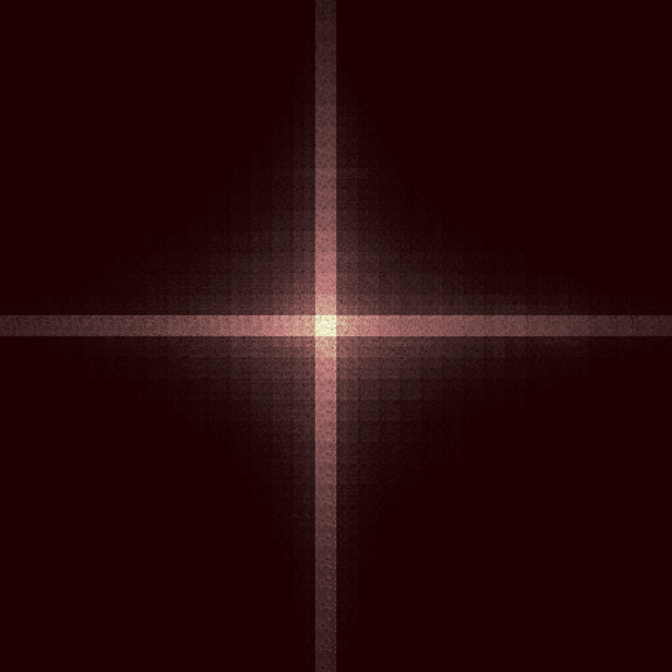}
    \subcaption{Block 17}
    \label{figure:erf_sequencer_448:16}
  \end{minipage}
  \begin{minipage}[t]{0.13\hsize}
    \centering
    \includegraphics[width=\linewidth]{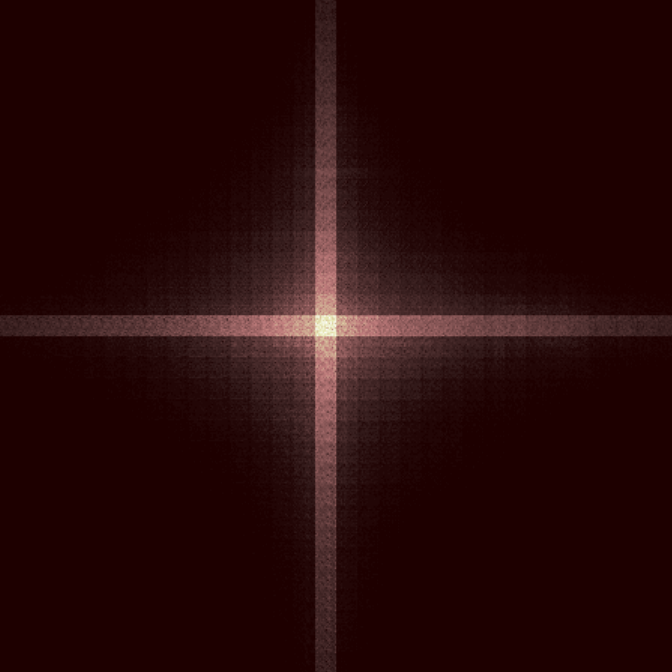}
    \subcaption{Block 18}
    \label{figure:erf_sequencer_448:17}
  \end{minipage}
  \caption{ERFs in Sequencer2D-S on images with resolution $448^2$.}
    \label{figure:erf_sequencer_448}
\end{figure}

%% file: erf_resnet_448.tex
\begin{figure}[tb]
  \raggedright
  \begin{minipage}[t]{0.13\hsize}
    \centering
    \includegraphics[width=\linewidth]{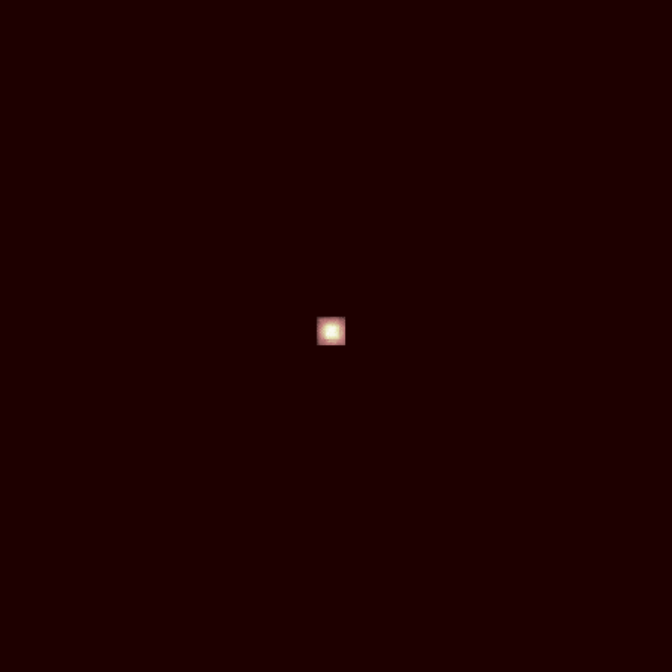}
    \subcaption{Block 1}
    \label{figure:erf_resnet_448:0}
  \end{minipage}
  \begin{minipage}[t]{0.13\hsize}
    \centering
    \includegraphics[width=\linewidth]{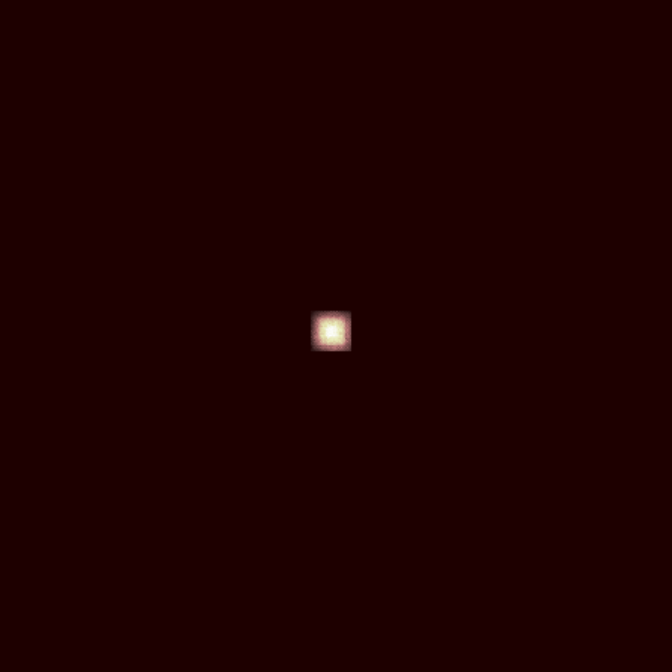}
    \subcaption{Block 2}
    \label{figure:erf_resnet_448:1}
  \end{minipage}
  \begin{minipage}[t]{0.13\hsize}
    \centering
    \includegraphics[width=\linewidth]{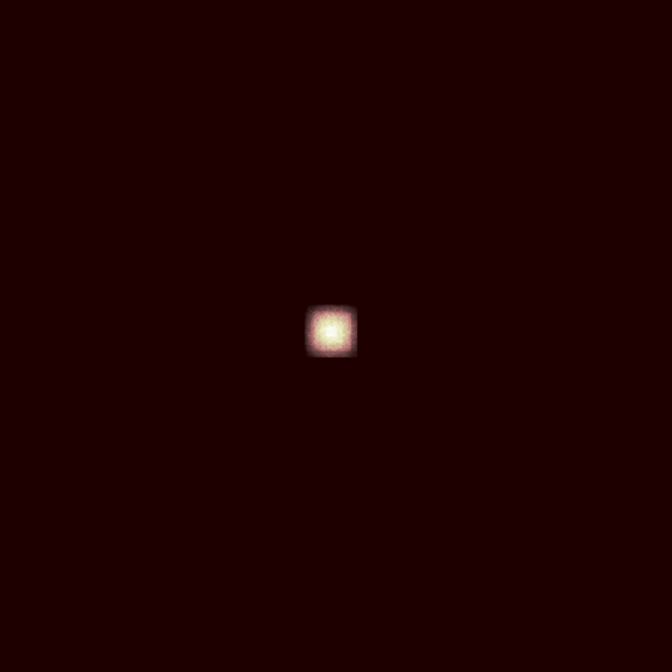}
    \subcaption{Block 3}
    \label{figure:erf_resnet_448:2}
  \end{minipage}
  \begin{minipage}[t]{0.13\hsize}
    \centering
    \includegraphics[width=\linewidth]{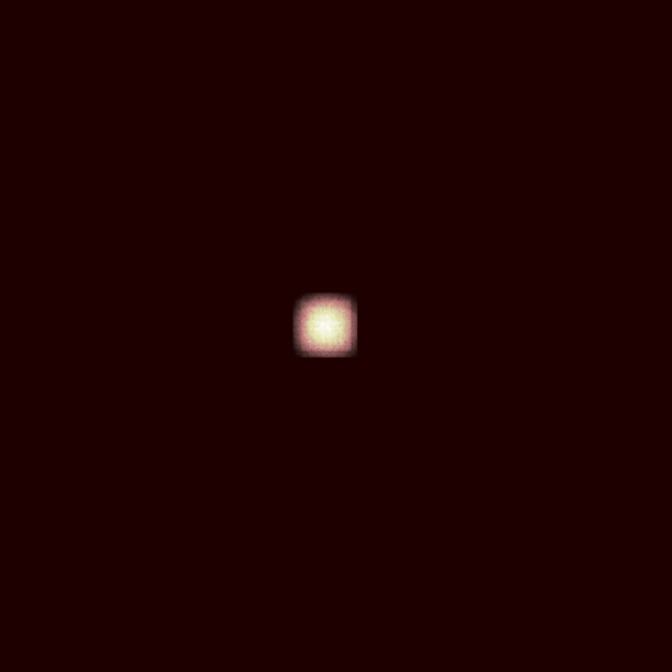}
    \subcaption{Block 4}
    \label{figure:erf_resnet_448:3}
  \end{minipage}
  \begin{minipage}[t]{0.13\hsize}
    \centering
    \includegraphics[width=\linewidth]{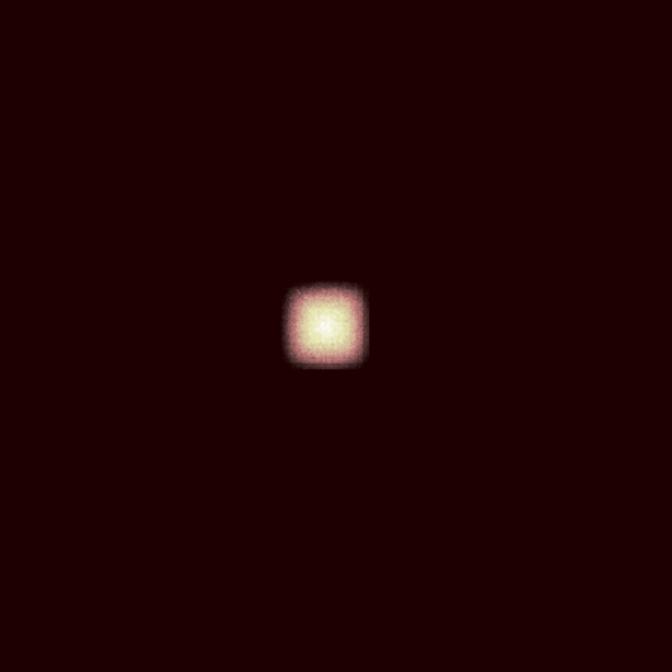}
    \subcaption{Block 5}
    \label{figure:erf_resnet_448:4}
  \end{minipage}
  \begin{minipage}[t]{0.13\hsize}
    \centering
    \includegraphics[width=\linewidth]{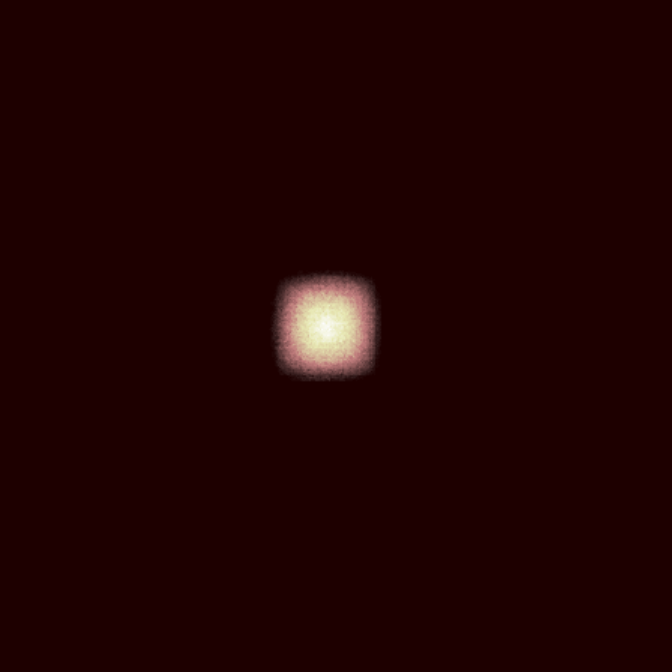}
    \subcaption{Block 6}
    \label{figure:erf_resnet_448:5}
  \end{minipage}
  \begin{minipage}[t]{0.13\hsize}
    \centering
    \includegraphics[width=\linewidth]{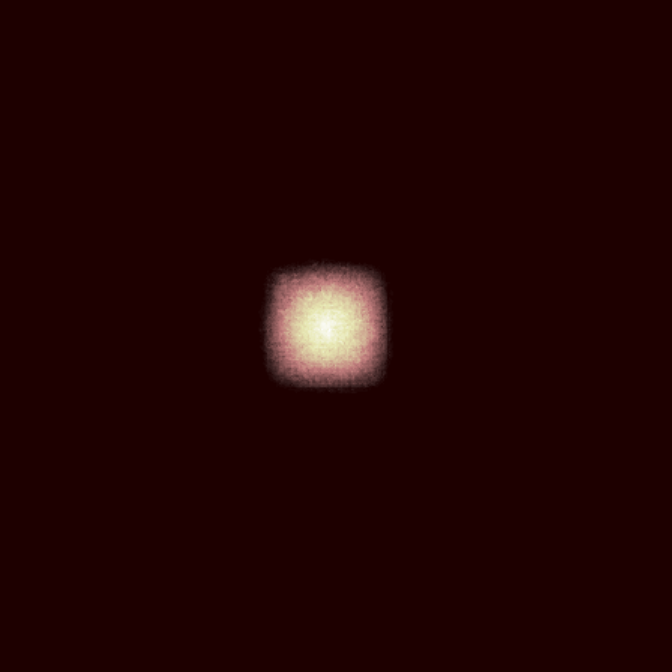}
    \subcaption{Block 7}
    \label{figure:erf_resnet_448:6}
  \end{minipage}
  \begin{minipage}[t]{0.13\hsize}
    \centering
    \includegraphics[width=\linewidth]{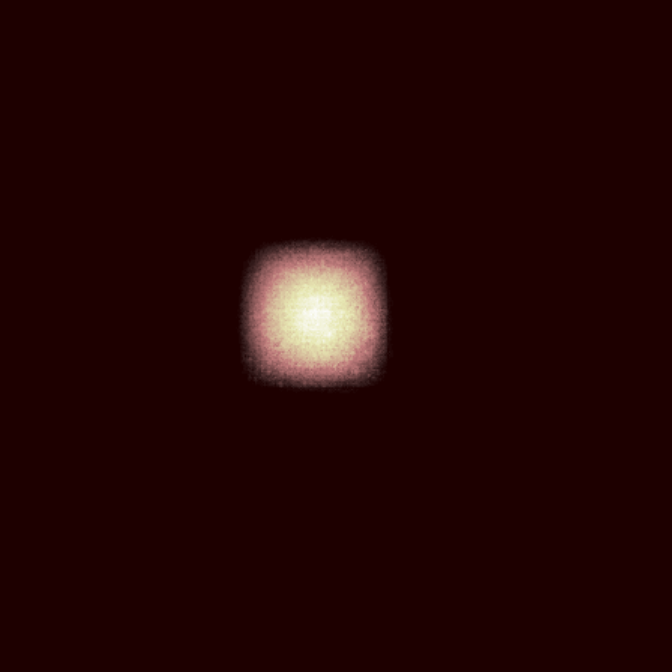}
    \subcaption{Block 8}
    \label{figure:erf_resnet_448:7}
  \end{minipage}
  \begin{minipage}[t]{0.13\hsize}
    \centering
    \includegraphics[width=\linewidth]{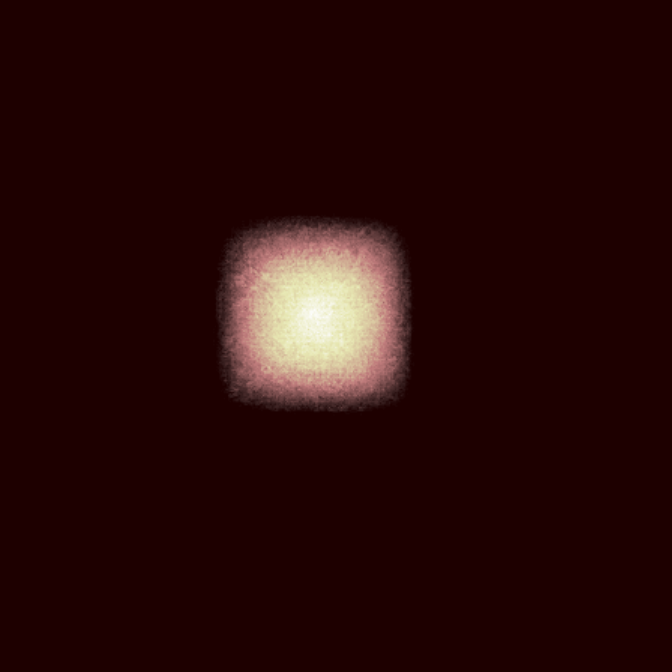}
    \subcaption{Block 9}
    \label{figure:erf_resnet_448:8}
  \end{minipage}
  \begin{minipage}[t]{0.13\hsize}
    \centering
    \includegraphics[width=\linewidth]{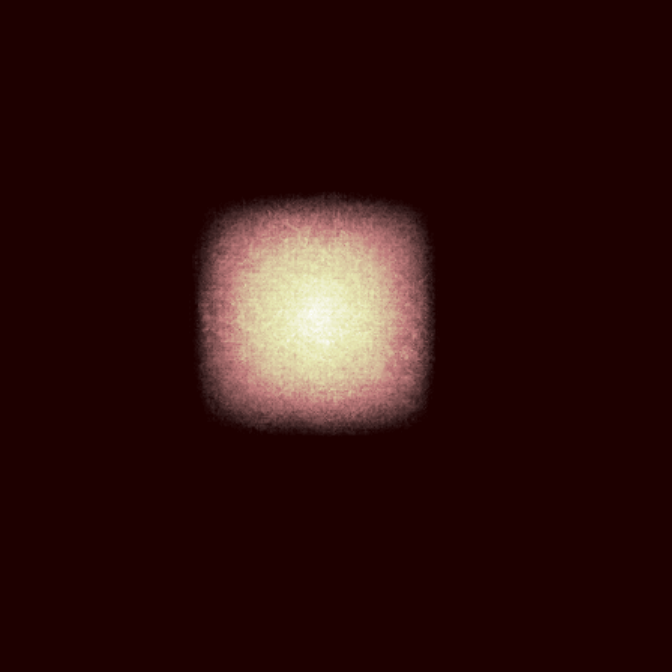}
    \subcaption{Block 10}
    \label{figure:erf_resnet_448:9}
  \end{minipage}
  \begin{minipage}[t]{0.13\hsize}
    \centering
    \includegraphics[width=\linewidth]{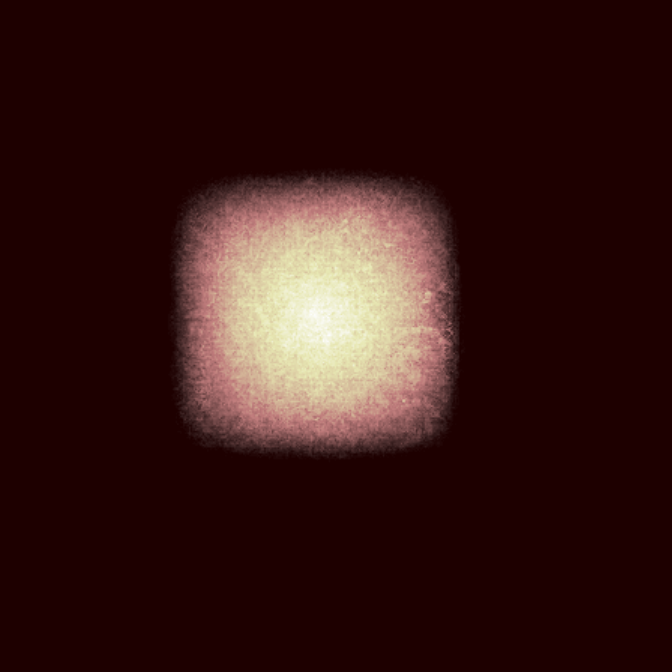}
    \subcaption{Block 11}
    \label{figure:erf_resnet_448:10}
  \end{minipage}
  \begin{minipage}[t]{0.13\hsize}
    \centering
    \includegraphics[width=\linewidth]{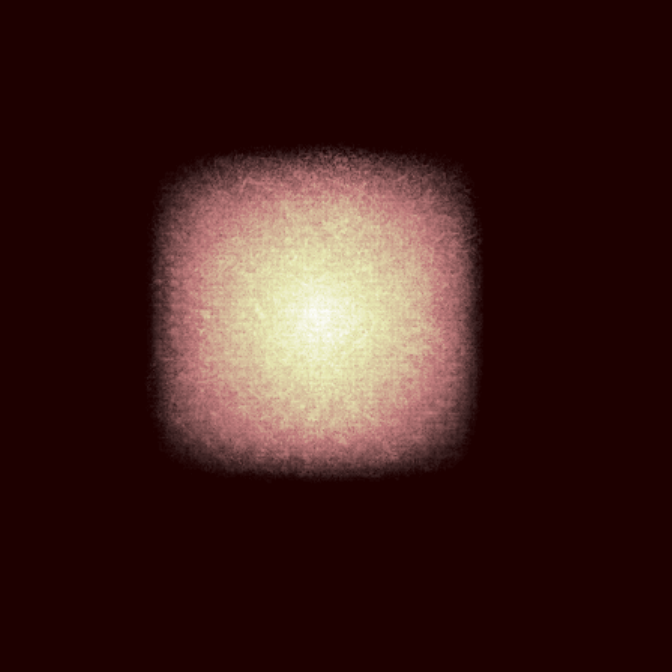}
    \subcaption{Block 12}
    \label{figure:erf_resnet_448:11}
  \end{minipage}
  \begin{minipage}[t]{0.13\hsize}
    \centering
    \includegraphics[width=\linewidth]{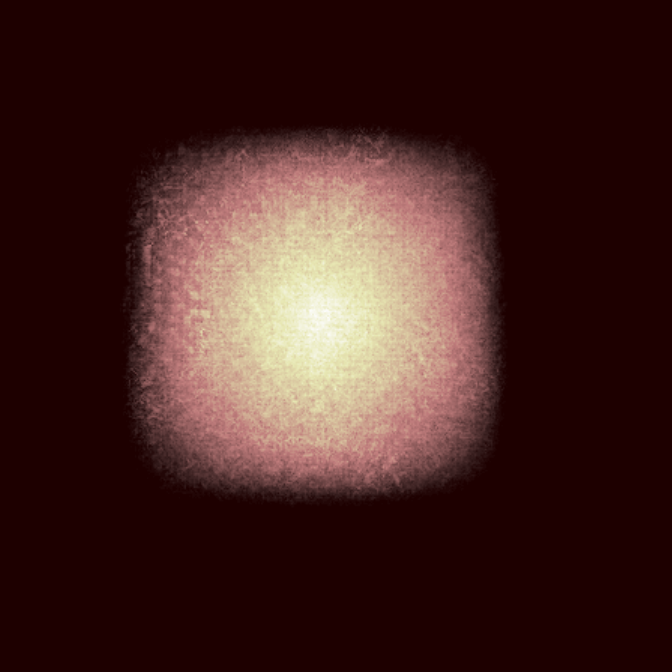}
    \subcaption{Block 13}
    \label{figure:erf_resnet_448:12}
  \end{minipage}
  \begin{minipage}[t]{0.13\hsize}
    \centering
    \includegraphics[width=\linewidth]{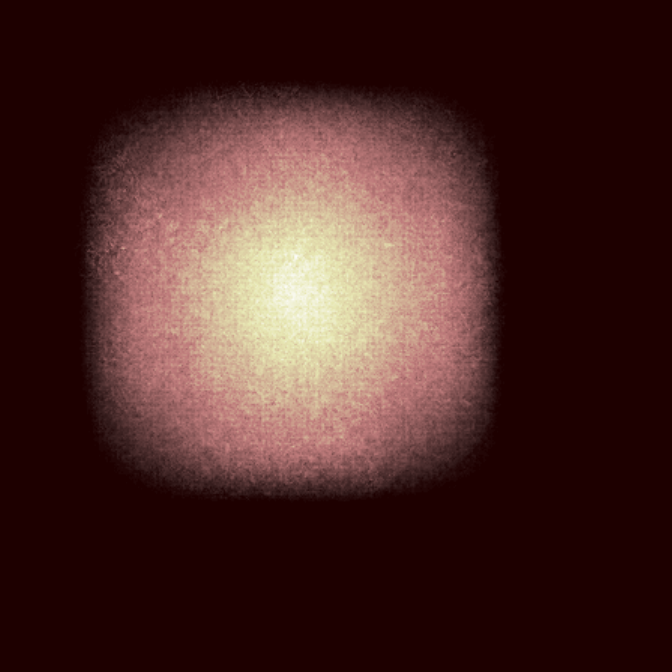}
    \subcaption{Block 14}
    \label{figure:erf_resnet_448:13}
  \end{minipage}
  \begin{minipage}[t]{0.13\hsize}
    \centering
    \includegraphics[width=\linewidth]{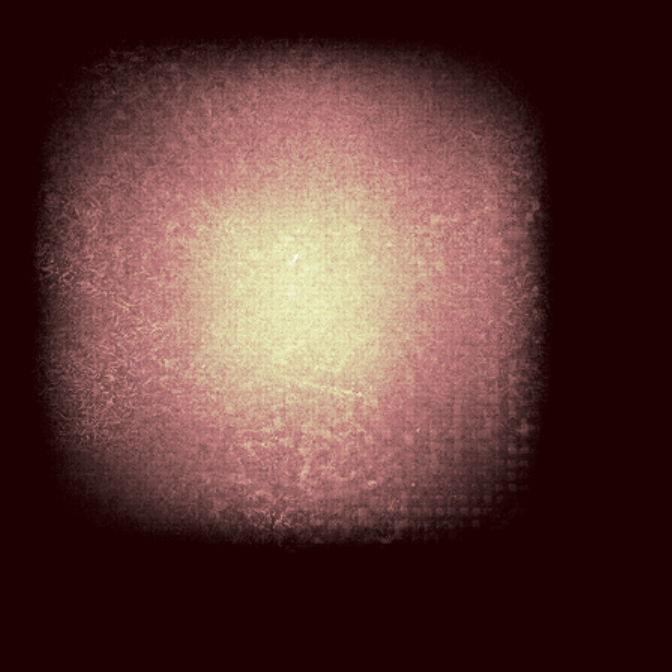}
    \subcaption{Block 15}
    \label{figure:erf_resnet_448:14}
  \end{minipage}
  \begin{minipage}[t]{0.13\hsize}
    \centering
    \includegraphics[width=\linewidth]{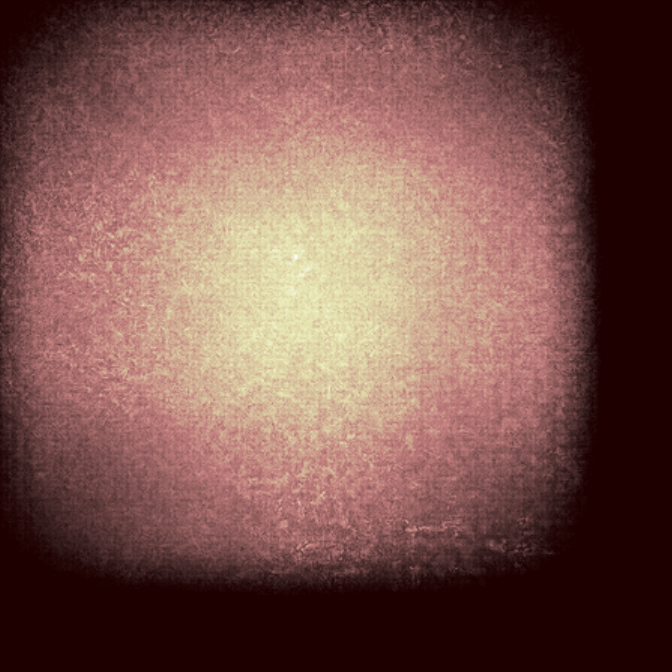}
    \subcaption{Block 16}
    \label{figure:erf_resnet_448:15}
  \end{minipage}
  \caption{ERFs in ResNet-50 \cite{he2016deep} on images with resolution $448^2$.}
  \label{figure:erf_resnet_448}
\end{figure}

%% file: erf_convnext_448.tex
\begin{figure}[tb]
  \raggedright
  \begin{minipage}[t]{0.13\hsize}
    \centering
    \includegraphics[width=\linewidth]{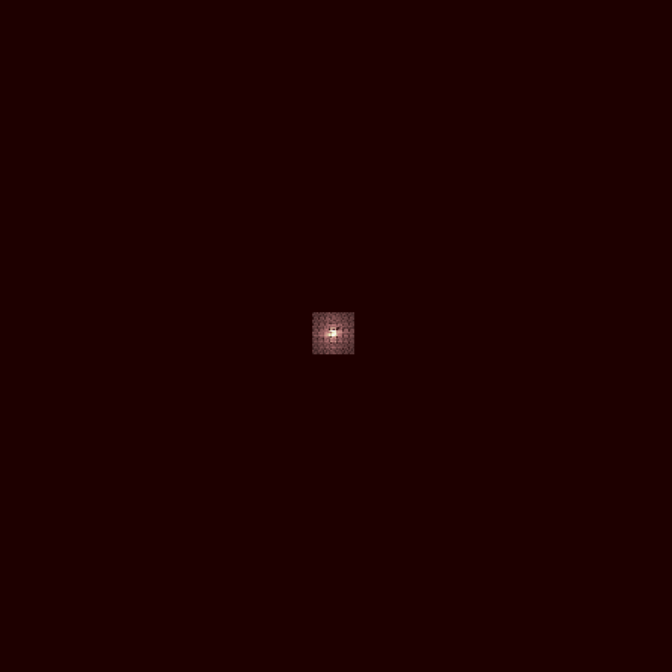}
    \subcaption{Block 1}
    \label{figure:erf_convnext_448:0}
  \end{minipage}
  \begin{minipage}[t]{0.13\hsize}
    \centering
    \includegraphics[width=\linewidth]{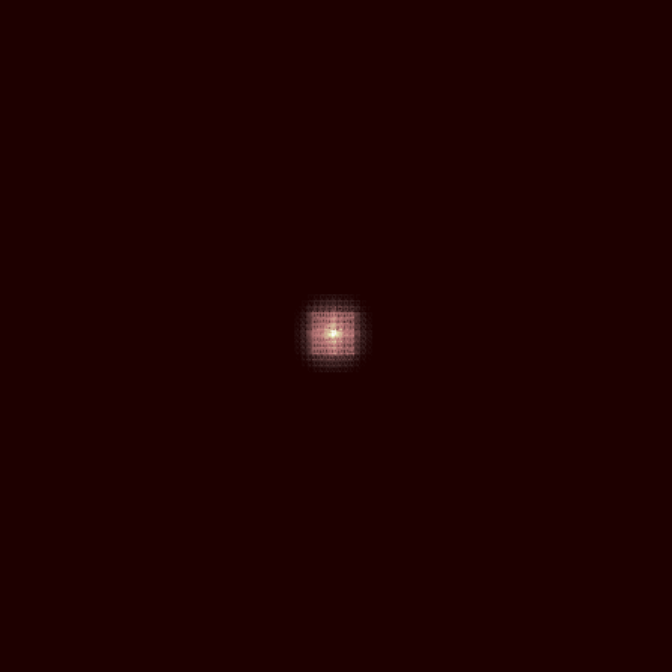}
    \subcaption{Block 2}
    \label{figure:erf_convnext_448:1}
  \end{minipage}
  \begin{minipage}[t]{0.13\hsize}
    \centering
    \includegraphics[width=\linewidth]{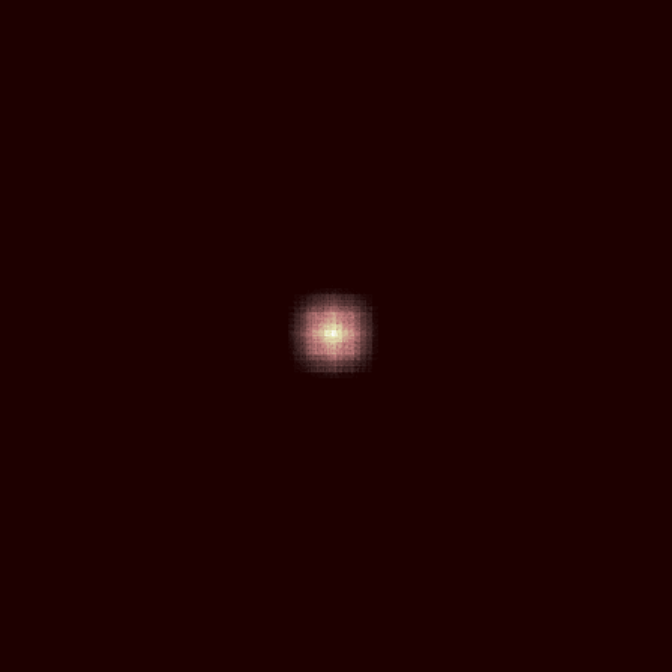}
    \subcaption{Block 3}
    \label{figure:erf_convnext_448:2}
  \end{minipage}
  \begin{minipage}[t]{0.13\hsize}
    \centering
    \includegraphics[width=\linewidth]{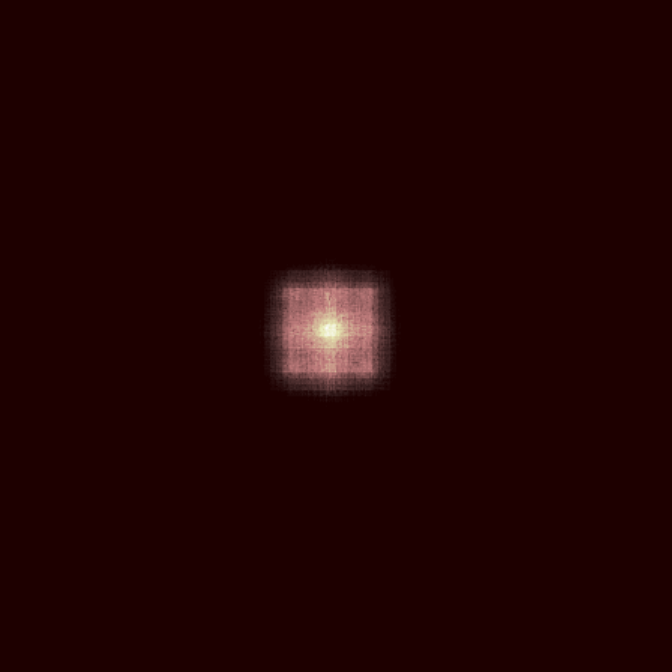}
    \subcaption{Block 4}
    \label{figure:erf_convnext_448:3}
  \end{minipage}
  \begin{minipage}[t]{0.13\hsize}
    \centering
    \includegraphics[width=\linewidth]{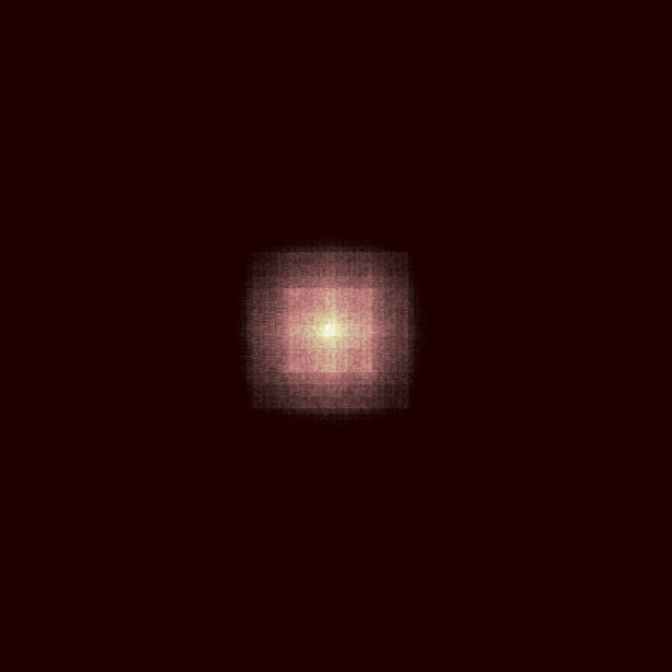}
    \subcaption{Block 5}
    \label{figure:erf_convnext_448:4}
  \end{minipage}
  \begin{minipage}[t]{0.13\hsize}
    \centering
    \includegraphics[width=\linewidth]{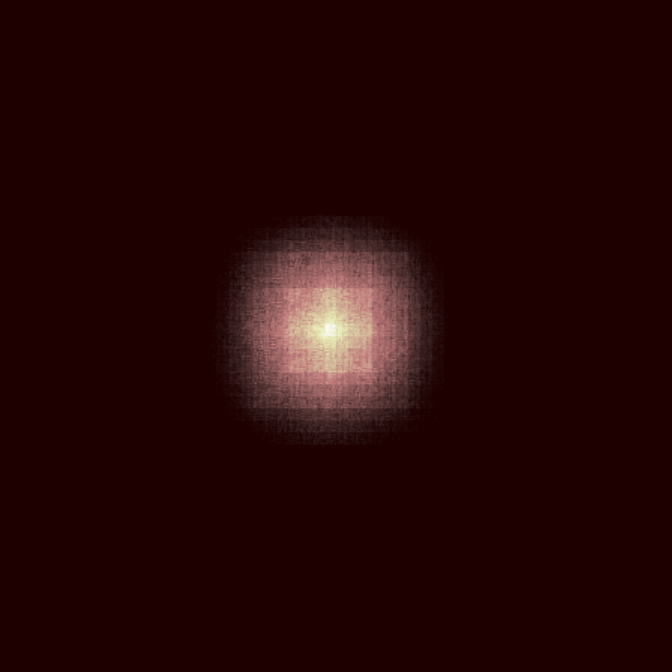}
    \subcaption{Block 6}
    \label{figure:erf_convnext_448:5}
  \end{minipage}
  \begin{minipage}[t]{0.13\hsize}
    \centering
    \includegraphics[width=\linewidth]{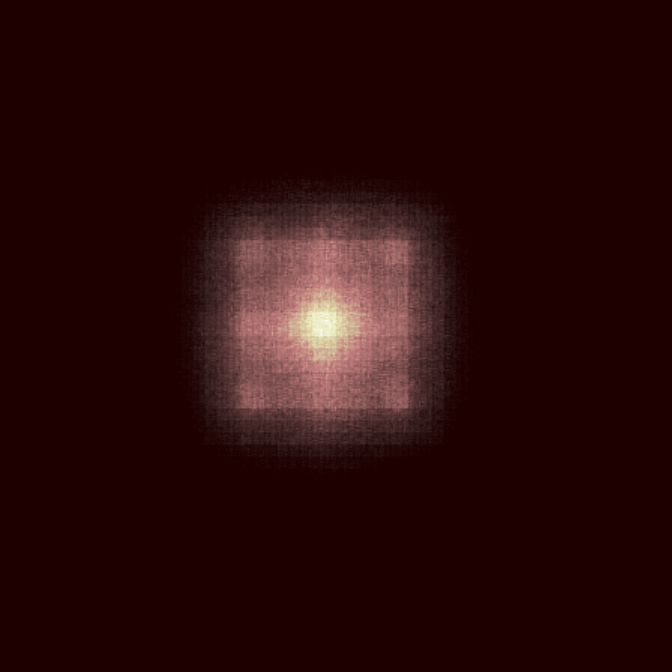}
    \subcaption{Block 7}
    \label{figure:erf_convnext_448:6}
  \end{minipage}
  \begin{minipage}[t]{0.13\hsize}
    \centering
    \includegraphics[width=\linewidth]{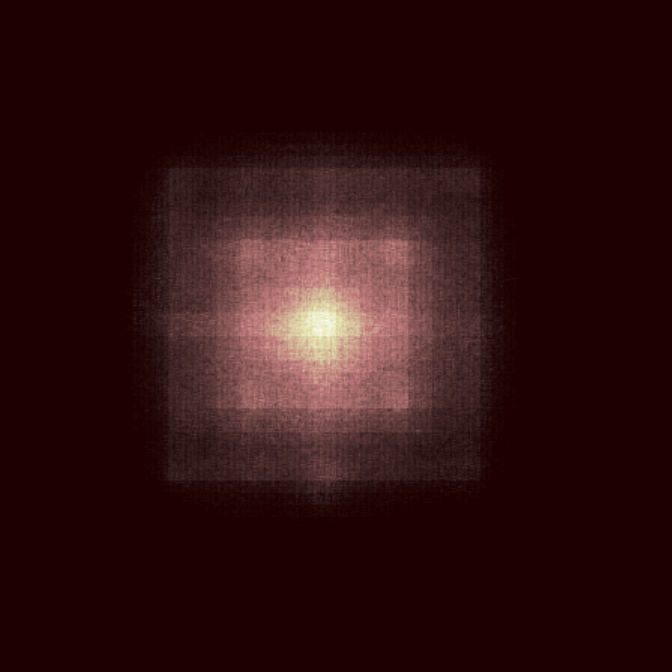}
    \subcaption{Block 8}
    \label{figure:erf_convnext_448:7}
  \end{minipage}
  \begin{minipage}[t]{0.13\hsize}
    \centering
    \includegraphics[width=\linewidth]{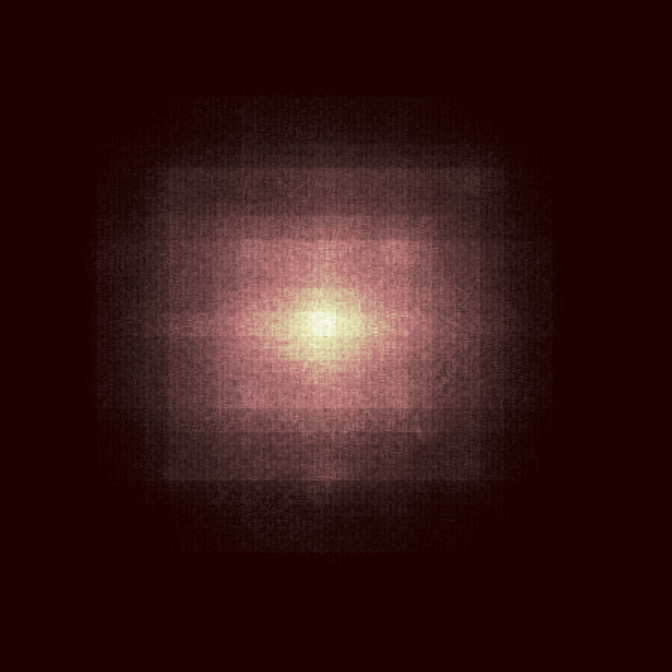}
    \subcaption{Block 9}
    \label{figure:erf_convnext_448:8}
  \end{minipage}
  \begin{minipage}[t]{0.13\hsize}
    \centering
    \includegraphics[width=\linewidth]{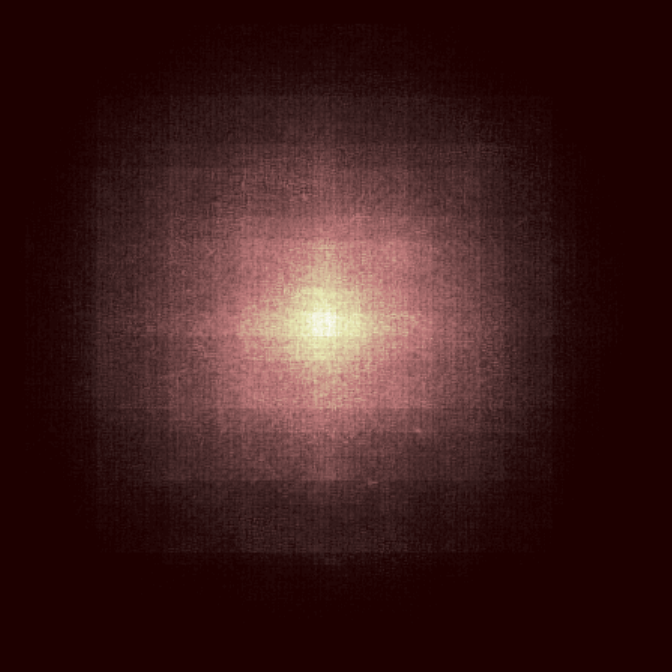}
    \subcaption{Block 10}
    \label{figure:erf_convnext_448:9}
  \end{minipage}
  \begin{minipage}[t]{0.13\hsize}
    \centering
    \includegraphics[width=\linewidth]{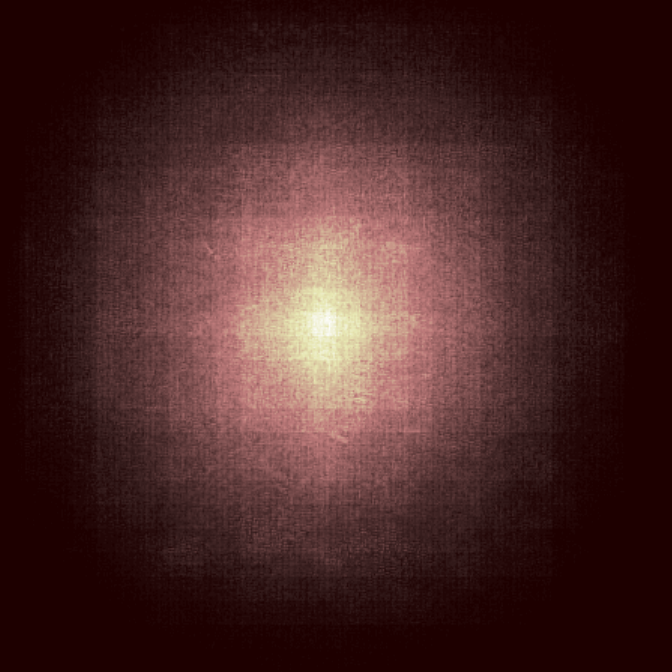}
    \subcaption{Block 11}
    \label{figure:erf_convnext_448:10}
  \end{minipage}
  \begin{minipage}[t]{0.13\hsize}
    \centering
    \includegraphics[width=\linewidth]{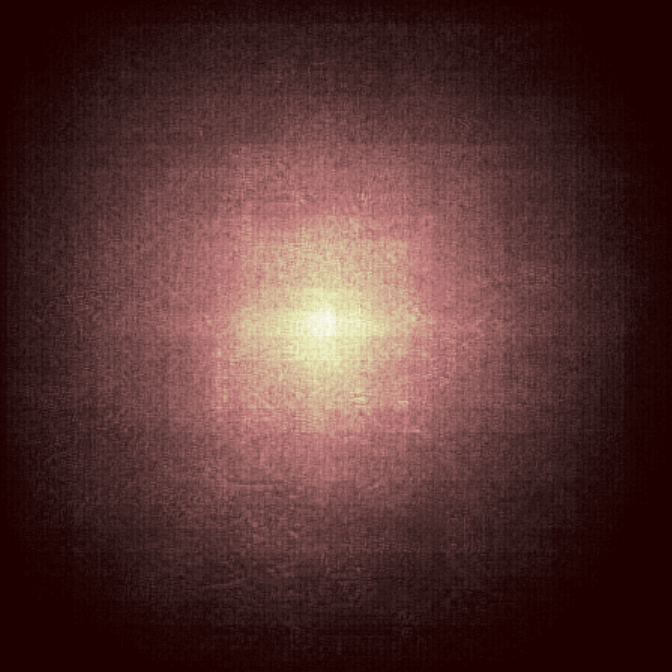}
    \subcaption{Block 12}
    \label{figure:erf_convnext_448:11}
  \end{minipage}
  \begin{minipage}[t]{0.13\hsize}
    \centering
    \includegraphics[width=\linewidth]{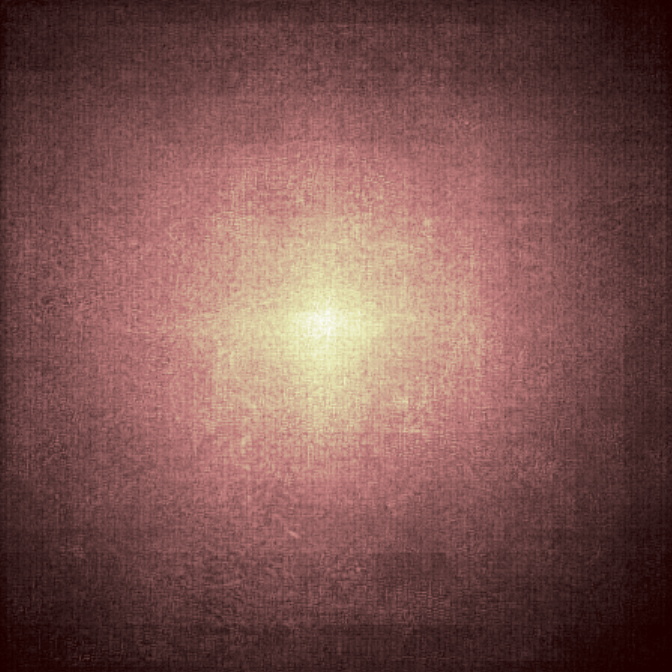}
    \subcaption{Block 13}
    \label{figure:erf_convnext_448:12}
  \end{minipage}
  \begin{minipage}[t]{0.13\hsize}
    \centering
    \includegraphics[width=\linewidth]{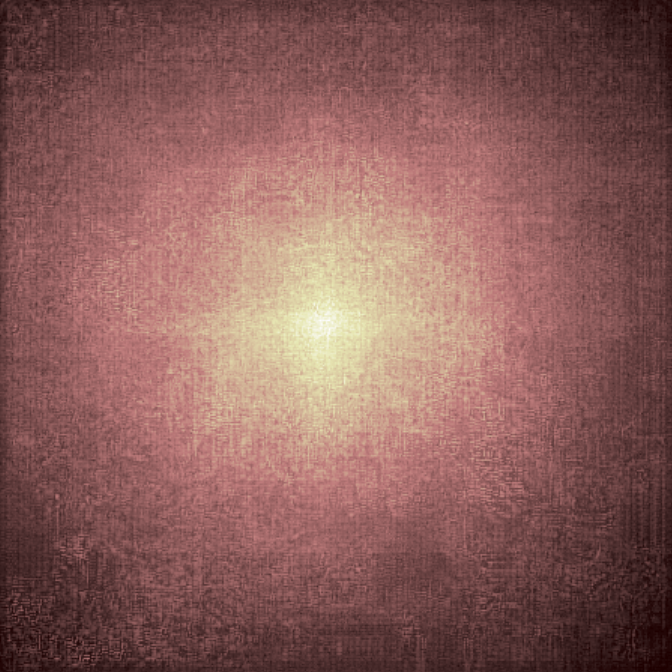}
    \subcaption{Block 14}
    \label{figure:erf_convnext_448:13}
  \end{minipage}
  \begin{minipage}[t]{0.13\hsize}
    \centering
    \includegraphics[width=\linewidth]{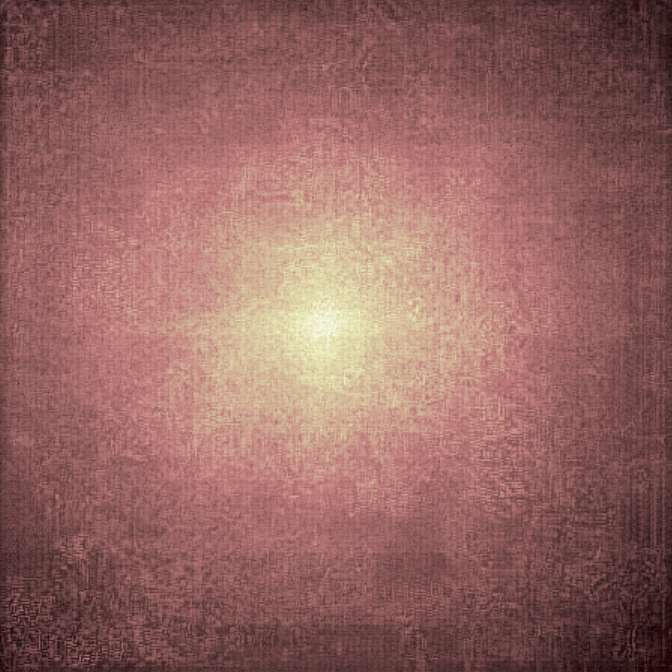}
    \subcaption{Block 15}
    \label{figure:erf_convnext_448:14}
  \end{minipage}
  \begin{minipage}[t]{0.13\hsize}
    \centering
    \includegraphics[width=\linewidth]{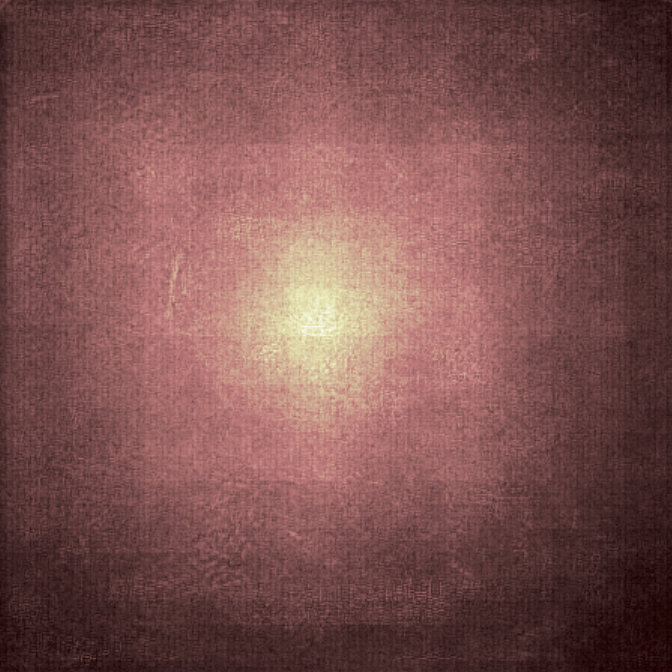}
    \subcaption{Block 16}
    \label{figure:erf_convnext_448:15}
  \end{minipage}
  \begin{minipage}[t]{0.13\hsize}
    \centering
    \includegraphics[width=\linewidth]{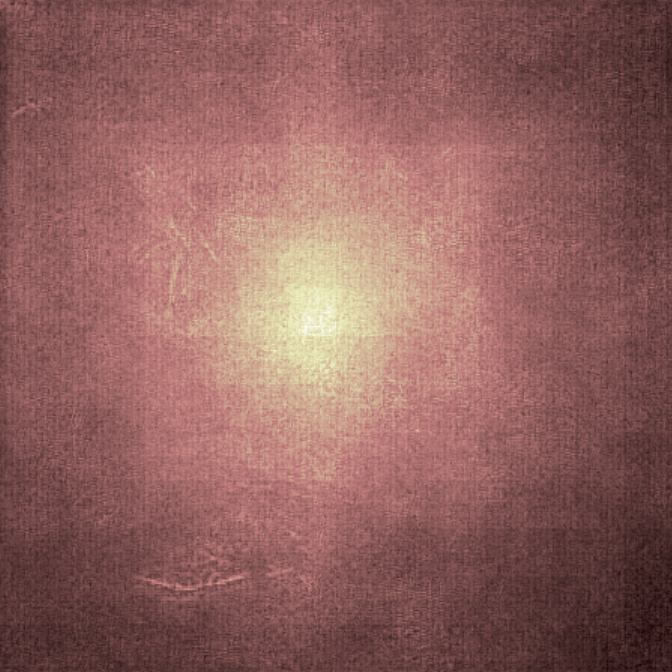}
    \subcaption{Block 17}
    \label{figure:erf_convnext_448:16}
  \end{minipage}
  \begin{minipage}[t]{0.13\hsize}
    \centering
    \includegraphics[width=\linewidth]{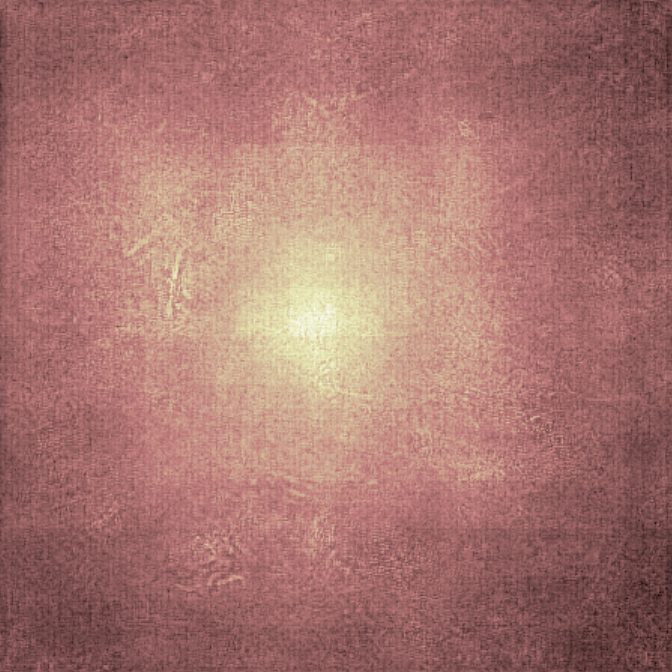}
    \subcaption{Block 18}
    \label{figure:erf_convnext_448:17}
  \end{minipage}
  \caption{ERFs in ConvNeXt-T \cite{liu2022convnet} on images with resolution $448^2$.}
    \label{figure:erf_convnext_448}
\end{figure}

%% file: erf_cyclemlp_448.tex
\begin{figure}[tb]
  \raggedright
  \begin{minipage}[t]{0.13\hsize}
    \centering
    \includegraphics[width=\linewidth]{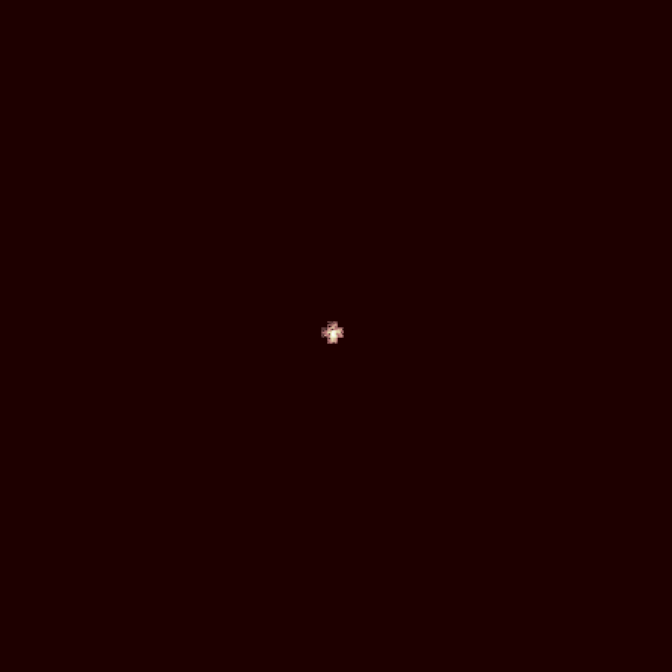}
    \subcaption{Block 1}
    \label{figure:erf_cyclemlp_448:0}
  \end{minipage}
  \begin{minipage}[t]{0.13\hsize}
    \centering
    \includegraphics[width=\linewidth]{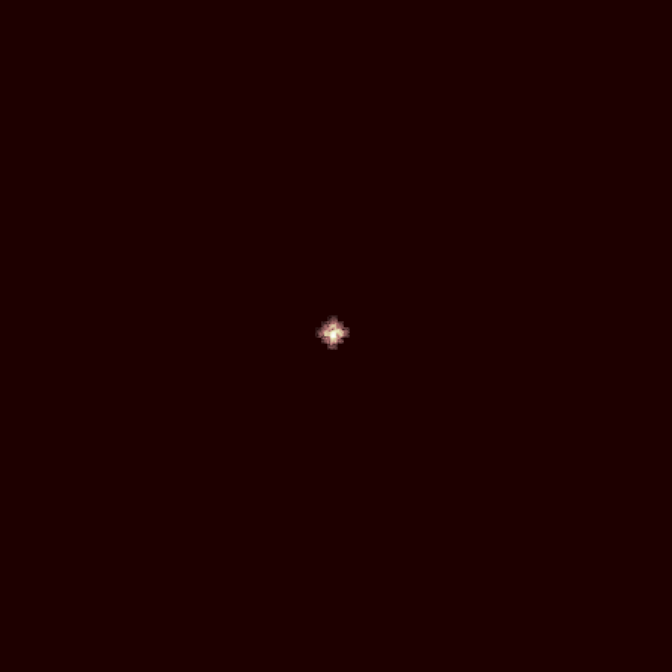}
    \subcaption{Block 2}
    \label{figure:erf_cyclemlp_448:1}
  \end{minipage}
  \begin{minipage}[t]{0.13\hsize}
    \centering
    \includegraphics[width=\linewidth]{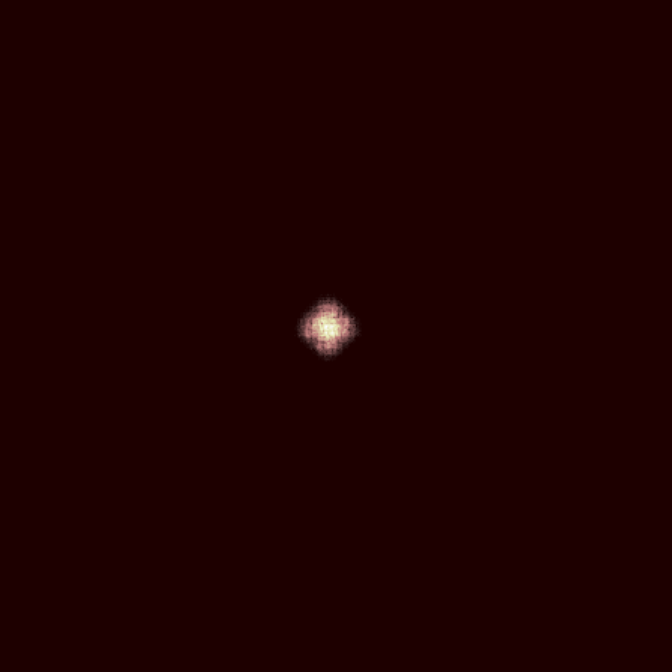}
    \subcaption{Block 3}
    \label{figure:erf_cyclemlp_448:2}
  \end{minipage}
  \begin{minipage}[t]{0.13\hsize}
    \centering
    \includegraphics[width=\linewidth]{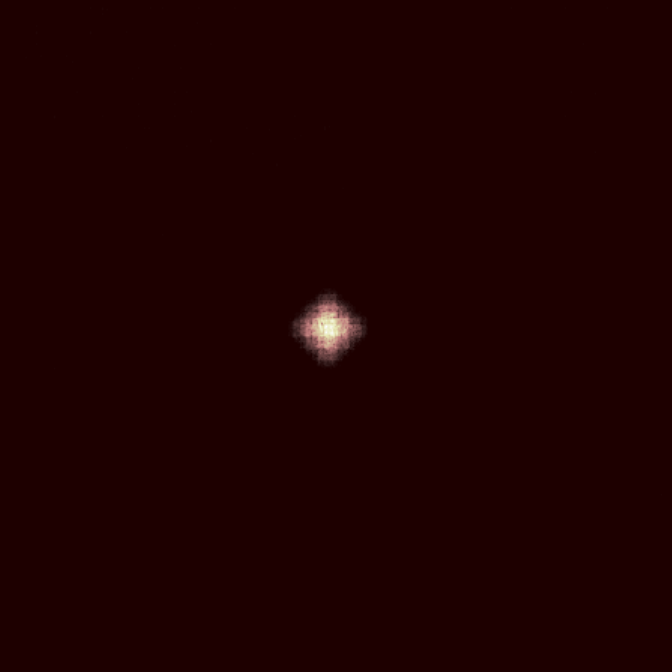}
    \subcaption{Block 4}
    \label{figure:erf_cyclemlp_448:3}
  \end{minipage}
  \begin{minipage}[t]{0.13\hsize}
    \centering
    \includegraphics[width=\linewidth]{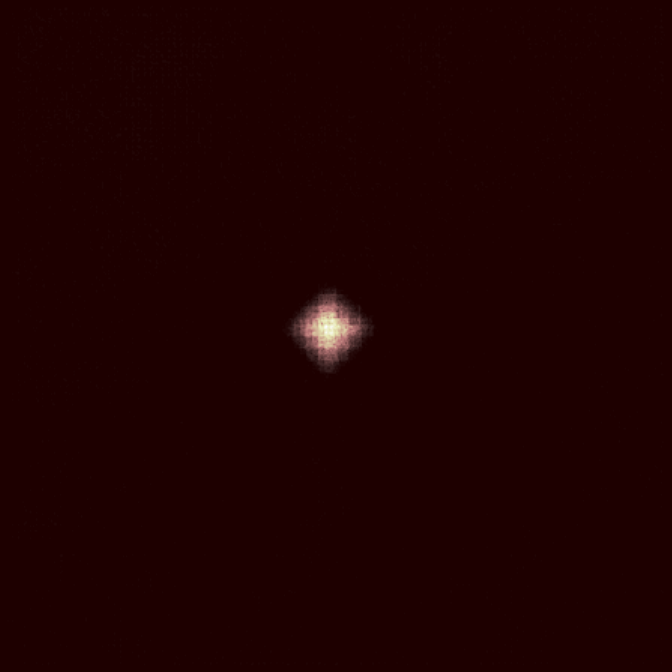}
    \subcaption{Block 5}
    \label{figure:erf_cyclemlp_448:4}
  \end{minipage}
  \begin{minipage}[t]{0.13\hsize}
    \centering
    \includegraphics[width=\linewidth]{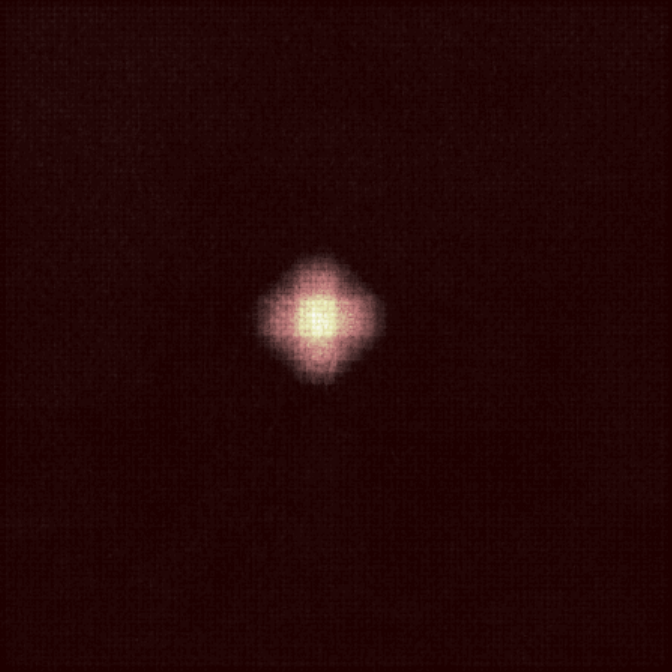}
    \subcaption{Block 6}
    \label{figure:erf_cyclemlp_448:5}
  \end{minipage}
  \begin{minipage}[t]{0.13\hsize}
    \centering
    \includegraphics[width=\linewidth]{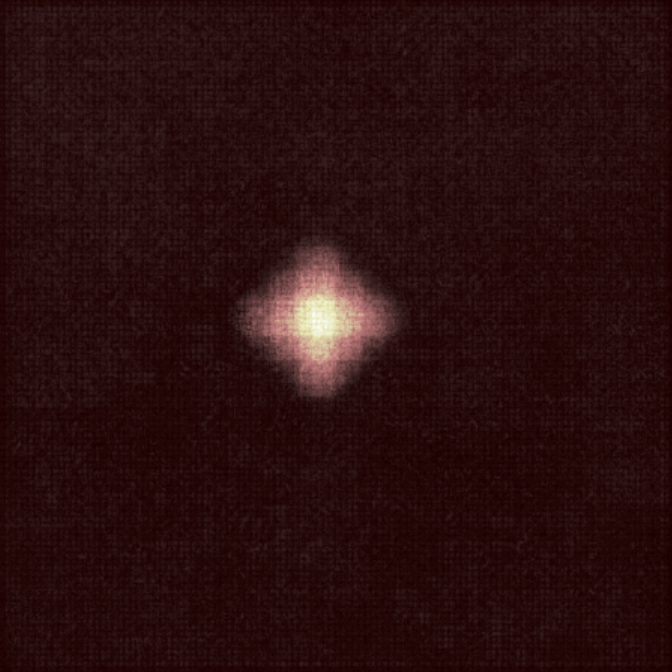}
    \subcaption{Block 7}
    \label{figure:erf_cyclemlp_448:6}
  \end{minipage}
  \begin{minipage}[t]{0.13\hsize}
    \centering
    \includegraphics[width=\linewidth]{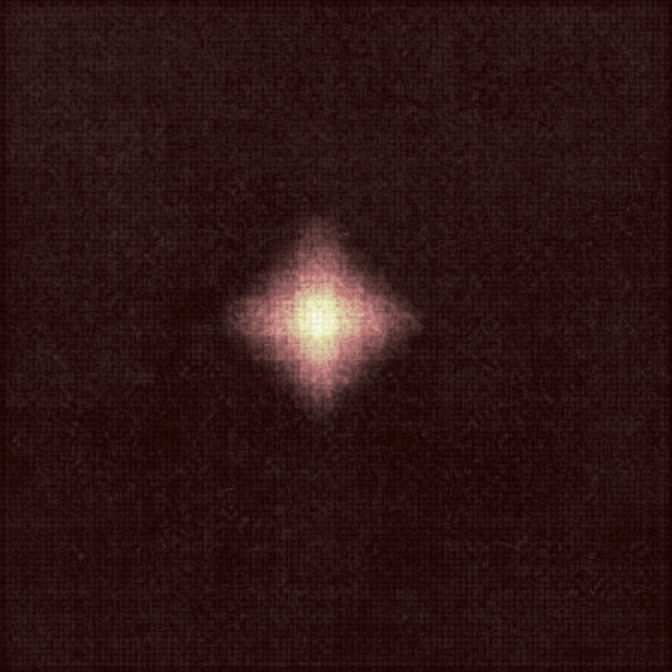}
    \subcaption{Block 8}
    \label{figure:erf_cyclemlp_448:7}
  \end{minipage}
  \begin{minipage}[t]{0.13\hsize}
    \centering
    \includegraphics[width=\linewidth]{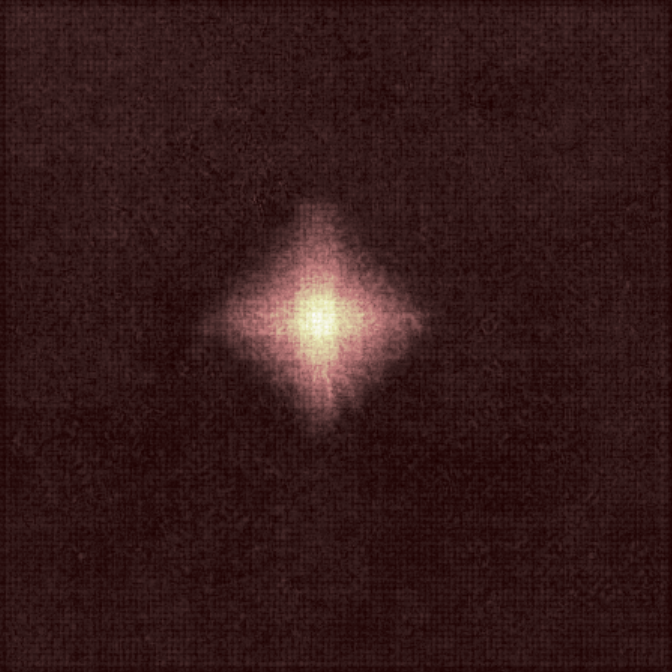}
    \subcaption{Block 9}
    \label{figure:erf_cyclemlp_448:8}
  \end{minipage}
  \begin{minipage}[t]{0.13\hsize}
    \centering
    \includegraphics[width=\linewidth]{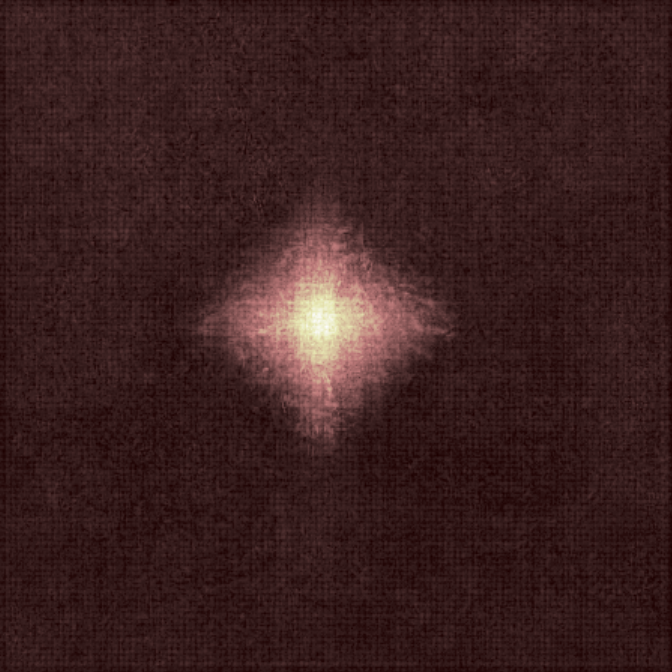}
    \subcaption{Block 10}
    \label{figure:erf_cyclemlp_448:9}
  \end{minipage}
  \begin{minipage}[t]{0.13\hsize}
    \centering
    \includegraphics[width=\linewidth]{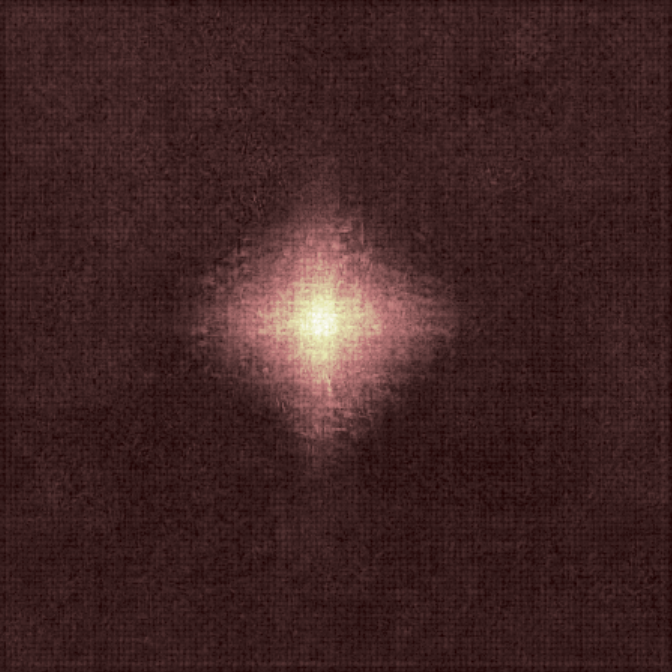}
    \subcaption{Block 11}
    \label{figure:erf_cyclemlp_448:10}
  \end{minipage}
  \begin{minipage}[t]{0.13\hsize}
    \centering
    \includegraphics[width=\linewidth]{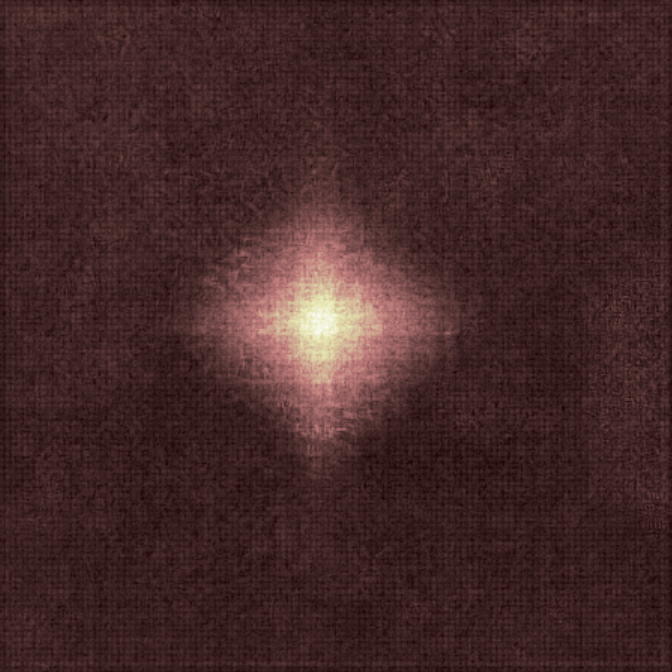}
    \subcaption{Block 12}
    \label{figure:erf_cyclemlp_448:11}
  \end{minipage}
  \begin{minipage}[t]{0.13\hsize}
    \centering
    \includegraphics[width=\linewidth]{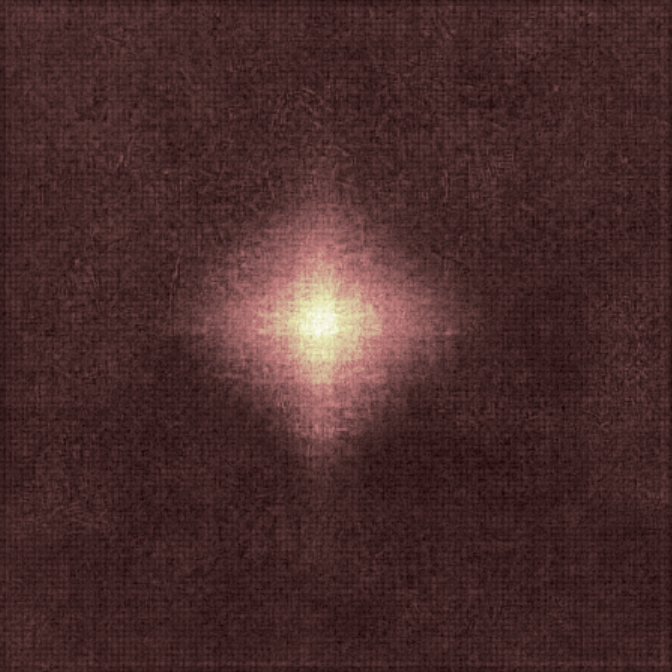}
    \subcaption{Block 13}
    \label{figure:erf_cyclemlp_448:12}
  \end{minipage}
  \begin{minipage}[t]{0.13\hsize}
    \centering
    \includegraphics[width=\linewidth]{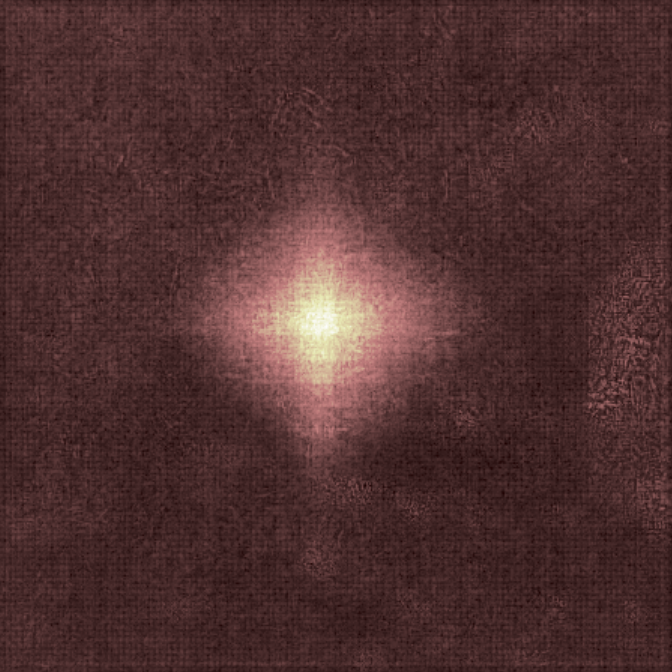}
    \subcaption{Block 14}
    \label{figure:erf_cyclemlp_448:13}
  \end{minipage}
  \begin{minipage}[t]{0.13\hsize}
    \centering
    \includegraphics[width=\linewidth]{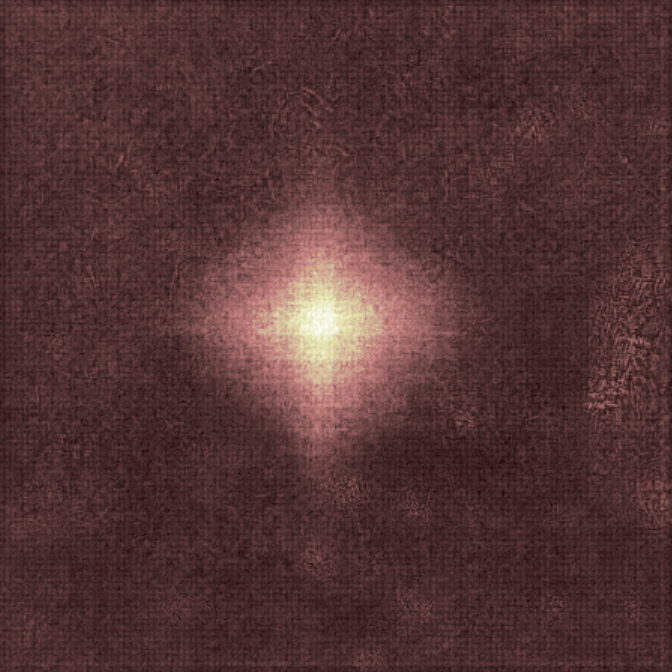}
    \subcaption{Block 15}
    \label{figure:erf_cyclemlp_448:14}
  \end{minipage}
  \begin{minipage}[t]{0.13\hsize}
    \centering
    \includegraphics[width=\linewidth]{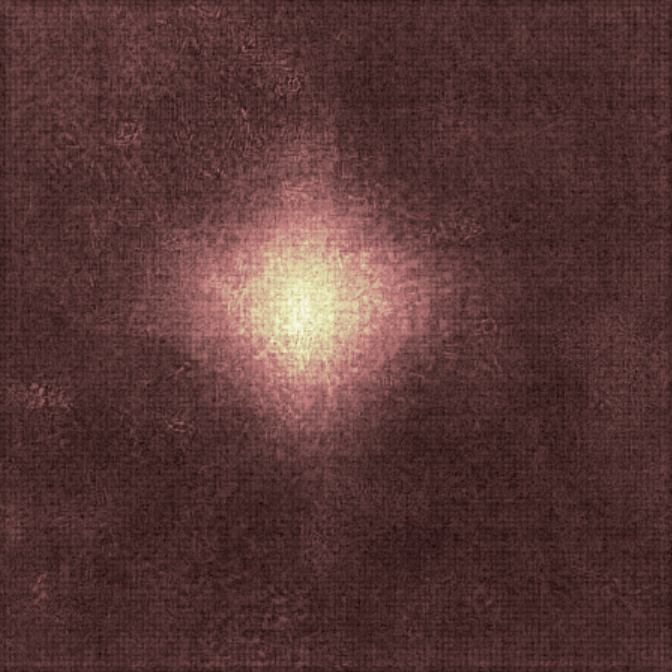}
    \subcaption{Block 16}
    \label{figure:erf_cyclemlp_448:15}
  \end{minipage}
  \begin{minipage}[t]{0.13\hsize}
    \centering
    \includegraphics[width=\linewidth]{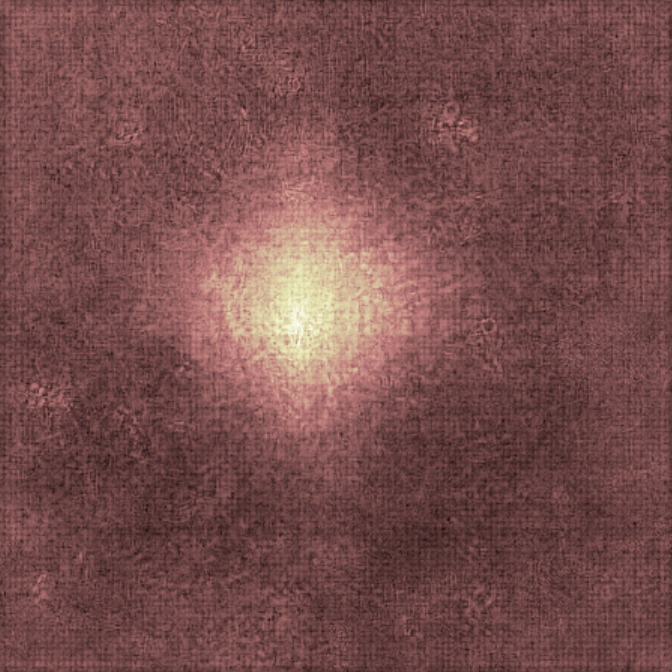}
    \subcaption{Block 17}
    \label{figure:erf_cyclemlp_448:16}
  \end{minipage}
  \begin{minipage}[t]{0.13\hsize}
    \centering
    \includegraphics[width=\linewidth]{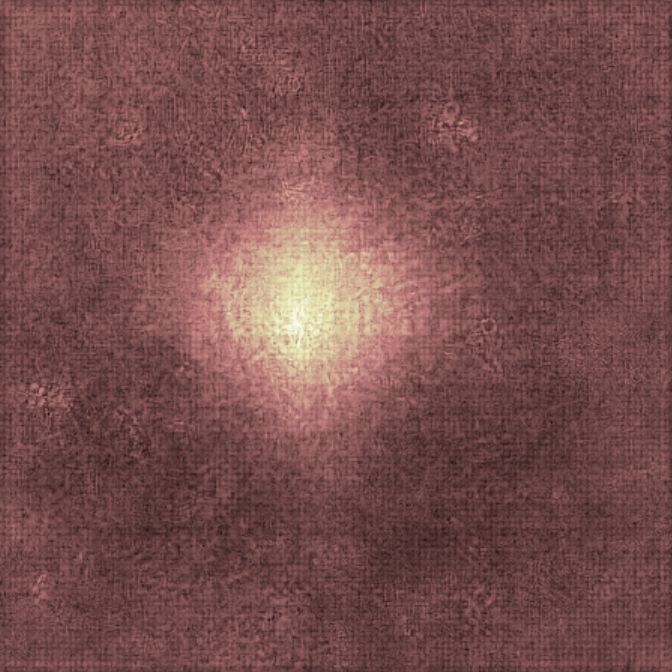}
    \subcaption{Block 18}
    \label{figure:erf_cyclemlp_448:17}
  \end{minipage}
  \caption{ERFs in CycleMLP-B2 \cite{chen2022cyclemlp} on images with resolution $448^2$.}
    \label{figure:erf_cyclemlp_448}
\end{figure}

%% file: erf_deit_448.tex
\begin{figure}[tb]
  \raggedright
  \begin{minipage}[t]{0.13\hsize}
    \centering
    \includegraphics[width=\linewidth]{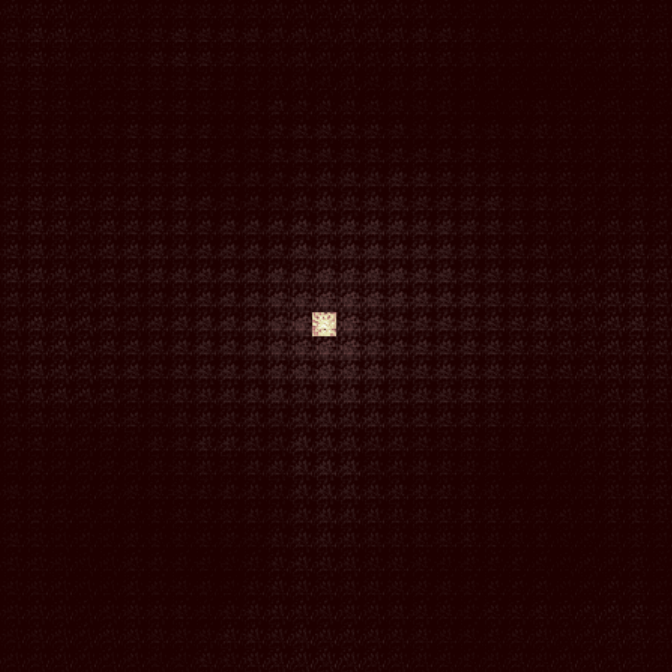}
    \subcaption{Block 1}
    \label{figure:erf_deit_448:0}
  \end{minipage}
  \begin{minipage}[t]{0.13\hsize}
    \centering
    \includegraphics[width=\linewidth]{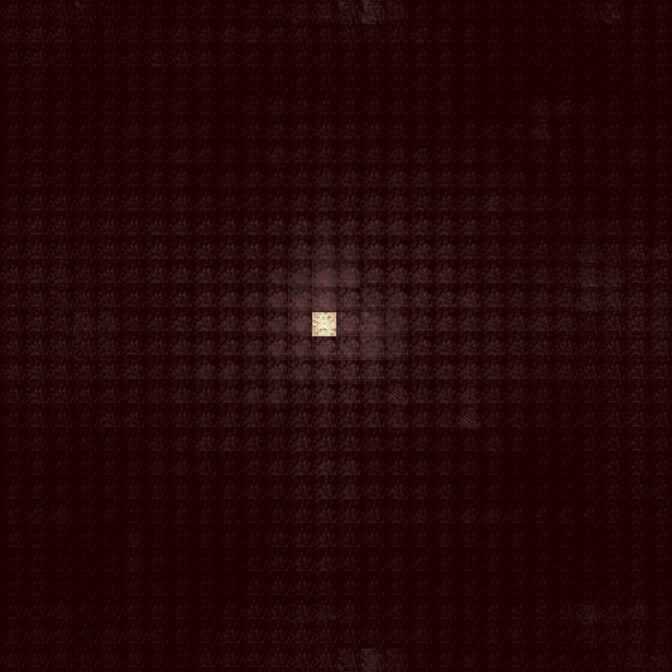}
    \subcaption{Block 2}
    \label{figure:erf_deit_448:1}
  \end{minipage}
  \begin{minipage}[t]{0.13\hsize}
    \centering
    \includegraphics[width=\linewidth]{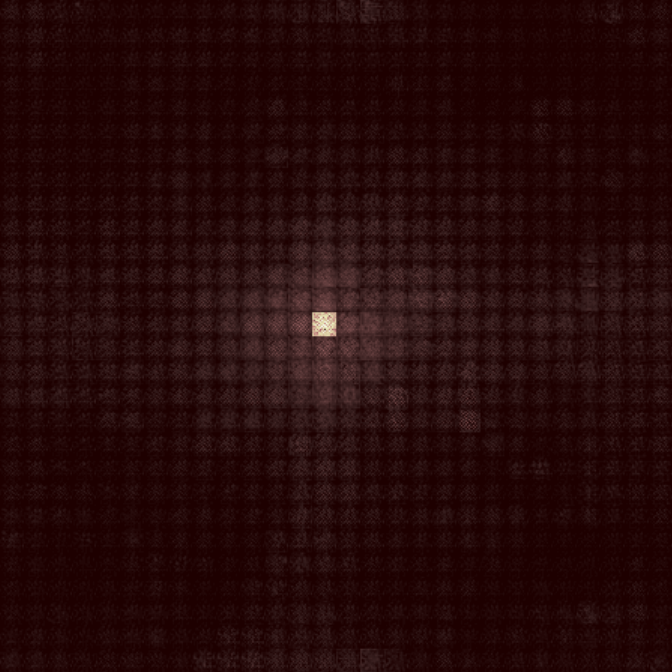}
    \subcaption{Block 3}
    \label{figure:erf_deit_448:2}
  \end{minipage}
  \begin{minipage}[t]{0.13\hsize}
    \centering
    \includegraphics[width=\linewidth]{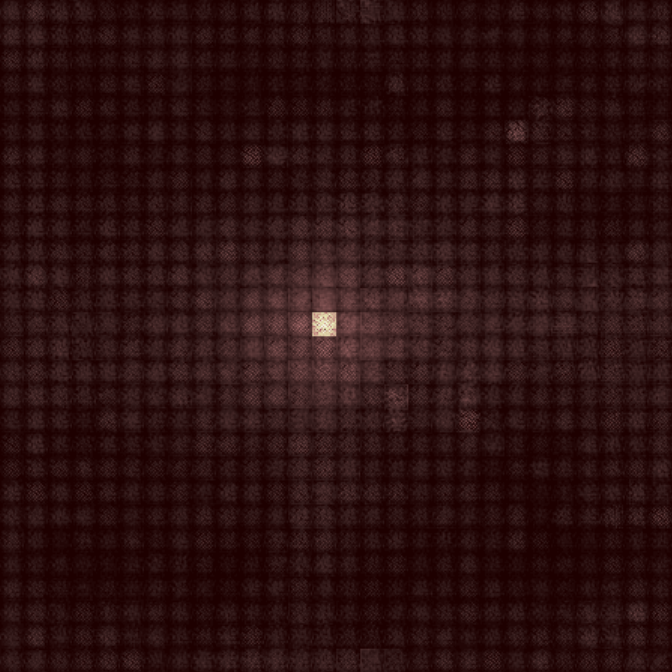}
    \subcaption{Block 4}
    \label{figure:erf_deit_448:3}
  \end{minipage}
  \begin{minipage}[t]{0.13\hsize}
    \centering
    \includegraphics[width=\linewidth]{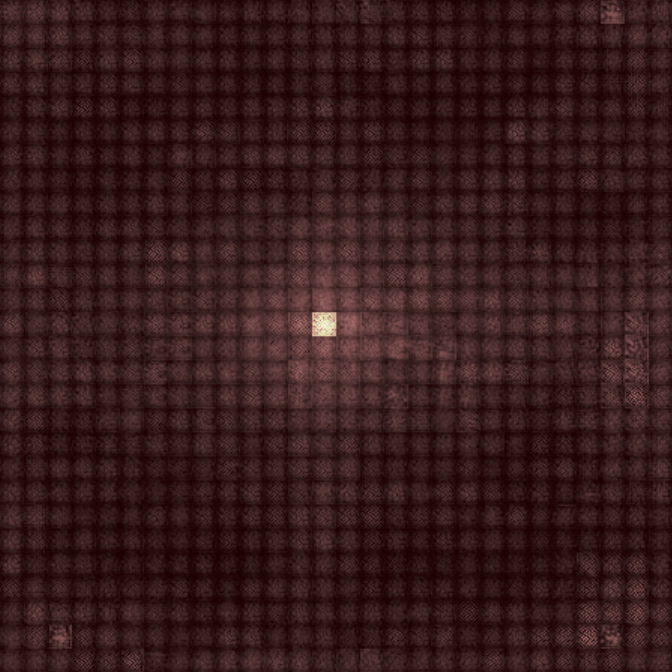}
    \subcaption{Block 5}
    \label{figure:erf_deit_448:4}
  \end{minipage}
  \begin{minipage}[t]{0.13\hsize}
    \centering
    \includegraphics[width=\linewidth]{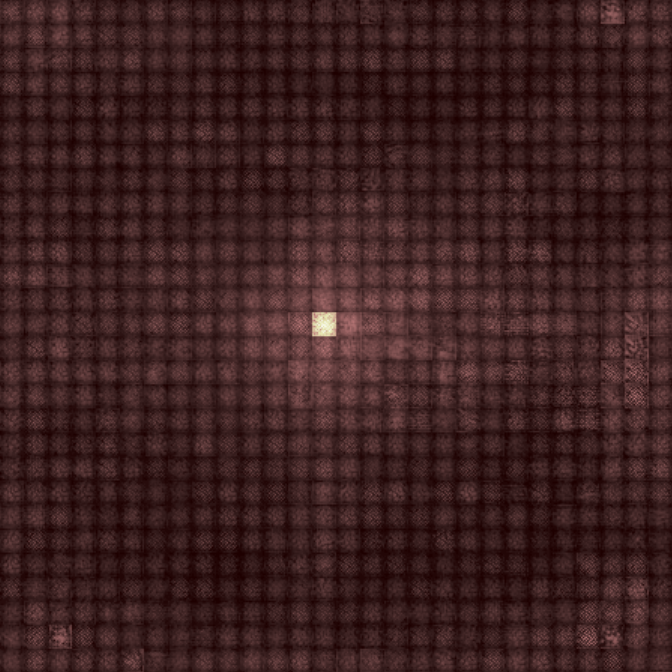}
    \subcaption{Block 6}
    \label{figure:erf_deit_448:5}
  \end{minipage}
  \begin{minipage}[t]{0.13\hsize}
    \centering
    \includegraphics[width=\linewidth]{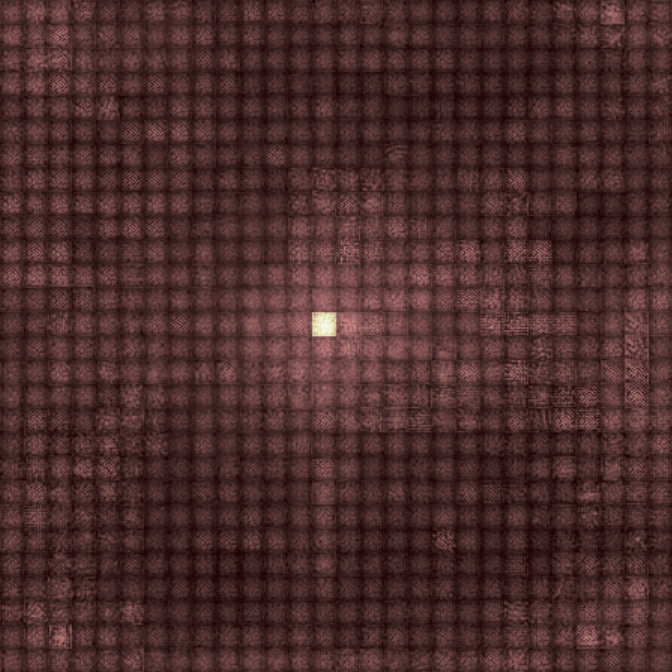}
    \subcaption{Block 7}
    \label{figure:erf_deit_448:6}
  \end{minipage}
  \begin{minipage}[t]{0.13\hsize}
    \centering
    \includegraphics[width=\linewidth]{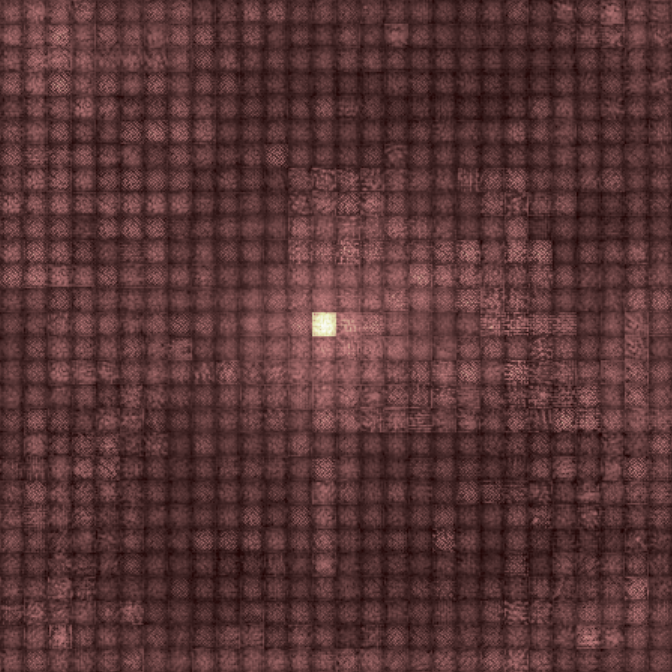}
    \subcaption{Block 8}
    \label{figure:erf_deit_448:7}
  \end{minipage}
  \begin{minipage}[t]{0.13\hsize}
    \centering
    \includegraphics[width=\linewidth]{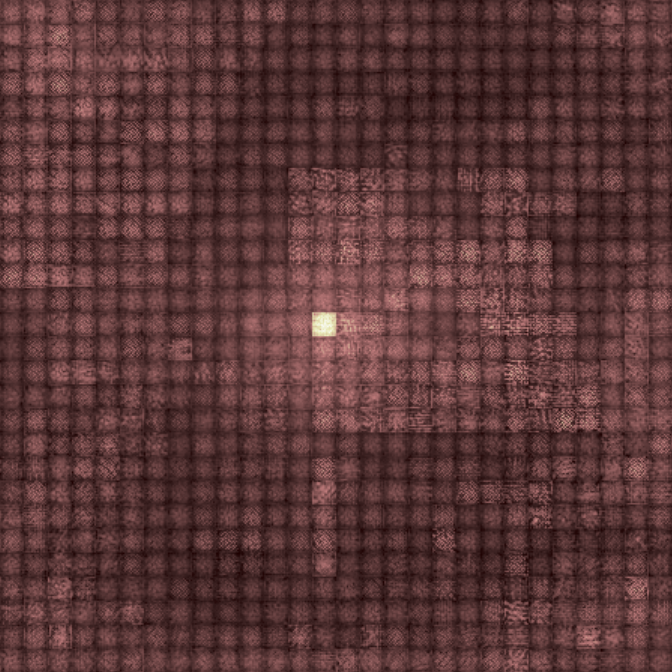}
    \subcaption{Block 9}
    \label{figure:erf_deit_448:8}
  \end{minipage}
  \begin{minipage}[t]{0.13\hsize}
    \centering
    \includegraphics[width=\linewidth]{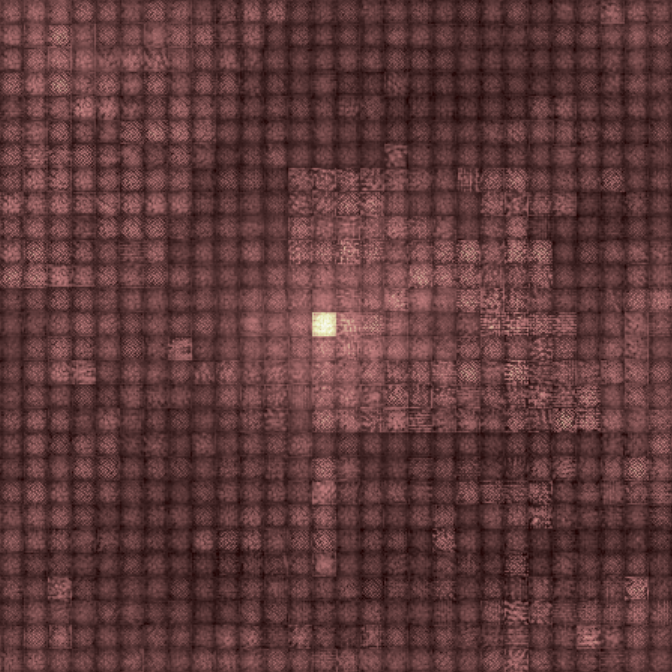}
    \subcaption{Block 10}
    \label{figure:erf_deit_448:9}
  \end{minipage}
  \begin{minipage}[t]{0.13\hsize}
    \centering
    \includegraphics[width=\linewidth]{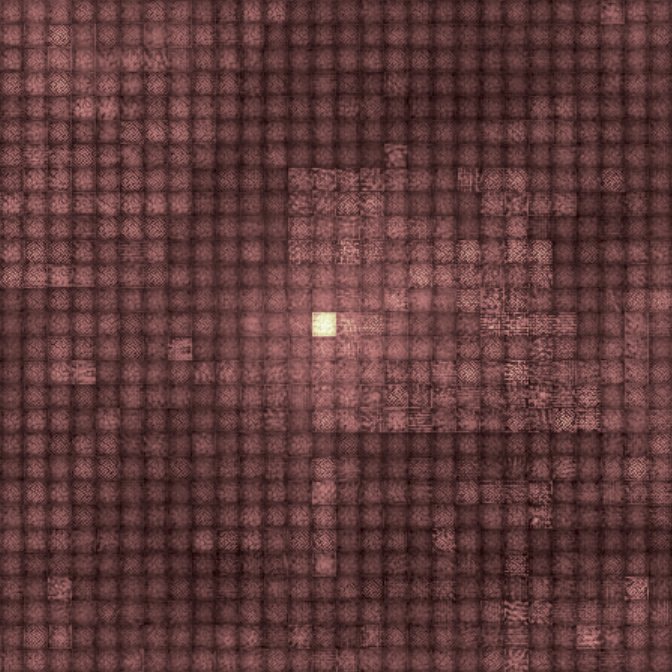}
    \subcaption{Block 11}
    \label{figure:erf_deit_448:10}
  \end{minipage}
  \begin{minipage}[t]{0.13\hsize}
    \centering
    \includegraphics[width=\linewidth]{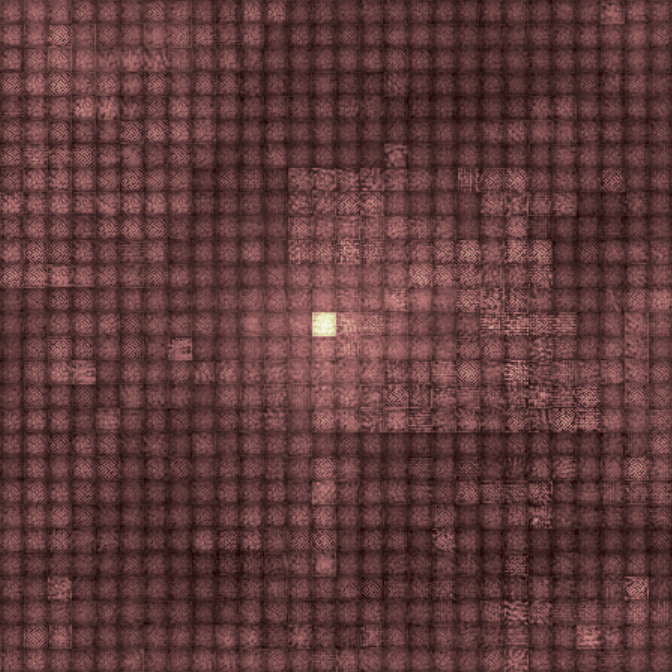}
    \subcaption{Block 12}
    \label{figure:erf_deit_448:11}
  \end{minipage}
  \caption{ERFs in DeiT-S \cite{touvron2020training} on images with resolution $448^2$.}
    \label{figure:erf_deit_448}
\end{figure}

%% file: neurips_2022.bbl
\begin{thebibliography}{10}

\bibitem{ba2016layer}
Jimmy~Lei Ba, Jamie~Ryan Kiros, and Geoffrey~E Hinton.
\newblock Layer normalization.
\newblock In {\em NeurIPS}, 2016.

\bibitem{beyer2020we}
Lucas Beyer, Olivier~J H{\'e}naff, Alexander Kolesnikov, Xiaohua Zhai, and
  A{\"a}ron van~den Oord.
\newblock Are we done with imagenet?
\newblock {\em arXiv preprint arXiv:2006.07159}, 2020.

\bibitem{brown2020language}
Tom Brown, Benjamin Mann, Nick Ryder, Melanie Subbiah, Jared~D Kaplan, Prafulla
  Dhariwal, Arvind Neelakantan, Pranav Shyam, Girish Sastry, Amanda Askell,
  et~al.
\newblock Language models are few-shot learners.
\newblock In {\em NeurIPS}, volume~33, pages 1877--1901, 2020.

\bibitem{byeon2015scene}
Wonmin Byeon, Thomas~M Breuel, Federico Raue, and Marcus Liwicki.
\newblock Scene labeling with lstm recurrent neural networks.
\newblock In {\em CVPR}, pages 3547--3555, 2015.

\bibitem{chen2021crossvit}
Chun-Fu~Richard Chen, Quanfu Fan, and Rameswar Panda.
\newblock Crossvit: Cross-attention multi-scale vision transformer for image
  classification.
\newblock In {\em ICCV}, pages 357--366, 2021.

\bibitem{chen2020generative}
Mark Chen, Alec Radford, Rewon Child, Jeffrey Wu, Heewoo Jun, David Luan, and
  Ilya Sutskever.
\newblock Generative pretraining from pixels.
\newblock In {\em ICML}, pages 1691--1703. PMLR, 2020.

\bibitem{chen2022cyclemlp}
Shoufa Chen, Enze Xie, Chongjian GE, Runjian Chen, Ding Liang, and Ping Luo.
\newblock Cycle{MLP}: A {MLP}-like architecture for dense prediction.
\newblock In {\em ICLR}, 2022.

\bibitem{cho2014properties}
Kyunghyun Cho, Bart van Merri{\"e}nboer, Dzmitry Bahdanau, and Yoshua Bengio.
\newblock On the properties of neural machine translation: Encoder--decoder
  approaches.
\newblock In {\em Proceedings of SSST-8, Eighth Workshop on Syntax, Semantics
  and Structure in Statistical Translation}, pages 103--111, 2014.

\bibitem{chollet2017xception}
Fran{\c{c}}ois Chollet.
\newblock Xception: Deep learning with depthwise separable convolutions.
\newblock In {\em CVPR}, pages 1251--1258, 2017.

\bibitem{cooley1965algorithm}
James~W Cooley and John~W Tukey.
\newblock An algorithm for the machine calculation of complex fourier series.
\newblock {\em Mathematics of computation}, 19(90):297--301, 1965.

\bibitem{cubuk2020randaugment}
Ekin~D Cubuk, Barret Zoph, Jonathon Shlens, and Quoc~V Le.
\newblock {RandAugment}: Practical automated data augmentation with a reduced
  search space.
\newblock In {\em CVPRW}, pages 702--703, 2020.

\bibitem{devries2017improved}
Terrance DeVries and Graham~W Taylor.
\newblock Improved regularization of convolutional neural networks with cutout.
\newblock {\em arXiv preprint arXiv:1708.04552}, 2017.

\bibitem{ding2021repmlpnet}
Xiaohan Ding, Honghao Chen, Xiangyu Zhang, Jungong Han, and Guiguang Ding.
\newblock Repmlpnet: Hierarchical vision mlp with re-parameterized locality.
\newblock In {\em CVPR}, 2022.

\bibitem{ding2022scaling}
Xiaohan Ding, Xiangyu Zhang, Yizhuang Zhou, Jungong Han, Guiguang Ding, and
  Jian Sun.
\newblock Scaling up your kernels to 31x31: Revisiting large kernel design in
  cnns.
\newblock In {\em CVPR}, 2022.

\bibitem{dong2022cswin}
Xiaoyi Dong, Jianmin Bao, Dongdong Chen, Weiming Zhang, Nenghai Yu, Lu~Yuan,
  Dong Chen, and Baining Guo.
\newblock Cswin transformer: A general vision transformer backbone with
  cross-shaped windows.
\newblock In {\em CVPR}, 2022.

\bibitem{dosovitskiy2020image}
Alexey Dosovitskiy, Lucas Beyer, Alexander Kolesnikov, Dirk Weissenborn,
  Xiaohua Zhai, Thomas Unterthiner, Mostafa Dehghani, Matthias Minderer, Georg
  Heigold, Sylvain Gelly, et~al.
\newblock An image is worth 16x16 words: Transformers for image recognition at
  scale.
\newblock In {\em ICLR}, 2021.

\bibitem{gholami2018squeezenext}
Amir Gholami, Kiseok Kwon, Bichen Wu, Zizheng Tai, Xiangyu Yue, Peter Jin,
  Sicheng Zhao, and Kurt Keutzer.
\newblock Squeezenext: Hardware-aware neural network design.
\newblock In {\em CVPRW}, pages 1638--1647, 2018.

\bibitem{goodfellow2014explaining}
Ian~J Goodfellow, Jonathon Shlens, and Christian Szegedy.
\newblock Explaining and harnessing adversarial examples.
\newblock In {\em ICLR}, 2015.

\bibitem{graves2007multi}
Alex Graves, Santiago Fern{\'a}ndez, and J{\"u}rgen Schmidhuber.
\newblock Multi-dimensional recurrent neural networks.
\newblock In {\em International conference on artificial neural networks},
  pages 549--558. Springer, 2007.

\bibitem{graves2008offline}
Alex Graves and J{\"u}rgen Schmidhuber.
\newblock Offline handwriting recognition with multidimensional recurrent
  neural networks.
\newblock In {\em NeurIPS}, volume~21, 2008.

\bibitem{guo2021hire}
Jianyuan Guo, Yehui Tang, Kai Han, Xinghao Chen, Han Wu, Chao Xu, Chang Xu, and
  Yunhe Wang.
\newblock Hire-mlp: Vision mlp via hierarchical rearrangement.
\newblock {\em arXiv preprint arXiv:2108.13341}, 2021.

\bibitem{he2016deep}
Kaiming He, Xiangyu Zhang, Shaoqing Ren, and Jian Sun.
\newblock Deep residual learning for image recognition.
\newblock In {\em CVPR}, pages 770--778, 2016.

\bibitem{hendrycks2021many}
Dan Hendrycks, Steven Basart, Norman Mu, Saurav Kadavath, Frank Wang, Evan
  Dorundo, Rahul Desai, Tyler Zhu, Samyak Parajuli, Mike Guo, et~al.
\newblock The many faces of robustness: A critical analysis of
  out-of-distribution generalization.
\newblock In {\em ICCV}, pages 8340--8349, 2021.

\bibitem{hendrycks2018benchmarking}
Dan Hendrycks and Thomas Dietterich.
\newblock Benchmarking neural network robustness to common corruptions and
  perturbations.
\newblock In {\em ICLR}, 2018.

\bibitem{hendrycks2016gaussian}
Dan Hendrycks and Kevin Gimpel.
\newblock Gaussian error linear units ({GELUs}).
\newblock {\em arXiv preprint arXiv:1606.08415}, 2016.

\bibitem{hendrycks2021natural}
Dan Hendrycks, Kevin Zhao, Steven Basart, Jacob Steinhardt, and Dawn Song.
\newblock Natural adversarial examples.
\newblock In {\em CVPR}, pages 15262--15271, 2021.

\bibitem{hochreiter1997long}
Sepp Hochreiter and J{\"u}rgen Schmidhuber.
\newblock Long short-term memory.
\newblock {\em Neural computation}, 9(8):1735--1780, 1997.

\bibitem{hou2022vision}
Qibin Hou, Zihang Jiang, Li~Yuan, Ming-Ming Cheng, Shuicheng Yan, and Jiashi
  Feng.
\newblock Vision permutator: A permutable mlp-like architecture for visual
  recognition.
\newblock {\em IEEE TPAMI}, 2022.

\bibitem{huang2017densely}
Gao Huang, Zhuang Liu, Laurens Van Der~Maaten, and Kilian~Q Weinberger.
\newblock Densely connected convolutional networks.
\newblock In {\em CVPR}, pages 4700--4708, 2017.

\bibitem{huang2016deep}
Gao Huang, Yu~Sun, Zhuang Liu, Daniel Sedra, and Kilian~Q Weinberger.
\newblock Deep networks with stochastic depth.
\newblock In {\em ECCV}, pages 646--661, 2016.

\bibitem{huang2019ccnet}
Zilong Huang, Xinggang Wang, Lichao Huang, Chang Huang, Yunchao Wei, and Wenyu
  Liu.
\newblock Ccnet: Criss-cross attention for semantic segmentation.
\newblock In {\em ICCV}, pages 603--612, 2019.

\bibitem{kalchbrenner2015grid}
Nal Kalchbrenner, Ivo Danihelka, and Alex Graves.
\newblock Grid long short-term memory.
\newblock In {\em In Proceedings of the IEEE Workshop on Spoken Language
  Technology}, 2015.

\bibitem{kasai2021finetuning}
Jungo Kasai, Hao Peng, Yizhe Zhang, Dani Yogatama, Gabriel Ilharco, Nikolaos
  Pappas, Yi~Mao, Weizhu Chen, and Noah~A Smith.
\newblock Finetuning pretrained transformers into rnns.
\newblock In {\em EMNLP}, pages 10630--10643, 2021.

\bibitem{katharopoulos2020transformers}
Angelos Katharopoulos, Apoorv Vyas, Nikolaos Pappas, and Fran{\c{c}}ois
  Fleuret.
\newblock Transformers are rnns: Fast autoregressive transformers with linear
  attention.
\newblock In {\em ICML}, pages 5156--5165. PMLR, 2020.

\bibitem{devlin2018bert}
Jacob Devlin Ming-Wei~Chang Kenton and Lee~Kristina Toutanova.
\newblock Bert: Pre-training of deep bidirectional transformers for language
  understanding.
\newblock In {\em NAACL-HLT}, pages 4171--4186, 2019.

\bibitem{kirillov2019panoptic}
Alexander Kirillov, Ross Girshick, Kaiming He, and Piotr Doll{\'a}r.
\newblock Panoptic feature pyramid networks.
\newblock In {\em CVPR}, pages 6399--6408, 2019.

\bibitem{krause20133d}
Jonathan Krause, Michael Stark, Jia Deng, and Li~Fei-Fei.
\newblock {3D} object representations for fine-grained categorization.
\newblock In {\em ICCVW}, pages 554--561, 2013.

\bibitem{krizhevsky2009learning}
Alex Krizhevsky, Geoffrey Hinton, et~al.
\newblock Learning multiple layers of features from tiny images.
\newblock Technical report, University of Toronto, 2009.

\bibitem{krizhevsky2012imagenet}
Alex Krizhevsky, Ilya Sutskever, and Geoffrey~E Hinton.
\newblock Imagenet classification with deep convolutional neural networks.
\newblock In {\em NeurIPS}, volume~25, pages 1097--1105, 2012.

\bibitem{lan2021couplformer}
Hai Lan, Xihao Wang, and Xian Wei.
\newblock Couplformer: Rethinking vision transformer with coupling attention
  map.
\newblock {\em arXiv preprint arXiv:2112.05425}, 2021.

\bibitem{lecun1998gradient}
Yann LeCun, L{\'e}on Bottou, Yoshua Bengio, and Patrick Haffner.
\newblock Gradient-based learning applied to document recognition.
\newblock {\em Proceedings of the IEEE}, 86(11):2278--2324, 1998.

\bibitem{lian2021mlp}
Dongze Lian, Zehao Yu, Xing Sun, and Shenghua Gao.
\newblock As-mlp: An axial shifted mlp architecture for vision.
\newblock In {\em ICLR}, 2022.

\bibitem{liang2016semantic}
Xiaodan Liang, Xiaohui Shen, Donglai Xiang, Jiashi Feng, Liang Lin, and
  Shuicheng Yan.
\newblock Semantic object parsing with local-global long short-term memory.
\newblock In {\em CVPR}, pages 3185--3193, 2016.

\bibitem{lin2017focal}
Tsung-Yi Lin, Priya Goyal, Ross Girshick, Kaiming He, and Piotr Doll{\'a}r.
\newblock Focal loss for dense object detection.
\newblock In {\em ICCV}, pages 2980--2988, 2017.

\bibitem{lin2014microsoft}
Tsung-Yi Lin, Michael Maire, Serge Belongie, James Hays, Pietro Perona, Deva
  Ramanan, Piotr Doll{\'a}r, and C~Lawrence Zitnick.
\newblock Microsoft coco: Common objects in context.
\newblock In {\em ECCV}, pages 740--755, 2014.

\bibitem{liu2021pay}
Hanxiao Liu, Zihang Dai, David So, and Quoc Le.
\newblock Pay attention to mlps.
\newblock In {\em NeurIPS}, volume~34, 2021.

\bibitem{liu2022swin}
Ze~Liu, Han Hu, Yutong Lin, Zhuliang Yao, Zhenda Xie, Yixuan Wei, Jia Ning, Yue
  Cao, Zheng Zhang, Li~Dong, et~al.
\newblock Swin transformer v2: Scaling up capacity and resolution.
\newblock In {\em CVPR}, 2022.

\bibitem{liu2021swin}
Ze~Liu, Yutong Lin, Yue Cao, Han Hu, Yixuan Wei, Zheng Zhang, Stephen Lin, and
  Baining Guo.
\newblock Swin transformer: Hierarchical vision transformer using shifted
  windows.
\newblock In {\em ICCV}, 2021.

\bibitem{liu2022convnet}
Zhuang Liu, Hanzi Mao, Chao-Yuan Wu, Christoph Feichtenhofer, Trevor Darrell,
  and Saining Xie.
\newblock A convnet for the 2020s.
\newblock In {\em CVPR}, 2022.

\bibitem{loshchilov2017decoupled}
Ilya Loshchilov and Frank Hutter.
\newblock Decoupled weight decay regularization.
\newblock In {\em ICLR}, 2019.

\bibitem{luo2016understanding}
Wenjie Luo, Yujia Li, Raquel Urtasun, and Richard Zemel.
\newblock Understanding the effective receptive field in deep convolutional
  neural networks.
\newblock In {\em NeurIPS}, volume~29, 2016.

\bibitem{madry2017towards}
Aleksander Madry, Aleksandar Makelov, Ludwig Schmidt, Dimitris Tsipras, and
  Adrian Vladu.
\newblock Towards deep learning models resistant to adversarial attacks.
\newblock In {\em ICLR}, 2018.

\bibitem{mao2021towards}
Xiaofeng Mao, Gege Qi, Yuefeng Chen, Xiaodan Li, Ranjie Duan, Shaokai Ye, Yuan
  He, and Hui Xue.
\newblock Towards robust vision transformer.
\newblock {\em arXiv preprint arXiv:2105.07926}, 2021.

\bibitem{netzer2011reading}
Yuval Netzer, Tao Wang, Adam Coates, Alessandro Bissacco, Bo~Wu, and Andrew~Y
  Ng.
\newblock Reading digits in natural images with unsupervised feature learning.
\newblock In {\em NeurIPS Workshop}, 2011.

\bibitem{nilsback2008automated}
Maria-Elena Nilsback and Andrew Zisserman.
\newblock Automated flower classification over a large number of classes.
\newblock In {\em ICVGIP}, pages 722--729, 2008.

\bibitem{paszke2019pytorch}
Adam Paszke, Sam Gross, Francisco Massa, Adam Lerer, James Bradbury, Gregory
  Chanan, Trevor Killeen, Zeming Lin, Natalia Gimelshein, Luca Antiga, et~al.
\newblock Pytorch: An imperative style, high-performance deep learning library.
\newblock In {\em NeurIPS}, volume~32, 2019.

\bibitem{radford2018improving}
Alec Radford, Karthik Narasimhan, Tim Salimans, and Ilya Sutskever.
\newblock Improving language understanding by generative pre-training.
\newblock {\em Technical report, OpenAI}, 2018.

\bibitem{radford2019language}
Alec Radford, Jeffrey Wu, Rewon Child, David Luan, Dario Amodei, Ilya
  Sutskever, et~al.
\newblock Language models are unsupervised multitask learners.
\newblock {\em OpenAI blog}, 1(8):9, 2019.

\bibitem{radosavovic2020designing}
Ilija Radosavovic, Raj~Prateek Kosaraju, Ross Girshick, Kaiming He, and Piotr
  Doll{\'a}r.
\newblock Designing network design spaces.
\newblock In {\em CVPR}, pages 10428--10436, 2020.

\bibitem{raffel2020exploring}
Colin Raffel, Noam Shazeer, Adam Roberts, Katherine Lee, Sharan Narang, Michael
  Matena, Yanqi Zhou, Wei Li, and Peter~J Liu.
\newblock Exploring the limits of transfer learning with a unified text-to-text
  transformer.
\newblock {\em JMLR}, 21:1--67, 2020.

\bibitem{rao2021global}
Yongming Rao, Wenliang Zhao, Zheng Zhu, Jiwen Lu, and Jie Zhou.
\newblock Global filter networks for image classification.
\newblock In {\em NeurIPS}, volume~34, 2021.

\bibitem{recht2019imagenet}
Benjamin Recht, Rebecca Roelofs, Ludwig Schmidt, and Vaishaal Shankar.
\newblock Do imagenet classifiers generalize to imagenet?
\newblock In {\em ICML}, pages 5389--5400. PMLR, 2019.

\bibitem{schuster1997bidirectional}
Mike Schuster and Kuldip~K Paliwal.
\newblock Bidirectional recurrent neural networks.
\newblock {\em IEEE TSP}, 45(11):2673--2681, 1997.

\bibitem{simonyan2014very}
Karen Simonyan and Andrew Zisserman.
\newblock Very deep convolutional networks for large-scale image recognition.
\newblock In {\em ICLR}, 2015.

\bibitem{szegedy2015going}
Christian Szegedy, Wei Liu, Yangqing Jia, Pierre Sermanet, Scott Reed, Dragomir
  Anguelov, Dumitru Erhan, Vincent Vanhoucke, and Andrew Rabinovich.
\newblock Going deeper with convolutions.
\newblock In {\em CVPR}, pages 1--9, 2015.

\bibitem{szegedy2016rethinking}
Christian Szegedy, Vincent Vanhoucke, Sergey Ioffe, Jon Shlens, and Zbigniew
  Wojna.
\newblock Rethinking the inception architecture for computer vision.
\newblock In {\em CVPR}, pages 2818--2826, 2016.

\bibitem{tan2019efficientnet}
Mingxing Tan and Quoc Le.
\newblock {EfficientNet}: Rethinking model scaling for convolutional neural
  networks.
\newblock In {\em ICML}, pages 6105--6114, 2019.

\bibitem{tang2021sparse}
Chuanxin Tang, Yucheng Zhao, Guangting Wang, Chong Luo, Wenxuan Xie, and Wenjun
  Zeng.
\newblock Sparse mlp for image recognition: Is self-attention really necessary?
\newblock {\em arXiv preprint arXiv:2109.05422}, 2021.

\bibitem{tatsunami2021raftmlp}
Yuki Tatsunami and Masato Taki.
\newblock Raftmlp: How much can be done without attention and with less spatial
  locality?
\newblock {\em arXiv preprint arXiv:2108.04384}, 2021.

\bibitem{tolstikhin2021mlp}
Ilya~O Tolstikhin, Neil Houlsby, Alexander Kolesnikov, Lucas Beyer, Xiaohua
  Zhai, Thomas Unterthiner, Jessica Yung, Andreas Steiner, Daniel Keysers,
  Jakob Uszkoreit, et~al.
\newblock Mlp-mixer: An all-mlp architecture for vision.
\newblock In {\em NeurIPS}, volume~34, 2021.

\bibitem{touvron2021resmlp}
Hugo Touvron, Piotr Bojanowski, Mathilde Caron, Matthieu Cord, Alaaeldin
  El-Nouby, Edouard Grave, Armand Joulin, Gabriel Synnaeve, Jakob Verbeek, and
  Herv{\'e} J{\'e}gou.
\newblock Resmlp: Feedforward networks for image classification with
  data-efficient training.
\newblock {\em arXiv preprint arXiv:2105.03404}, 2021.

\bibitem{touvron2020training}
Hugo Touvron, Matthieu Cord, Matthijs Douze, Francisco Massa, Alexandre
  Sablayrolles, and Herv{\'e} J{\'e}gou.
\newblock Training data-efficient image transformers \& distillation through
  attention.
\newblock In {\em ICML}, 2021.

\bibitem{touvron2021going}
Hugo Touvron, Matthieu Cord, Alexandre Sablayrolles, Gabriel Synnaeve, and
  Herv{\'e} J{\'e}gou.
\newblock Going deeper with image transformers.
\newblock In {\em ICCV}, pages 32--42, 2021.

\bibitem{van2016pixel}
Aaron Van~Oord, Nal Kalchbrenner, and Koray Kavukcuoglu.
\newblock Pixel recurrent neural networks.
\newblock In {\em International conference on machine learning}, pages
  1747--1756. PMLR, 2016.

\bibitem{vaswani2017attention}
Ashish Vaswani, Noam Shazeer, Niki Parmar, Jakob Uszkoreit, Llion Jones,
  Aidan~N Gomez, {\L}ukasz Kaiser, and Illia Polosukhin.
\newblock Attention is all you need.
\newblock In {\em NeurIPS}, volume~30, 2017.

\bibitem{visin2016reseg}
Francesco Visin, Marco Ciccone, Adriana Romero, Kyle Kastner, Kyunghyun Cho,
  Yoshua Bengio, Matteo Matteucci, and Aaron Courville.
\newblock Reseg: A recurrent neural network-based model for semantic
  segmentation.
\newblock In {\em CVPRW}, pages 41--48, 2016.

\bibitem{visin2015renet}
Francesco Visin, Kyle Kastner, Kyunghyun Cho, Matteo Matteucci, Aaron
  Courville, and Yoshua Bengio.
\newblock Renet: A recurrent neural network based alternative to convolutional
  networks.
\newblock {\em arXiv preprint arXiv:1505.00393}, 2015.

\bibitem{wang2019learning}
Haohan Wang, Songwei Ge, Zachary Lipton, and Eric~P Xing.
\newblock Learning robust global representations by penalizing local predictive
  power.
\newblock In {\em NeurIPS}, volume~32, 2019.

\bibitem{wang2021pyramid}
Wenhai Wang, Enze Xie, Xiang Li, Deng-Ping Fan, Kaitao Song, Ding Liang, Tong
  Lu, Ping Luo, and Ling Shao.
\newblock Pyramid vision transformer: A versatile backbone for dense prediction
  without convolutions.
\newblock In {\em ICCV}, 2021.

\bibitem{rw2019timm}
Ross Wightman.
\newblock Pytorch image models.
\newblock \url{https://github.com/rwightman/pytorch-image-models}, 2019.

\bibitem{yu2021s}
Tan Yu, Xu~Li, Yunfeng Cai, Mingming Sun, and Ping Li.
\newblock S$^2$-mlpv2: Improved spatial-shift mlp architecture for vision.
\newblock {\em arXiv preprint arXiv:2108.01072}, 2021.

\bibitem{yu2022s2}
Tan Yu, Xu~Li, Yunfeng Cai, Mingming Sun, and Ping Li.
\newblock S$^2$-mlp: Spatial-shift mlp architecture for vision.
\newblock In {\em WACV}, pages 297--306, 2022.

\bibitem{yu2021metaformer}
Weihao Yu, Mi~Luo, Pan Zhou, Chenyang Si, Yichen Zhou, Xinchao Wang, Jiashi
  Feng, and Shuicheng Yan.
\newblock Metaformer is actually what you need for vision.
\newblock In {\em CVPR}, 2022.

\bibitem{yuan2021tokens}
Li~Yuan, Yunpeng Chen, Tao Wang, Weihao Yu, Yujun Shi, Zi-Hang Jiang,
  Francis~EH Tay, Jiashi Feng, and Shuicheng Yan.
\newblock Tokens-to-token vit: Training vision transformers from scratch on
  imagenet.
\newblock In {\em ICCV}, pages 558--567, 2021.

\bibitem{yun2019cutmix}
Sangdoo Yun, Dongyoon Han, Seong~Joon Oh, Sanghyuk Chun, Junsuk Choe, and
  Youngjoon Yoo.
\newblock Cutmix: Regularization strategy to train strong classifiers with
  localizable features.
\newblock In {\em ICCV}, pages 6023--6032, 2019.

\bibitem{zhang2021morphmlp}
David~Junhao Zhang, Kunchang Li, Yunpeng Chen, Yali Wang, Shashwat Chandra,
  Yu~Qiao, Luoqi Liu, and Mike~Zheng Shou.
\newblock Morphmlp: A self-attention free, mlp-like backbone for image and
  video.
\newblock {\em arXiv preprint arXiv:2111.12527}, 2021.

\bibitem{zhang2017mixup}
Hongyi Zhang, Moustapha Cisse, Yann~N Dauphin, and David Lopez-Paz.
\newblock mixup: Beyond empirical risk minimization.
\newblock In {\em ICLR}, 2018.

\bibitem{zhong2020random}
Zhun Zhong, Liang Zheng, Guoliang Kang, Shaozi Li, and Yi~Yang.
\newblock Random erasing data augmentation.
\newblock In {\em AAAI}, volume~34, pages 13001--13008, 2020.

\bibitem{zhou2017scene}
Bolei Zhou, Hang Zhao, Xavier Puig, Sanja Fidler, Adela Barriuso, and Antonio
  Torralba.
\newblock Scene parsing through ade20k dataset.
\newblock In {\em CVPR}, pages 633--641, 2017.

\bibitem{zhou2021deepvit}
Daquan Zhou, Bingyi Kang, Xiaojie Jin, Linjie Yang, Xiaochen Lian, Zihang
  Jiang, Qibin Hou, and Jiashi Feng.
\newblock Deepvit: Towards deeper vision transformer.
\newblock {\em arXiv preprint arXiv:2103.11886}, 2021.

\bibitem{zhou2021refiner}
Daquan Zhou, Yujun Shi, Bingyi Kang, Weihao Yu, Zihang Jiang, Yuan Li, Xiaojie
  Jin, Qibin Hou, and Jiashi Feng.
\newblock Refiner: Refining self-attention for vision transformers.
\newblock {\em arXiv preprint arXiv:2106.03714}, 2021.

\end{thebibliography}
